\documentclass{article}

\usepackage{microtype}
\usepackage{graphicx}
\usepackage{subfigure}
\usepackage{booktabs} 
\usepackage{tabularx} 
\usepackage{siunitx}  
\usepackage{multirow}
\usepackage{tikz}
\usepackage{float}
\usepackage{longtable}
\usepackage{svg}
\usepackage{hyperref}

\usepackage[accepted]{icml2024}

\usepackage{amsmath}
\usepackage{amssymb}
\usepackage{mathtools}
\usepackage{amsthm}
\usepackage{xcolor}
\usepackage{ulem}
\usepackage[capitalize,noabbrev]{cleveref}

\usepackage{enumitem}
\usepackage{subfigure}
\usepackage{makecell}
\theoremstyle{plain}

\theoremstyle{definition}

\theoremstyle{remark}

\usepackage[textsize=tiny]{todonotes}

\icmltitlerunning{From Goal-Conditioned to Language-Conditioned Agents via Vision-Language Models}

\DeclareMathOperator*{\argmax}{arg\!\max\,}

\newcommand{\R}{\mathbb{R}}
\newcommand{\configset}{\mathcal{Q}}
\newcommand{\admiset}{\mathcal{Q}^\text{a}} 
\newcommand{\proj}{\mathcal{P}_{\admiset}}
\newcommand{\markov}{\mathcal{M}}
\newcommand{\states}{\mathcal{S}}
\newcommand{\actions}{\mathcal{A}}
\newcommand{\reward}{R}
\newcommand{\imageset}{\mathcal{I}}
\newcommand{\textset}{\mathcal{T}}
\newcommand{\dataset}{\mathcal{D}}
\newcommand{\bq}{q}

\newcommand{\txt}{x}
\newcommand{\img}{I}
\newcommand{\frender}{f^\text{render}}
\newcommand{\ftext}{f^\text{text}}
\newcommand{\fimage}{f^\text{image}}
\newcommand{\fconfig}{f^\text{config}}
\newcommand{\fdist}{\hat{f}^\text{config}}
\newcommand{\imtextsim}{S_\mathrm{it}}
\newcommand{\qtextsim}{S_\mathrm{qt}}
\newcommand{\qtextsurr}{\hat{S}_\mathrm{qt}}
\newcommand{\sco}[1]{\hat{S}^{(#1)}_\mathrm{qt}}
\newcommand{\natproj}{\varphi_{\configset}}
\renewcommand{\emph}[1]{\textit{#1}}

\begin{document}

\twocolumn[
\icmltitle{From Goal-Conditioned to Language-Conditioned Agents\\ via Vision-Language Models}




\begin{icmlauthorlist}
\icmlauthor{Théo Cachet}{naver,sorbonne}
\icmlauthor{Christopher R. Dance}{naver}
\icmlauthor{Olivier Sigaud}{sorbonne}
\end{icmlauthorlist}

\icmlaffiliation{naver}{NAVER LABS Europe, Meylan}
\icmlaffiliation{sorbonne}{Institute of Intelligent Systems and Robotics, Sorbonne University, Paris}

\icmlcorrespondingauthor{Théo Cachet}{theo.cachet@gmail.com}

\icmlkeywords{vision-language models, distillation, reinforcement learning, language-condition agents}

\vskip 0.3in
]


\printAffiliationsAndNotice{}  

\begin{abstract}
Vision-language models (VLMs) have tremendous potential for \emph{grounding} language, and thus enabling \emph{language-conditioned agents (LCAs)} to perform diverse tasks specified with text.  
This has motivated the study of LCAs based on reinforcement learning (RL) with rewards given by rendering images of an environment and evaluating those images with VLMs. 
If single-task RL is employed, such approaches are limited by the cost and time required to train a policy for each new task. 
Multi-task RL (MTRL) is a natural alternative, but requires a carefully designed corpus of training tasks and does not always generalize reliably to new tasks.
Therefore, this paper introduces a novel decomposition of the problem of building an LCA: first find an environment \emph{configuration} that has a high VLM score for text describing a task; then use a (pretrained) goal-conditioned policy to reach that configuration.  
We also explore several enhancements to the speed and quality of VLM-based LCAs, notably, the use of distilled models, and the evaluation of configurations from multiple viewpoints to resolve the ambiguities inherent in a single 2D view.
We demonstrate our approach on the Humanoid environment, showing that it results in LCAs that outperform MTRL baselines in zero-shot generalization, without requiring any textual task descriptions or other forms of environment-specific annotation during training.
\end{abstract}

\section{Introduction}
\label{sec:intro}

\begin{figure*}[ht]
\centering
\includegraphics[width=.999\linewidth]{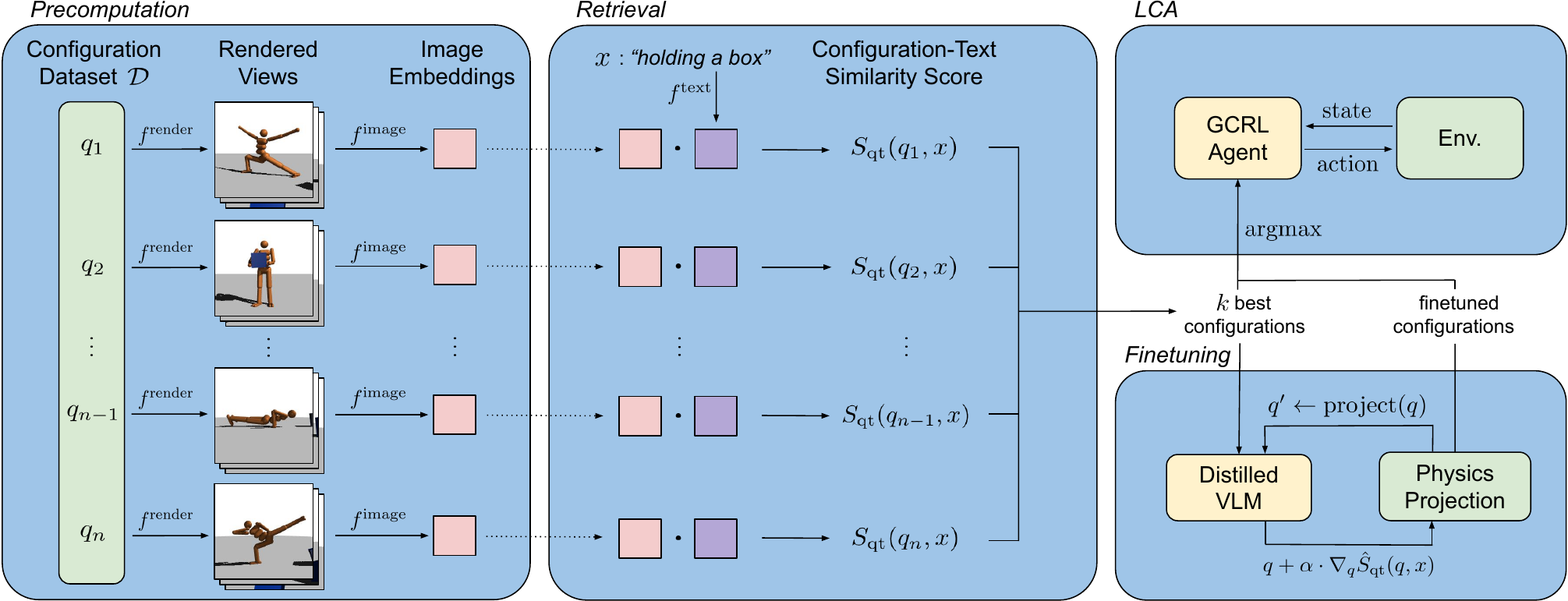}
\caption{Overview of the proposed approach. We use rendering functions and a VLM image encoder to precompute embeddings of all configurations in a dataset. Given text $\txt$ describing a new task, we embed it with the VLM text encoder, evaluate its cosine similarity with the precomputed image embeddings, select $k$ highest-scoring configurations, and (optionally) finetune them using a distilled model. Finally, the best configuration is fed to a pretrained goal-conditioned agent to execute the task. Note that our approach can retrieve and finetune configurations for a text $x$ without ever having seen that text before, and that the GCRL agent can reach a goal configuration without having been specifically trained for that goal; thus, our approach results in a \emph{zero-shot} LCA.} 
\label{fig:overview}
\end{figure*}

The widespread adoption of large language models (LLMs) and text-to-image models in modern society demonstrates the convenience of natural language interaction with AI models~\cite{bubeck2023sparks, rombach2022high}.
This motivates the study of \emph{language-conditioned agents (LCAs)} that can execute diverse commands, specified with text, in a given environment~\citep{colas2020languageconditioned, zhou2023language}. 
Large-scale internet-scraped text and image data is a key enabler of current LLMs and text-to-image models~\cite{schuhmann2022laion, gadre2023datacomp, penedo2023refinedweb}. 
However, data of comparable scale relating text with environments does not exist and human annotation is costly, leading to the question: \emph{how can we learn LCAs given the scarcity of annotated data?}

Recently, foundation models (FMs) have emerged as promising tools to tackle this challenge. 
LLMs have been used to orchestrate predefined motion primitives~\cite{huang2022language, zeng2022socratic, ichter2023do}, but the learning of such primitives also suffers from the lack of annotated data. 
Another approach is to use LLMs or vision-language models (VLMs) to define reward functions for a given text, either by source-code generation~\cite{yu2023language, perez2023larg, ma2023eureka}, or by image-text similarity evaluation~\cite{mahmoudieh2022zero, tam2022semantic, cui2022can, rocamonde2023vision, adeniji2023language}. 
One may couple such rewards with single-task reinforcement learning (STRL), but such approaches are limited by the need to train a policy for each new task, making them too slow for real-time application. 
Training multi-task reinforcement learning (MTRL) agents is a natural alternative~\citep{fan2022minedojo}, which we explore in this paper.  
However, it is not obvious how to construct a training corpus of textual task descriptions that enables MTRL agents to generalize reliably to new tasks~\citep{cobbe2020leveraging, yu2020meta}. 

Therefore, this paper explores a novel decomposition of the problem of building LCAs into \emph{VLM-based text-to-goal generation} and \emph{goal-reaching}, as shown in Figure~\ref{fig:overview}: 
we find a configuration such that rendered images of the environment in this configuration have high VLM scores for a given text; then we use a goal-conditioned reinforcement learning (GCRL) agent to reach that configuration\footnote{Videos and an interactive demo can be found at \url{https://europe.naverlabs.com/text2control}}.
In this paper, \emph{configurations} capture enough of the state to render images of the environment (e.g., current object positions of objects, but not velocities or past positions).   
The decomposition into text-to-goal and goal-reaching was previously proposed by~\citet{colas2020languageconditioned}, but that work relies upon 
highly restricted algorithmic annotations
to learn the text-to-goal component, rather than a VLM.
Our decomposition has several advantages over the MTRL approach: 
it circumvents the problem of choosing a corpus of texts for MTRL; moreover, the reward function for GCRL is less oscillatory and faster to evaluate than the VLM score that might be used by MTRL.
On the other hand, our decomposition precludes tasks with no natural final configuration, such as juggling or cycling, which we hope to study in future work.

We also explore several enhancements to the quality and speed of VLM-based LCAs.
Whereas previous work has used the VLM score of a single view of an environment to determine the reward for a given state, we explore the use of \emph{multiple viewpoints} to mitigate problems of occlusion and distance ambiguity inherent in a single 2D view.
Moreover, we investigate the use of large datasets of diverse configurations with precomputed VLM embeddings, for rapid \emph{retrieval of configurations} corresponding to a given text.
These datasets may be also used to train \emph{distilled models} (of the composition of rendering with the VLM image encoder) for rapid evaluation of VLM scores, accelerating both text-to-goal generation and the training of MTRL agents.
We show that the derivatives of such distilled models with respect to configuration are better behaved than those of the original VLM score, and are well-suited to the finetuning of retrieved configurations.  
We evaluate the proposed methods on the Humanoid environment~\citep{brockman2016openai}, which we augment with feet and a cube, on a set of 256 textually defined tasks, using five state-of-the-art VLMs.  
To summarize the results:
\begin{itemize}[itemsep=0.5mm, parsep=0mm]
\item Our LCA, based on the novel decomposition into VLM-based text-to-goal generation and goal-reaching, attains higher returns than otherwise equivalent MTRL baselines, when performing zero-shot command execution, for 205 out of the 256 tasks.
\item We quantify the viewpoint sensitivities of VLM scores, and give concrete examples showing that multiview evaluation avoids configurations that appear acceptable from one view but are actually pathological.
\item Our distilled model reduces the computation time of VLM-based rewards by up to $40\, 000\times$, while remaining sufficiently accurate that finetuning configurations using the distilled model increases the true VLM score.
\end{itemize}

The paper is structured as follows: we discuss related work (Section~\ref{related_work}), present our methods (Section~\ref{sec:methods}), experimentally evaluate them (Section~\ref{sec:experiments}), and suggest future work (Section~\ref{sec:further-work}). 
The Appendix presents full details about the methods, models, environment, tasks, rendering, as well as task-by-task results for the presented text-to-goal methods and LCAs. 

\section{Related Work}
\label{related_work}

Central to our work is the challenge of \emph{grounding language}: 
the process of linking language with an agent’s observations and actions~\citep{harnad1990symbol}. 
In this section, we first discuss the use of environment-specific annotations, LLMs and VLMs for grounding language and building LCAs. 
Then, we focus on text-to-goal generation methods.

\paragraph{Annotation.} A natural approach to grounding is to gather textual annotations for an environment. For instance, state descriptions have been used to learn language-conditioned goal generators~\citep{colas2020languageconditioned} and language-conditioned reward functions~\cite{bahdanau2018learning, nair2022learning}. Additionally, state-sequence descriptions can be coupled with imitation learning~\cite{stepputtis2020language, lynch2021language, chen2023polarnet} or with inverse reinforcement learning~\cite{fu2019language, zhou2021inverse} to create LCAs. To reduce the cost of human annotation, some works generate annotations algorithmically~\cite{jiang2019language, hill2020human, stooke2021open}, but so far such works are limited to simple tasks such as \textit{``place X near Y''}.

\paragraph{Foundation models (FMs).} Another way to circumvent costly human annotation is to exploit FMs. Given a textual task, LLMs have been used to write the source code of reward functions~\cite{yu2023language, perez2023larg, ma2023eureka} or to orchestrate predefined skills~\cite{huang2022language, ichter2023do}. LLMs have also been coupled with VLMs to leverage visual information~\cite{zeng2022socratic, huang2023voxposer, ajay2023compositional}. However LLMs require 
environment-specific prompting and are prone to hallucination~\cite{huang2023survey}. 

Other works have also explored language grounding using VLMs. For instance, VLMs can be coupled with rendering functions to derive reward functions from natural language~\cite{mahmoudieh2022zero, fan2022minedojo, rocamonde2023vision, baumli2023vision}, to pretrain language-conditioned policies~\cite{adeniji2023language}, to derive extrinsic reward functions for exploration~\cite{tam2022semantic}, and to detect task-completion~\cite{du2023visionlanguage}.
Unfortunately, VLM-based reward functions are costly to evaluate, and they are highly oscillatory (`noisy'), as illustrated in Figure~3 of~\citet{adeniji2023language}, leading to slow and unreliable RL. 
Our approach also relies on VLMs, but we circumvent these difficulties by only using VLMs to find configurations with high VLM scores, as we now discuss.

\paragraph{Text-to-goal methods.} Text-to-goal methods identify (sets of) states that align with a given textual description. These states may then be fed to goal-conditioned policies~\cite{colas2020languageconditioned, akakzia2021grounding} or used to construct hybrid controllers~\citep{Raman2013}, and thus to create LCAs.

With the emergence of large-scale text-to-text~\cite{raffel2020exploring}, text-to-image~\cite{rombach2022high, ramesh2022hierarchical}, text-to-audio~\cite{copet2023simple} and text-to-video~\cite{singer2022make} models, it is interesting to investigate how FMs might serve as components of a text-to-goal procedure. Recently, \citet{gao2023pretrained} proposed to leverage language-conditioned diffusion models to generate images representing goals. However, generating images that closely correspond to a given environment and instruction is challenging, even for state-of-the-art image editing techniques~\cite{brooks2023instructpix2pix, dunlap2023diversify}. Moreover, additional steps are required to derive rewards or goal states from the resulting images: for instance, Gao et al. extend VICE~\citep{fu2018variational} to infer rewards, which is computationally costly and error-prone. 
In contrast, our approach directly generates goal configurations, eliminating the need for image editing, and for such additional steps.

\section{Methods}
\label{sec:methods}
We address the overall problem of finding a language-conditioned policy: given text describing a (potentially previously unseen) task to be performed in an environment, the policy should result in configurations of that environment that correspond well to the given text.
Our approach is to decompose this into two subproblems: finding configurations of the environment with high VLM scores for a given text; and designing goal-conditioned policies to reach such configurations.
This decomposition has two main advantages over end-to-end training of LCAs. 
First, it is easy to generate a huge number of configurations to train a goal-conditioned policy, helping with generalization to new tasks, and avoiding the need to collect a large training set of task descriptions. 
Second, as grounding is independent of low-level control in our approach, it is possible to determine whether failures of the LCA are due to poor alignment of the goal with the task description, or to failure of the low-level controller to reach the goal. 

The section begins with definitions relating configurations with VLM scores, before presenting methods for optimizing such scores that enable a range of speed-quality tradeoffs. Then we discuss the combination of text-to-goal methods with goal-conditioned policies, to craft zero-shot LCAs.

\subsection{Definitions}
Our approach selects \emph{configurations} of an \emph{environment} using \emph{rendering functions} and \emph{VLMs}, as we now explain.

\paragraph{Environment.} The environment is modelled as a controlled Markov process 
\begin{align*}
\markov = (\states, \actions, \rho, \tilde{P}),
\end{align*} 
with state space $\states$, action space $\actions$, initial state distribution $\rho$ and transition distribution $\tilde{P}$.  
The controlled Markov process may be augmented with a reward function $\reward : \states \times \actions \to \R$ to define a Markov decision process (MDP).
In this work, we define various reward functions, using
distances to goal configurations, or VLM scores for a given text (in our baselines).

\paragraph{Configurations.}
Each state of $\states$ is associated with a specific \emph{configuration} $\bq \in \configset$,
which captures the state dimensions relevant to rendering images of the environment.
For instance, a configuration might consist of the angles or positions of an environment's bodies, but not their velocities.
As we use gradient methods to optimize configurations, we assume the configuration space $\configset$ can be represented as a subset of a real Euclidean space.
Typically, not all configurations are admissible: there may be inequality constraints corresponding to the requirement that objects do not interpenetrate (unilateral constraints) or to safety requirements.
We denote the admissible subset of configurations by $\admiset$. 

\paragraph{Rendering functions.} A \emph{rendering function} $\frender : \configset \to \imageset$ maps configurations to images of the environment. 
Suitable rendering functions can be found in MuJoCo~\cite{todorov2012mujoco} and OpenGL~\cite{neider1993opengl}.

\paragraph{VLMs.} A VLM, such as CLIP~\cite{radford2021learning}, typically consists of 
\begin{enumerate}[itemsep=2pt, topsep=0pt, parsep=0pt]
    \item An image encoder \( \fimage : \imageset \to \R^d  \) that maps images to an embedding space of dimension $d$; and 
    \item A text encoder \( \ftext : \textset \rightarrow \R^d \) that maps texts to the same embedding space. The space of texts $\textset$ consists of finite strings on a finite vocabulary.
\end{enumerate}
We assume that the outputs of these encoders are normalized so that $\| \fimage(\cdot) \| = 1$ on $\imageset$ and  $\| \ftext(\cdot) \| = 1$ on $\textset$. 
The \emph{image-text similarity score} (or \emph{VLM score}) for a given image $\img \in \imageset$ and text $\txt \in \textset$ is then defined as the cosine similarity of their embeddings:
\begin{equation}
\imtextsim(\img, \txt) \coloneqq \fimage(\img) \cdot \ftext(\txt).
\label{eq:img_text_clip_score}
\end{equation}

\paragraph{Configuration-text score.} 
Composing a rendering function with the image-text similarity score enables us to define a \emph{configuration-text similarity score} for a given configuration $\bq\in \configset$ and text $\txt \in \textset$: 
\begin{equation}
S_\text{qt}^\text{sv}(\bq, \txt) \coloneqq \imtextsim(\frender(\bq), \txt),
\label{eq:state_text_clip_score}
\end{equation}
where `sv' emphasizes that this is for a single view.
To resolve ambiguities inherent in a single 2D view, we propose to extend this definition to a \emph{multiview configuration-text similarity score} by averaging the similarity scores
for multiple rendering functions $\frender_1, \dots, \frender_m$: 
\begin{align}
\qtextsim(\bq, \txt) &\coloneqq \frac{1}{m} \sum_{k=1}^m \imtextsim(\frender_k(\bq), \txt) \nonumber \\
&= \fconfig(\bq) \cdot \ftext(\txt),
\label{eq:mv_state_text_clip_score}
\end{align}
where the \emph{configuration embedding} $\fconfig : \configset \to \R^d$ is defined by
\begin{align}
\fconfig(\bq) \coloneqq \frac{1}{m} \sum_{k=1}^m \fimage \circ \frender_k(\bq).
\label{eq:confemb}
\end{align}

\paragraph{Distilled model.} Given multiple rendering functions and billion-parameter VLMs, evaluating the configuration-text similarity score can be costly. 
Therefore, we propose to \emph{distill} the configuration embedding into a neural network 
\begin{equation}
\fdist(\bq) \approx \fconfig(\bq).
\label{eq:distilled_model}
\end{equation}
Not only is the distilled model $\fdist$ faster than $\fconfig$, it is also readily differentiated with respect to the configuration: this proves useful when optimizing scores with respect to the configuration.
In contrast, the gradients of $\frender_k$ and hence $\fconfig$ may not be defined; and even if they are, the embedding $\fconfig$ may be highly oscillatory (see Figure~\ref{fig:noisiness} in the Appendix), making its gradients of questionable utility.

We use the distilled model in three ways: to sample a diverse dataset for retrieval of configurations; to finetune the resulting configurations; and to train STRL and MTRL baselines.

\subsection{Text-to-Goal Generation Methods}
\label{sec:text-to-goal-methods}
Given a text $\txt$, we wish to find an admissible configuration $\bq^*$ that maximizes the configuration-text similarity score:
\begin{align*}
\bq^* \in \argmax_{\bq \in \admiset} \fconfig(\bq) \cdot \ftext(\txt).
\end{align*}
As the similarity score can be highly multimodal and costly to evaluate, we adopt a three-step approach to optimizing it for a given text,
which involves: (i) retrieving high-scoring configurations from a dataset of precomputed configuration embeddings; (ii) starting from those configurations, performing gradient ascent on an approximate version of the similarity score, based on the distilled model of the configuration embedding; and (iii) selecting from the resulting configurations using the exact similarity score, based on the VLM. 
These steps are explained in detail below.
Optionally, one might stop immediately after step (i), returning the single best configuration in the dataset; or after step (ii), returning the best configuration according to the approximate score.

\subsubsection{Retrieving Configurations}
\label{sec:retrieval}
As the configuration embedding $\fconfig$ is independent of the text, one way to mitigate the cost of optimizing the configuration-text similarity is to work with a dataset of configurations $\dataset \coloneqq \{ \bq_1, \bq_2, \dots, \bq_n \} \subset \admiset$ with \emph{precomputed} configuration embeddings $\fconfig(\bq)$ for $\bq \in \dataset$. Given a text $\txt$, one may then retrieve a configuration with a high configuration-text similarity score, simply by taking dot products with those precomputed embeddings:
\begin{equation}
\bq^\text{retrieved} \in \argmax_{\bq \in \dataset} \fconfig(\bq) \cdot \ftext(\txt) .
\label{eq:state_retrieval}
\end{equation}
The effectiveness of this retrieval method hinges on the choice of the retrieval dataset $\dataset$. To ensure high-scoring configurations for \textit{any} text, dataset $\dataset$ must encompass diverse configurations. We therefore propose the following three dataset-generation methods aiming for diversity, while ensuring admissibility of the resulting configurations (full details are given in Appendix~\ref{sec:appendix_data_sampling}):

\paragraph{Random policy dataset.} 
This dataset consists of configurations resulting from a random policy interacting with the environment. 
To encourage diversity, the random policy samples actions uniformly over the action space.

\paragraph{Uniform sampling dataset.}
This dataset is generated by uniformly sampling from the configuration space $\configset$ and projecting them onto admissible configurations $\admiset$. 

\paragraph{Embedding-diversity dataset.}
This dataset is created by focusing on the diversity of configuration embeddings, using the distilled model (\ref{eq:distilled_model}). We optimize a set of configurations by minimizing the maximum cosine similarity between embeddings of distinct configurations with the following loss:
\begin{equation*}
L(\bq_1, \dots, \bq_n) 
= \frac{1}{n} \sum_{i=1}^{n} \max_{j \in [n] \backslash \{i\}}\fdist(\bq_j) \cdot \fdist(\bq_i).
\label{eq:embedding_diversity}
\end{equation*}

\subsubsection{Finetuning Configurations}
\label{sec:finetuning}
Even a huge retrieval dataset may lack configurations aligned with a given text. 
Therefore, we propose a method to finetune retrieved configurations. 
As the exact configuration-text similarity score is costly and may not be differentiable, we attempt to maximize a surrogate $\qtextsurr$ for that score, based on the distilled model~(\ref{eq:distilled_model}): 
\begin{equation}
\qtextsurr(\bq, \txt) \coloneqq \fdist(\bq) \cdot\ftext(\txt).
\label{eq:approx_qt_score}
\end{equation}
By the product rule, the gradient of this surrogate score is
\begin{equation*}
\nabla_{\bq} \qtextsurr(\bq, \txt) =  \nabla_{\bq}\fdist(\bq) \cdot \ftext(\txt).
\end{equation*}
As some configurations may not be admissible, we assume a projection operator $\proj : \configset \to \admiset$ is available, which maps configurations to nearby admissible configurations.
In our experiments, we implement $\proj$ with one step of the MuJoCo physics engine. 
(This step does no dynamics, it just recovers from any interpenetrations.) 
We use this projection to perform projected gradient ascent, with the update
\begin{equation}
\label{eq:modified-gradient}
\bq^{(j+1)} = \proj\left(\bq^{(j)} + \alpha \cdot \nabla_{\bq} \qtextsurr(\bq^{(j)}, \txt) \right),
\end{equation}
where $\alpha$ is the learning rate.
The full procedure is presented in Appendix~\ref{sec:appendix_finetuning}. In practice, both the gradient and projection calculations are highly parallelizable, allowing concurrent optimization of multiple solutions.

\subsection{Zero-Shot LCA}
To craft an LCA, we feed a text $\txt$ to one of the text-to-goal methods described above, resulting in a goal configuration $\bq^\text{goal}_\txt$. Then we feed that goal to a task-agnostic goal-conditioned policy $\pi$, which takes actions distributed as $a \sim \pi(\cdot | s, \bq^\text{goal}_\txt)$ when in state $s$. We train this policy on a collection of goals with GCRL~\citep{Colas2022autotelic, liu2022goal}, using a reward based on the Euclidean distance between the current configuration and a goal configuration $\bq$:
\begin{equation}
R(s, a | \bq) = \| \natproj(s) - \bq \| - \| \natproj(P(s, a)) - \bq   \|
,
\label{eq:gcrl_reward}
\end{equation}
where $\natproj : \states \to \configset$ is the mapping from state to configuration, with configurations seen as Euclidean vectors, and $P$ gives the next state (assuming transitions distributed as $\tilde{P}$ are deterministic). At test time, we simply condition the GCRL agent on a goal configuration to execute the task, without additional training or pre-processing. 

\section{Experiments}
\label{sec:experiments}
In this section, we evaluate our text-to-goal methods and LCAs to address the following questions:
\textbf{(Q1)}~How does the choice of dataset-sampling method affect the performance of VLM-based configuration retrieval? 
\textbf{(Q2)}~What is the impact of using multiple viewpoints to assess configurations in 3D environments?
\textbf{(Q3)}~What is the speed-quality trade-off of the proposed text-to-goal methods?
\textbf{(Q4)}~How do the proposed LCAs compare with STRL and MTRL baselines?

 \subsection{Evaluation Benchmark}

\paragraph{Environment and rendering functions.}
We evaluate our approach on the Humanoid environment from OpenAI's Gym framework~\cite{brockman2016openai}, which we augment with feet and with a cube, enabling stable standing and agent-object interaction.
We create three MuJoCo rendering functions \textit{front view}, \textit{right view} and \textit{left view} (Figure~\ref{fig:multiview_qualitative}) to render configurations of that environment.
Full details are in Appendix~\ref{appendix:env}. 

\paragraph{Tasks.}
We evaluate our work on 256 diverse tasks, which are descriptions of environment configurations in natural language. The list of tasks and the process used to generate it are described in Appendix~\ref{appendix:tasks}. 
The tasks range from simple concepts (\textit{``standing up''}), to compositional relationships (\textit{``sitting on top of a box''}) and abstract concepts (\textit{``standing up in Usain Bolt's celebration pose''}).

 \subsection{Dataset-Sampling Evaluation (Q1)}
\label{sec:state_retrieval_exp}

We sample three datasets of $2.5\times 10^6$ configurations using the three dataset-sampling methods described in Section~\ref{sec:retrieval} and precompute their configuration embeddings~(\ref{eq:confemb}) with the front-view rendering function and the EVA-02-E-14+ VLM~\citep{sun2023eva}. 
Then for all tasks, we retrieve the configuration~(\ref{eq:state_retrieval}) with the highest configuration-text score from each dataset. 

\begin{table}
  \centering
  \setlength{\tabcolsep}{3pt}
  \small 
  \begin{tabular}{@{} l S[table-format=1.4] S[table-format=1.4] S[scientific-notation=true, table-format=1.1e-4] S[table-format=5] @{}}
    \toprule
    {Evaluator Model} & {\thead{Random \\ Policy \\ Dataset}} & {\thead{Uniform \\ Sampling \\ Dataset}} & {\thead{Embedding \\ Diversity \\ Dataset}} \\
    \midrule
    EVA02-E-14+ (LAION2B) & 0.6132 & 0.6128 & \textbf{0.6260}  \\
    EVA02-E-14 (LAION2B)  & 0.3416 & 0.3403 & \textbf{0.3496}  \\
    ViT-H-14 (DFN5B) & 0.4016 & 0.4011 & \textbf{0.4102}   \\
    ViT-H-14 (MetaCLIP) & 0.4114 & 0.4092 & \textbf{0.4197}  \\
    ViT-bigG-14 (DataComp1B) & 0.2700 & 0.2681 & \textbf{0.2771}  \\
    \bottomrule
  \end{tabular}
  \caption{Average single-view configuration-text scores of retrieved configurations by dataset, evaluated with different VLMs. 
  EVA02-E-14(+), DFN5B, MetaCLIP and DataComp1B are from~\citet{sun2023eva}, \citet{xu2023MetaCLIP} and~\citet{gadre2023datacomp} respectively.}
\label{tab:score_other_VLMs}
\end{table}

Table~\ref{tab:score_other_VLMs} presents the configuration-text scores of the retrieved configurations as assessed by various VLMs, averaged over all tasks.
We observe that the configurations retrieved from the embedding-diversity dataset attain the highest scores for all VLMs.
Therefore we use that dataset-sampling method in the remaining experiments.
Further results comparing different dataset sizes and using multiview scores are discussed in Appendix~\ref{sec:appendix_data_sampling}.

\subsection{Multiview Configuration-Text Score (Q2)}
\label{sec:multiview}
We now study the impact of using multiple views to retrieve configurations, as in equation~(\ref{eq:mv_state_text_clip_score}). 

\begin{table}[h]
  \centering
  \setlength{\tabcolsep}{2pt}
  \small 
  \begin{tabular}{@{} l c S[table-format=1.4] S[table-format=1.4] S[table-format=1.4] S[table-format=1.4]}
    \toprule
    {Evaluator Model} & {\thead{Retrieval \\ \# Views}} & {\thead{Front \\ View}} & {\thead{Mid-Left \\ View}} & {\thead{Left \\ View}} \\
    \midrule
    \multirow{2}{*}{EVA02-E-14+ (LAION2B)} & 1 & \textbf{0.6260} & 0.6050 & 0.5835 \\
    & 3 & 0.6147 & \textbf{0.6138} & \textbf{0.6104} \\
    \midrule
    \multirow{2}{*}{EVA02-E-14 (LAION2B)} & 1 & \textbf{0.3496} & 0.3340 & 0.3164  \\
    & 3 & 0.3442 & \textbf{0.3435} & \textbf{0.3386} \\
    \midrule
    \multirow{2}{*}{ViT-H-14 (DFN5B)} & 1 & \textbf{0.4102} & 0.3992 & 0.3816 \\
    & 3  & 0.4092 & \textbf{0.4053} & \textbf{0.3999} \\
    \midrule
   \multirow{2}{*}{ViT-H-14 (MetaCLIP)} & 1 & \textbf{0.4197} & 0.4131 & 0.4021 \\
    & 3  & 0.4194 & \textbf{0.4202} & \textbf{0.4172} \\
    \midrule
    \multirow{2}{*}{ViT-bigG-14 (DataComp1B)} & 1 & \textbf{0.2771} & 0.2661 & 0.2554 \\
    & 3  & 0.2769 & \textbf{0.2734} & \textbf{0.2722} \\
    \bottomrule
  \end{tabular}
  \caption{Single-view configuration-text scores, evaluated by various VLMs, of configurations retrieved using one view and three views from the embedding-diversity dataset, seen from different viewpoints. 
  Appendix~\ref{sec:mv_variant_appendix} gives results for other viewpoints.
  }
  \label{tab:mv_eval}
\end{table}

Table~\ref{tab:mv_eval} compares configurations retrieved from the embedding-diversity dataset using three views (at $0^\circ, \pm 45^\circ$) with configurations retrieved using the front view ($0^\circ$) only. 
These two sets of configurations are evaluated with single-view configuration-text scores for
different viewpoints. 
We include a \emph{mid-left} view at $-22.5^\circ$ to assess generalization to rendering functions distinct from those used for retrieval.
When evaluated on the \emph{front} view, the configurations retrieved using the front view only attain higher scores than those retrieved using three views. 
But when evaluated from \emph{other} views, their scores decrease significantly.
In contrast, the configurations retrieved using three views exhibit little variation in score across the viewpoints tested.

\begin{figure}[h]
\begin{center}
\centerline{\includegraphics[width=.9999\linewidth]{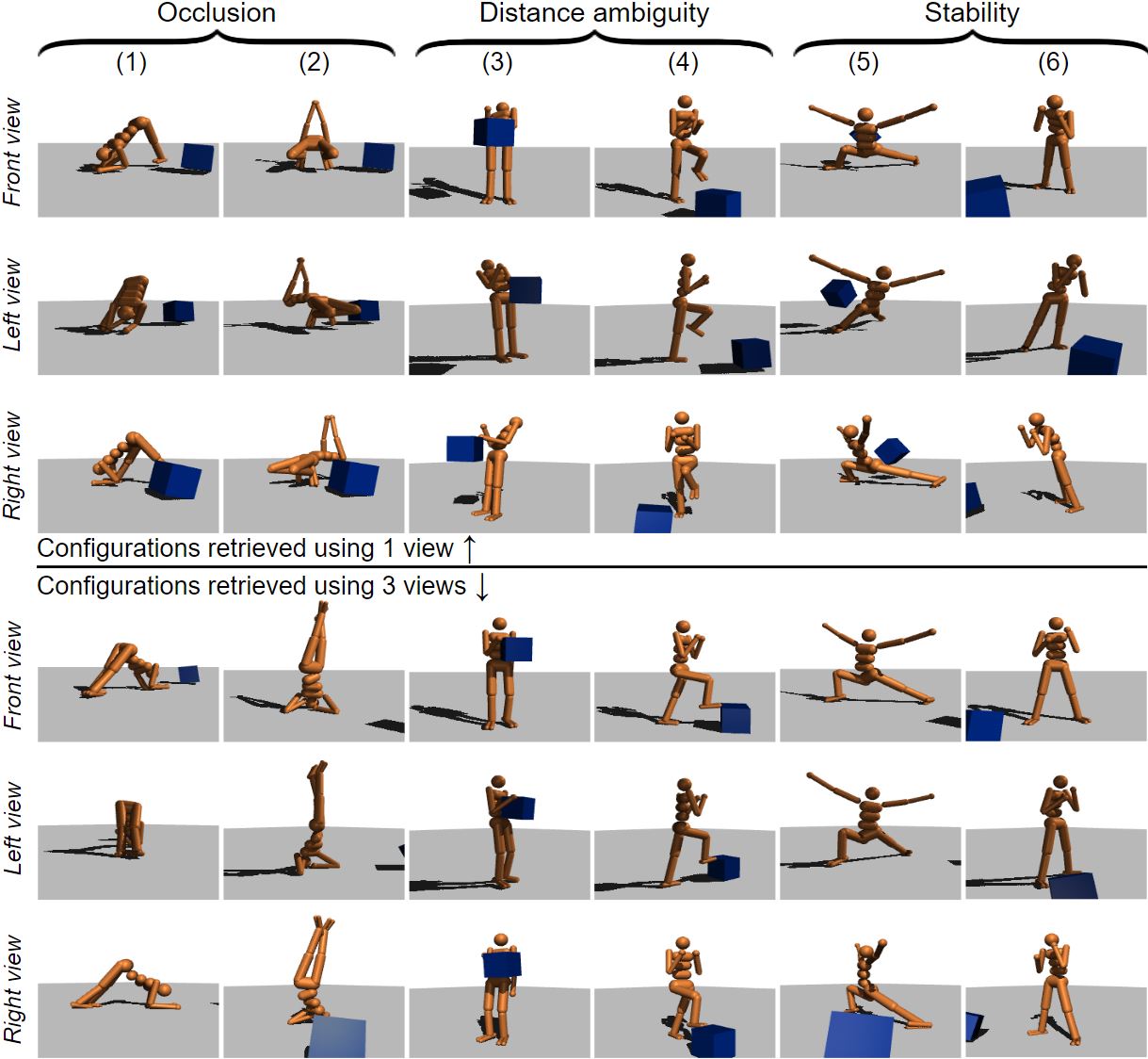}}
\caption{
Each column represents a task: (1) \textit{``downward-facing dog yoga pose''}, (2) \textit{``headstand''}, (3) \textit{``holding a blue box''}, (4) \textit{``box step-up''}, (5) \textit{``warrior yoga pose''} and (6) \textit{``boxing guard''}.
Within each column, the top three rows show the front, left and right views of the best configuration from the embedding-diversity dataset, retrieved using the front view only. The bottom three rows show the best configuration retrieved using the three views. 
}
\label{fig:multiview_qualitative}
\end{center}
\vskip -0.3in
\end{figure}

Figure~\ref{fig:multiview_qualitative} provides some insight into the lack of robustness of single-view retrieval, illustrating six tasks for which a high single-view configuration-text score is misleading.
We identify the following three reasons for these failures: 
\begin{itemize}[leftmargin=*, itemsep=0.2mm]
    \item \emph{Occlusion.} At times, some components hide other components. For instance, in the top three rows of column 1 of Figure~\ref{fig:multiview_qualitative}, the right leg of the humanoid is behind its body. This is not apparent in the front view (top row), but it becomes apparent in the left and right views. Using three views to evaluate configurations mitigates this issue.
    \item \emph{Distance ambiguity.} Based on a single view, the distance between objects in a scene is ambiguous, even for a human. In columns 3 and 4, the humanoid and the cube appear close to each other in the front view (top row), but the side views (second and third rows) reveal that they are actually far apart. 
    \item \emph{Stability.} While the top row of columns 5 and 6 appears to show configurations that correspond well to the specified tasks, the left and right views reveal that the positions are unstable. While our text-to-goal method does not explicitly aim for stability, stability emerges from optimizing the multiview configuration-text score.
\end{itemize}

We conclude that evaluating configurations using multiple views mitigates some of the problems arising from the lack of information of single 2D images. While using multiple views linearly increases the computational cost of precomputing the configuration embeddings, that improvement comes at no extra cost at retrieval time.

\subsection{Configuration Finetuning (Q3, Quality)}
\label{sec:finetuning_exp}
We now study the optimization of retrieved configurations.
First, we use the configurations from the embedding-diversity dataset and their precomputed embeddings to train a distilled model (as detailed in Appendix~\ref{app:distilled}). 
Then for each task, we use that model to finetune the 256 top-scoring configurations (based on the multiview score) from the embedding-diversity dataset (as detailed in Appendix~\ref{sec:appendix_finetuning}). 
We compare two options after finetuning: either we return the configuration with the highest approximate score~(\ref{eq:approx_qt_score}) based on the distilled model; or we evaluate all finetuned configurations with EVA02-E-14+ and select the one with the highest exact multiview score~(\ref{eq:mv_state_text_clip_score}).
The choice of hyperparameters for finetuning is discussed in Appendix~\ref{sec:appendix_finetuning}: one reason for using 256 configurations is that it corresponds to one batch of VLM evaluations.

\begin{table}[h]
  \centering
  \setlength{\tabcolsep}{1.7pt}
  \small 
  \begin{tabular}{@{} l S[table-format=1.4] S[table-format=1.4] S[table-format=1.4]}
    \toprule
    {Evaluator Model} & {\thead{Retrieval \\ Only}} & {\thead{Retrieval+ \\ Finetuning}} & {\thead{Retrieval+ \\ Finetuning+ \\ Selection}}  \\
    \midrule
    EVA02-E-14+ (LAION2B) & 0.6265 & 0.6289 & \textbf{0.6343}  \\
    EVA02-E-14 (LAION2B)  & 0.3506 & 0.3528 & \textbf{0.3564} \\
    ViT-H-14 (DFN5B) & 0.4187 & 0.4224 & \textbf{0.4229}  \\
    ViT-H-14 (MetaCLIP) & 0.4226 & 0.4243 & \textbf{0.4255} \\
    ViT-bigG-14 (DataComp1B) & 0.2771 & 0.2795 & \textbf{0.2800} \\
    \bottomrule
  \end{tabular}
  \caption{Multiview configuration-text scores of retrieved configurations, finetuned configurations, and finetuned configurations selected by EVA02-E-14+. These scores are evaluated by various VLMs and averaged over the 256 tasks.}
  \label{tab:finetuning_3v}
\end{table}

\begin{figure}[h]
\vskip -0.02in
\begin{center}
\centerline{\includegraphics[width=.9999\linewidth]{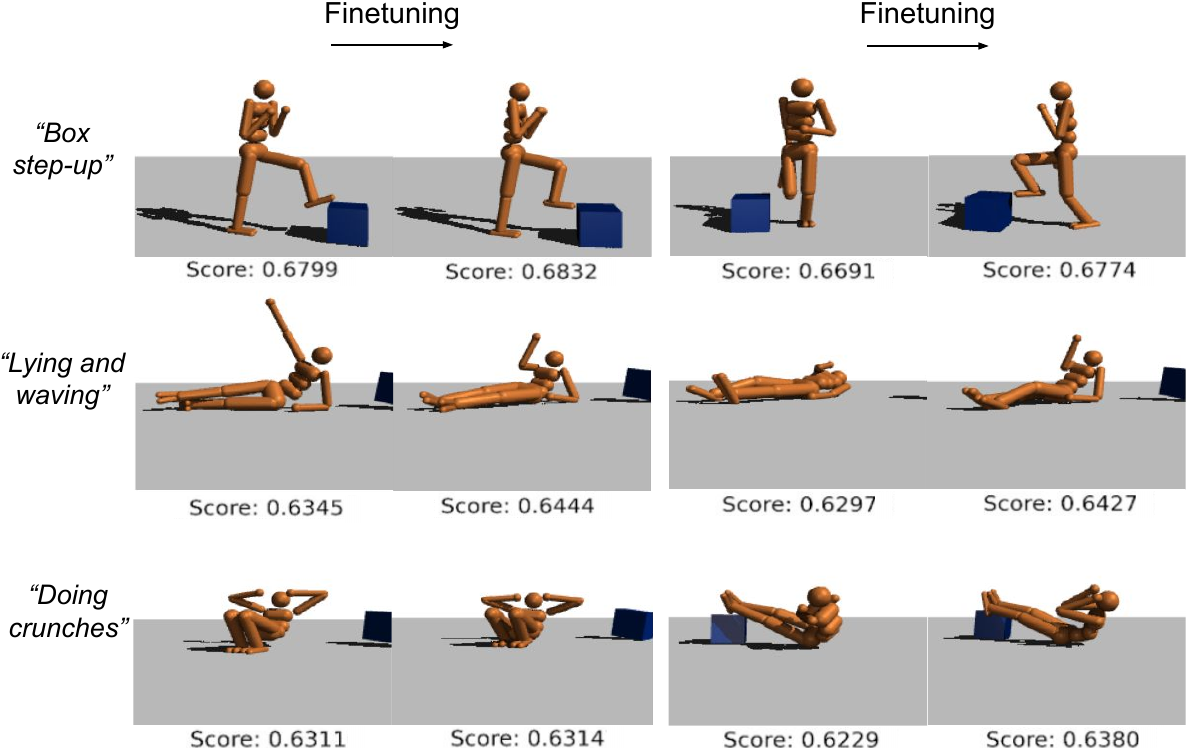}}
\caption{
Each row shows a single task. Columns 1 and 3 show the top-1 and another of the top-20 retrieved configurations; columns~2 and~4 show their finetuned counterparts (without VLM selection).}
\label{fig:finetuning_qualitative}
\end{center}
\vskip -0.2in
\end{figure}

Table~\ref{tab:finetuning_3v} presents results of finetuning, indicating that the proposed finetuning methods further improve the configuration-text alignment of retrieved configurations, even when evaluated by other VLMs. 
Figure~\ref{fig:finetuning_qualitative} illustrates retrieved configurations and their finetuned counterparts, showing that even small numerical increases in the configuration-text score can correspond to significant changes in configuration.
Finetuned configurations are sometimes more convincing (and hardly ever less convincing) to a human viewer.

Finally,  Appendix~\ref{sec:appendix_finetuning} studies the effect of the number of finetuning steps. 

\subsection{Time Efficiency (Q3, Speed)}
\label{sec:time-efficiency}

Evaluating 256 three-view configuration embeddings with our distilled model takes $4.8\times 10^{-4}$ s, whereas computing those embeddings with EVA02-E-14+ takes 21.1 s (for the computing infrastructure described in Appendix~\ref{appendix:architectures-and-ppo}). Thus, the distilled model reduces the computation time of configuration embeddings by $40\, 000\times$. 

The timings of our text-to-goal methods are as follows: embedding the text takes 0.013~s, retrieving high-scoring configurations with precomputed embeddings (by brute force, from a dataset of $2.5\times 10^6$ samples) takes 0.015~s; and the optional stages of finetuning and selecting the highest-scoring configuration (by applying EVA02-E-14+ to three views, corresponding to three batches) take 2.8~s and 21.1~s respectively.\footnote{Our current implementation is based on a CPU version of MuJoCo: finetuning may be much faster with a GPU version. Selecting the highest-scoring configurations could also be much faster: for instance, one evaluation of EVA02-E-14+ requires 2362B FLOPs, whereas SigLIP~\citep{zhai2023sigmoid} requires 233B.}   

This compares favourably with approaches based on goal-image generation~\cite{gao2023pretrained}, which would require $\ge 3\times 20$~s were they to generate three goal images to overcome the ambiguities of 2D images.
It also compares favourably with using LLMs to generate source code specifying goals. For instance,  LLAMA-2-70B~\citep{touvron2023llama} generates about 10 tokens per second on our infrastructure, and requires about 15~s to generate a Humanoid configuration.
In conclusion, our text-to-goal approaches are fast enough for real-time language-guided control of humanoid agents.

\subsection{Language-Conditioned Agents (Q4)}
We now detail our baselines and the GCRL policy. Then we compare the resulting LCAs. 

\paragraph{Observation and action spaces.} 
Each of the LCAs receives a 343-dimensional observation vector. This includes the current position, orientation and velocity of each component of the Humanoid and the cube, along with the timestep (as detailed in Appendix~\ref{appendix:env}). For MTRL, this observation is augmented with the text embedding of the task. For GCRL, it is augmented with the goal configuration. The agents take 19-dimensional actions, corresponding to actuator torques.

\paragraph{MTRL and STRL baselines.}
Previous works~\citep{mahmoudieh2022zero, rocamonde2023vision, fan2022minedojo, adeniji2023language} use the single-view configuration-text score to define the reward function
\begin{equation}
R^{\text{past work}}_{\txt}(s, a) = \qtextsim(\natproj(s), \txt)
\label{eq:reward_exact_past_work}
\end{equation}
for action $a$, in state $s$, with text $\txt$. We propose three improvements. First, we use three-view configuration-text scores, which improve robustness as shown in Section 4.3. Second, we define the \emph{VLM reward} as the time difference of configuration-text scores:
\begin{equation}
R_{\txt}(s, a) = \qtextsim(\natproj(P(s, a)), \txt) - \qtextsim(\natproj(s), \txt).
\label{eq:mtrl_reward_exact}
\end{equation}
This reward encourages policies to attain high final scores, rather than to maintain high average scores as discussed in Appendix~\ref{sec:appendix_lca_eval}. Finally, we 
define the \emph{approximate VLM reward}, using the  distilled model (the same one that we used for finetuning):
\begin{equation}
\hat{R}_{\txt}(s, a) = \qtextsurr(\natproj(P(s, a)), \txt) - \qtextsurr(\natproj(s), \txt).
\label{eq:mtrl_reward}
\end{equation}
This reward is up to $40\, 000\times$ faster to evaluate and less oscillatory than $R_{\txt}$. 

Unlike previous work, which only trains STRL agents, we also train MTRL agents.
Both STRL and MTRL agents use reward $\hat{R}_{\txt}$.
The MTRL agents learn a multi-task policy $\pi(a | s, \ftext(\txt))$, which is conditioned on the VLM text-embedding. 
This enables zero-shot execution of new tasks, so it is directly comparable with the proposed LCA based on text-to-goal generation. 
To compare the performance of MTRL on test and training tasks, 
we randomly partition the 256 tasks into two sets of 128, and train one MTRL agent on each set.

\paragraph{GCRL agent.} We train the GCRL agent to maximize reward~(\ref{eq:gcrl_reward}) using goals sampled uniformly from the embedding-diversity dataset. At test time, we condition the agent on configurations retrieved and potentially finetuned using the methods of Section~\ref{sec:text-to-goal-methods}. 

\paragraph{Architectures and training.} The MTRL, STRL and GCRL agents have identical architectures (as far as possible given their different inputs). All are trained with proximal policy optimization~\cite{schulman2017proximal} with similar hyperparameters. Full details are given in Appendix~\ref{appendix:architectures-and-ppo}.

\begin{figure}[h]
\begin{center}
\centerline{\includegraphics[width=.9999\linewidth]{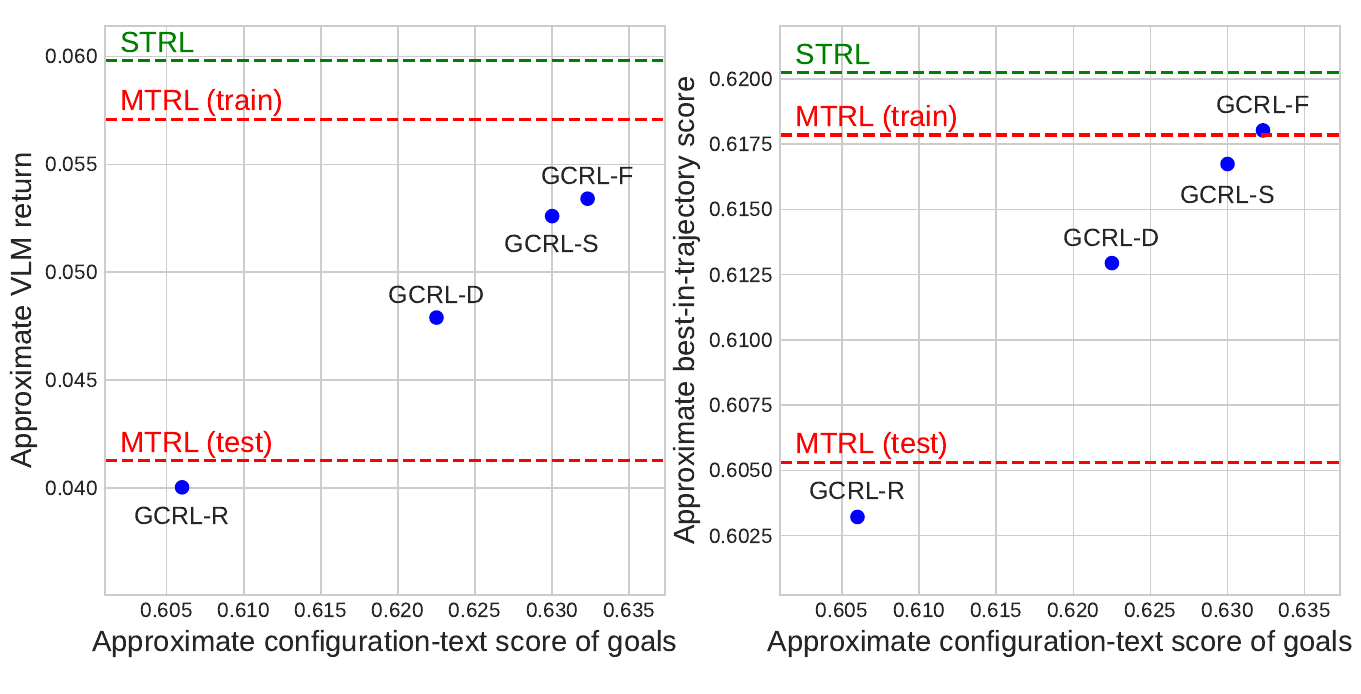}}
\caption{Approximate VLM returns and approximate best-in-trajectory configuration-text scores for the STRL and MTRL baselines, and for the GCRL agent given four types of goal configuration: retrieval from the random policy dataset (\emph{GCRL-R}); and retrieval from the diversity dataset (\emph{GCRL-D}), plus finetuning (\emph{GCRL-F}), plus selection based on the exact score (\emph{GCRL-S}). \emph{MTRL (train)} and \emph{MTRL (test)} are for the baseline models evaluated on their training and test tasks respectively. The metrics are averaged over all 256 tasks, with 10 episodes per task.}
\label{fig:LCA_eval_approx}
\end{center}
\vskip -0.3in
\end{figure}

\paragraph{Comparison.} 
Figure~\ref{fig:LCA_eval_approx} compares the approximate VLM returns, given by discounted sums of reward~(\ref{eq:mtrl_reward}), and the approximate best-in-trajectory configuration-text scores 
$$\max \{\qtextsurr(\natproj(s), \txt) : s\in \text{trajectory} \}$$ 
for the STRL and MTRL baselines, and for the GCRL agent given four different types of goal. 
Here we choose to show the approximate metrics because the baselines were trained to optimize the approximate VLM return.
Even though the GCRL agents were not trained to optimize the VLM return, GCRL-D, GCRL-F and GCRL-S (acronyms defined in the caption) all attain higher approximate VLM returns than MTRL (test). 

This demonstrates that LCAs based on a decomposition into goal-generation and goal-attainment can generalize more effectively to previously unseen tasks than MTRL agents. 
Moreover, the STRL agents achieve an average approximate best-in-trajectory score of $0.6202$, whereas the finetuned goal configurations have an average score of $0.6324$. Thus, if the GCRL agent were able to accurately reach its goals, then it would outperform the STRL agents on that metric.

Appendix~\ref{sec:appendix_lca_eval} presents analogous results using \emph{exact} configuration-text scores, which do not differ greatly differ from those for approximate scores. It also presents results for each task individually. In summary, the exact and approximate VLM returns of GCRL-F exceed those of MTRL (test) in 198 and 203 of the 256 tasks respectively; while the exact VLM return of GCRL-S exceeds that of MTRL (test) in 205 tasks.

\begin{figure}[h]
\begin{center}
\centerline{\includegraphics[width=.9999\linewidth]{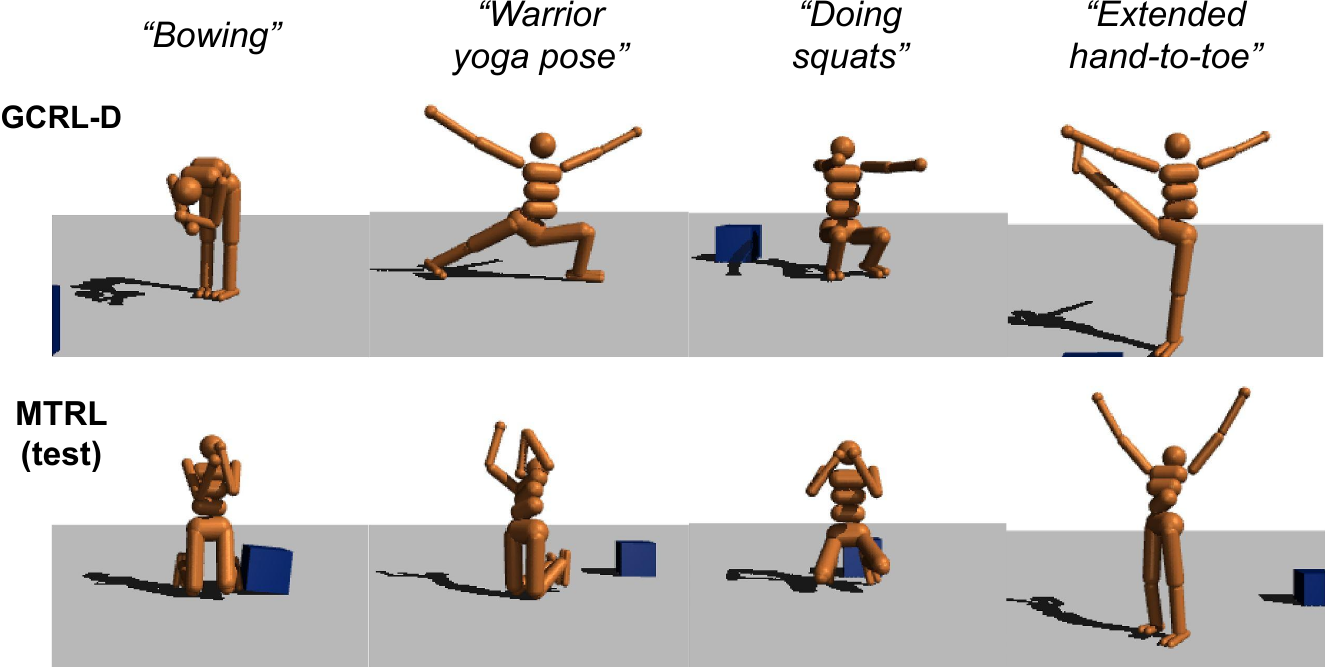}}
\caption{Front-view of the best-in-trajectory configuration (with respect to the approximate VLM score) reached by GCRL-D and by the MTRL (test) baseline in a single rollout. Analogous figures for all LCAs on all 256 tasks are in Appendix~\ref{sec:reached-configs}.}
\label{fig:LCA_best_in_traj_main}
\end{center}
\vskip -0.3in
\end{figure}

\paragraph{Generalization.}
As illustrated in Figure~\ref{fig:LCA_best_in_traj_main}, the generalization of GCRL-D to new tasks is often superior to that of MTRL (test).
We propose the following two hypotheses explaining this observation:
\begin{enumerate}
    \item The training set of GCRL is more helpful than that of MTRL: it is larger and it is optimized for diversity.
    It is larger as training tasks for GCRL only consist of configurations to reach, and these are simpler to collect than the texts required for MTRL training tasks. 
    Optimizing for diversity is clearly beneficial: indeed, Figure~\ref{fig:LCA_eval_approx} shows that GCRL-R (random policy) does worse than MTRL-test, whereas GCRL-D (diversity) does better.
    \item GCRL training tasks are easier to learn than those of MTRL. 
    Indeed, the GCRL agent only needs to learn how to reach already-found high-scoring configurations, whereas the MTRL agent must both find high-scoring configurations and learn to reach them. 
    Moreover, Appendix~\ref{app:compositionality} demonstrates that the VLM score is highly oscillatory, which could perturb RL training. 
    While Appendix~\ref{app:distilled} shows that the approximate VLM score used for MTRL is smoother than the true VLM score, it is still much more oscillatory than the Euclidean distance used for GCRL.
\end{enumerate}
These advantages come with a trade-off. 
While MTRL directly optimizes the approximate VLM return, GCRL uses configurations as a proxy to maximize that return. However, some configurations may be hard-to-reach, and proximity to a goal configuration in terms of Euclidean distance does not necessarily translate to a high approximate VLM score, as discussed in Appendix~\ref{app:failures}.

\section{Limitations and Further Work}
\label{sec:further-work}
We would like to extend the range of tasks that may be performed to include \emph{compositional relationships} between a diverse set of objects. As the current generation of VLMs is limited in its ability to assess such relationships, as discussed in Appendix~\ref{app:compositionality}, we await a new generation of VLMs to enable such work.  
Also, our current approach is limited to static configurations: in future, it will be interesting to extend the approach to
\emph{dynamic tasks}.

\section*{Acknowledgements}
We thank Quan Sun \textit{et al.} as well as the OpenCLIP community for releasing pretrained state-of-the-art VLMs, enabling such research. We also thank Jean-Michel Renders, Tomi Silander, Diane Larlus and David Emukpere and the reviewers for their many helpful comments.

\section*{Impact Statement}

\textbf{Positive impacts.} The development of language-conditioned agents (LCAs) will further fluidify human-technology interaction. Given the convenience of language as a medium for humans to express desires and commands, this will likely result in an increase in such interactions. The possibility of steering technologies easily will empower every potential user, including those without technical expertise. LCAs are particularly promising as they might enhance the independence of individuals with mobility limitations, such as elderly or paraplegic persons. They may also be valuable for fine-grained control of robots in hazardous environments, such as radioactive or extraterrestrial terrains.

\textbf{Negative impacts.} The proposed technology can be used to enable language-conditioned design and interaction with virtual environments, such as games, potentially making those environments more addictive, possibly causing users to become dependent upon them, lose touch with reality and become isolated. Additionally, a person of malicious intent could specify tasks that might be harmful (e.g., kick something hard) or insulting (e.g., make some kind of insulting gesture) to the LCA, and use it to generate corresponding images or to control a physical robot. As with any work that might be embodied in a physical robot, there is a risk of physical harm to objects and entities in that robot’s immediate vicinity: the training of the RL agents would need to be adapted to take safety factors into account; the agents would need to be trained/finetuned on real-world data; and operators would need to be trained on appropriate safety guidelines.

\bibliography{icml-2024}
\bibliographystyle{icml2024}

\newpage
\appendix
\onecolumn

\section{Environment, Cameras and Configurations}
\label{appendix:env}
In this section, we first provide details about the Humanoid-plus-cube environment, then we describe the camera settings used to render that environment, and finally we detail each dimension of the environment's configuration space.

\subsection{Environment Details}
We use the Humanoid environment from OpenAI's Gym framework~\cite{brockman2016openai}, originally designed for locomotion, which we modify as follows:

\begin{itemize}[leftmargin=*, itemsep=0.8mm, parsep=0mm, topsep=1mm, partopsep=0mm]
    \item The original Humanoid does not have feet. This is problematic, as we expect the agent to balance in diverse poses. 
    We therefore augment the XML definition of the Humanoid model with feet, borrowed from the Humanoid model in the DeepMind Control Suite~\cite{tassa2018deepmind}. This modification adds two dimensions to the action space (for ankle control) and four dimensions to the state space (ankle-foot angles and angular velocities).
    \item We add a cube to the environment, with sides of length 0.15 m. This allows agent-object interactions, other than agent-floor interactions. This adds 14 dimensions to the state space (7 dimensions for the cube position and orientation and 7 for the corresponding velocities). The cube is modelled as a homogeneous volume with mass 0.5 kg.  
    \item Episodes terminate after 100 steps. We therefore include the episode time-step in the state. To do otherwise would negatively impact policy performance~\cite{pardo2018time}.
    \item We delete some of the original 376 state dimensions, as they are not relevant. Notably, the 84-dimensional vector of external forces is \emph{always} zero when using MuJoCo $>$ 2.0.\footnote{This is explained in the environment documentation at \url{https://www.gymlibrary.dev/environments/mujoco/humanoid/}.}
\end{itemize}

In summary, this results in a Humanoid-plus-cube environment, with action space $[-1, 1]^{19}$ and an observation space of 343 dimensions. 

\paragraph{Initial state distribution.} 
The original Humanoid environment specifies a \emph{default state}, in which the Humanoid is standing up, facing forward, with all velocities equal to zero. We extend this default state by setting the ankle angles such that the feet are at $6.5^\circ$ to the floor (toes pointing up), setting the Cartesian coordinates of the cube to be $(-0.15, 0.4, 0.15)$, and setting its orientation to be the quarternion $(0, 0, 0, 1)$, so that its sides are parallel to the axes. All corresponding velocities are zero. 
As in the original Humanoid environment, we sample the initial state of each episode by adding uniform noise on $[-0.01, 0.01]$ independently to all positions and velocities of this default state.

\subsection{Camera Settings}

\begin{figure}[H]
\begin{center}
\centerline{\includegraphics[width=.6\linewidth]{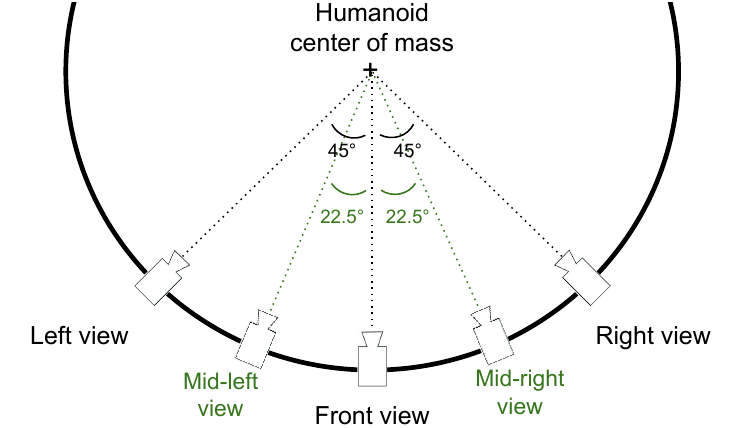}}
\caption{Positions and orientations of the different cameras with respect to the Humanoid center of mass.}
\label{fig:renderers}
\end{center}
\vskip -0.2in
\end{figure}

Rendering is performed using MuJoCo rendering functions, which we control by defining cameras in the XML file describing the environment. Our rendering functions share most of their settings: directional lightening with shadow casting, default MuJoCo camera field of view, a wood-colored Humanoid with color \verb|#CC9966|, a grey floor with color \verb|#B8B8B8|, and a blue cube with color \verb|#0033FF|. Each camera generates RGB images of size $224\times224\times3$, which is the default image resolution of most VLM image encoders. The cameras all point toward and follow the Humanoid's center of mass (COM). They are placed 2.65 m away from the Humanoid COM: this ensures that the cube and the whole body of the Humanoid are visible in the rendered images, for most configurations. The camera heights are initialized 35 centimeters above the Humanoid COM: this ensures that the upper surface of the Humanoid is partly visible, even when the Humanoid is lying on the floor. 

As illustrated in Figure~\ref{fig:renderers}, the cameras differ in their orientations and positions. The front camera is first placed facing the Humanoid in its default state. The left, mid-left, mid-right and right cameras are equally distant from the Humanoid COM, but are rotated by $-45^\circ$, $-22.5^\circ$, $22.5^\circ$ and $45^\circ$ relative to the front camera respectively, about a vertical axis. The left, right and front views are used to compute the multiview configuration-text score, while the mid-left and mid-right views are only used in the robustness-to-viewpoint evaluations of Section~\ref{sec:multiview} and Appendix~\ref{sec:mv_variant_appendix}.

Clearly, VLM scores depend upon the choice of colours and textures, employed while rendering. There is thus ample room for future research to explore alternative rendering functions, possibly using image-to-image~\citep{zhang2023adding} or video-to-video realism enhancement models~\cite{zeltner2023real, chu2023video} on top of rendered images.

\subsection{Configurations}
In our modified Humanoid environment, the configuration space $\configset$ has 31 dimensions, as listed in Table~\ref{tab:configurations}. These encompass the torso height, the relative positions of its joints, the Cartesian coordinates of the cube, and a quaternion giving the cube's orientation.

\begin{table}[H]
\centering
\begin{tabular}{|c|l|}
\hline
\thead{\textbf{Configuration} \\ \textbf{Dimension}}  & \textbf{Description} \\
\hline
0 & z-coordinate of the torso (centre)   \\                      
1 & x-orientation of the torso (centre)  \\                      
2 & y-orientation of the torso (centre)   \\                     
3 & z-orientation of the torso (centre)   \\                     
4 & w-orientation of the torso (centre)   \\                     
5 & z-angle of the abdomen (in lower waist)  \\                 
6 & y-angle of the abdomen (in lower waist)   \\                
7 & x-angle of the abdomen (in pelvis)   \\                      
8 & x-coordinate of angle between pelvis and right hip (in right thigh) \\
9 & z-coordinate of angle between pelvis and right hip (in right thigh) \\
10 & y-coordinate of angle between pelvis and right hip (in right thigh) \\
11 & angle between right hip and the right shin (in right knee) \\
12 & x-coordinate of angle between pelvis and left hip (in left thigh) \\
13 & z-coordinate of angle between pelvis and left hip (in left thigh) \\
14 & y-coordinate of angle between pelvis and left hip (in left thigh) \\
15 & angle between left hip and the left shin (in left knee) \\
16 & coordinate-1 (multi-axis) angle between torso and right arm (in right upper arm) \\
17 & coordinate-2 (multi-axis) angle between torso and right arm (in right upper arm) \\
18 & angle between right upper arm and right lower arm (in right elbow) \\
19 & coordinate-1 (multi-axis) angle between torso and left arm (in left upper arm) \\
20 & coordinate-2 (multi-axis) angle between torso and left arm (in left upper arm) \\
21 & angle between left upper arm and left lower arm (in left elbow) \\
22 & angle between left shin and left ankle \\
23 & angle between right shin and right ankle \\
24 & x-coordinate of the cube with respect to the torso x-coordinate \\
25 & y-coordinate of the cube with respect to the torso y-coordinate \\
26 & z-coordinate of the cube \\
27 & x-orientation of the cube \\
28 & y-orientation of the cube \\
29 & z-orientation of the cube \\
30 & w-orientation of the cube \\
\hline
\end{tabular}
\caption{Description of each dimension of the configuration space of the Humanoid-plus-cube environment. (The terminology is that of the original Humanoid documentation.)}
\label{tab:configurations}
\end{table}

\clearpage

\section{Dataset Sampling and Retrieval}
\label{sec:appendix_data_sampling}
This section gives a detailed explanation of the configuration-dataset sampling methods introduced in Section~\ref{sec:retrieval} of the main paper. Then it discusses the effect of dataset size on the quality of the retrieved configurations, and on the time and memory required for retrieval. 

\subsection{Random Policy Dataset}
This dataset is constructed by collecting all configurations encountered by a random policy. Episodes start in a state sampled from the initial state distribution detailed in Section~\ref{appendix:env}, with one exception: we sampled the initial height of the cube uniformly on the interval $[0.15, 1]$, as such configurations are relevant but unlikely to be encountered without this change. The policy then performs actions that are uniformly sampled in the action space at each time step.

\subsection{Uniform Sampling Dataset}

This dataset is constructed by uniformly sampling the configuration space $\configset$, whose components were described in  Table~\ref{tab:configurations}. 
We uniformly and independently sample each component of the configuration within the following bounds:

\begin{itemize}[leftmargin=*, itemsep=0.8mm, parsep=0mm, topsep=1mm, partopsep=0mm]
    \item Configuration dimension 0 represents the height of the Humanoid's torso, which is not bounded in the Humanoid XML file. Instead, we sample this component from the interval $[0.1, 1.45]$. A height of 0.1 m is the lowest the torso can be without penetrating the floor. If the Humanoid's torso is at a height of $1.45$ m, and it is standing up straight, then its feet are 0.2 m above the floor.
    \item Configuration dimensions 1--4 and 27--30 are quaternions. Therefore we sample them from the interval $[-1, 1]$.
    \item For configuration dimensions 5--23, which represent relative joint angles, we use the bounds specified in the Humanoid XML file.
    \item Configuration dimensions 24 and 25 represent the x and y components of the cube's Cartesian coordinates relative to the Humanoid's torso. We sample these from the interval $[-1, 1]$. This ensures that the cube will be visible in most rendered images. 
    \item Configuration dimension 26 is the height of the cube, which is unbounded. We sample this from the interval $[0.145, 1]$ to avoid cube-floor interpenetration.\footnote{Retrospectively, the upper limit of 1 m on the cube height is rather low for some tasks, such as task 121, \textit{``pretending to lift a heavy weight overhead, like a weightlifter''}. However, it is appropriate for most tasks and is not the cause for the inferiority of the uniform sampling method relative to the embedding-diversity method.} 
\end{itemize}

Sampled configurations may not be admissible, we therefore project them to nearby admissible configurations with $\proj$ (the operator defined in Section~\ref{sec:finetuning}). 
Because of this projection, the method does not truly result in uniform samples. However,  configurations involving contact are obviously important (for instance, standing with feet in contact with the floor, or arms holding a cube) and truly uniform sampling would typically have zero probability of generating such samples. 

The uniform sampling procedure generates millions of configurations in under an hour, with most of the time spent projecting configurations. 
However, computing the multiview embeddings of $2.5 \times 10^{6}$ sampled configurations takes about 58 hours.

\subsection{Embedding-Diversity Dataset}

The process used to construct the embedding-diversity dataset was as follows:
\begin{enumerate}[topsep=0pt,itemsep=1pt,parsep=0pt]
\item Initialize the dataset with $500\, 000$ configurations sampled with the random policy.
\item Compute their exact configuration embeddings~(\ref{eq:confemb}) using the rendering functions and the VLM.
\item While the dataset has not reached the required size:
\begin{enumerate}[topsep=0pt,itemsep=1pt,parsep=0pt]
\item Train a distilled model using the current dataset (as detailed in Appendix~\ref{app:distilled}).
\item Sample $n = 256\, 000$ configurations uniformly without replacement from the current dataset.
\item Minimize the maximum pairwise embedding similarity of these configurations by performing $k=100$ steps of projected gradient descent on the loss given in Section~\ref{eq:embedding_diversity}. (We only optimize $n=256\, 000$ per batch, due to the $O(n^2)$ cost of computing this loss.) After each gradient step, project onto nearby admissible configurations using the operator $\proj$. 
\item Compute the exact configuration embeddings~(\ref{eq:confemb}) of the optimized configurations. 
\item Include these configurations in the configuration dataset.  
\end{enumerate}
\end{enumerate}
This method takes about 64 hours to generate a dataset of $2.5\times 10^6$ configurations with their corresponding multiview configuration embeddings. 
Even though the process involves repeated training of a distilled model, as well as the optimization of an objective involving many evaluations of the distilled model, most of the run time is spent computing the exact configuration embeddings.  
This is because computing the exact multiview configuration embeddings is $40\,000\times$ slower than evaluating the distilled model, 
as discussed in Section~\ref{sec:time-efficiency}.

\subsection{Effect of Dataset Size}

\begin{figure}[h]
\begin{center}
\centerline{
\includegraphics[width=.5\linewidth]{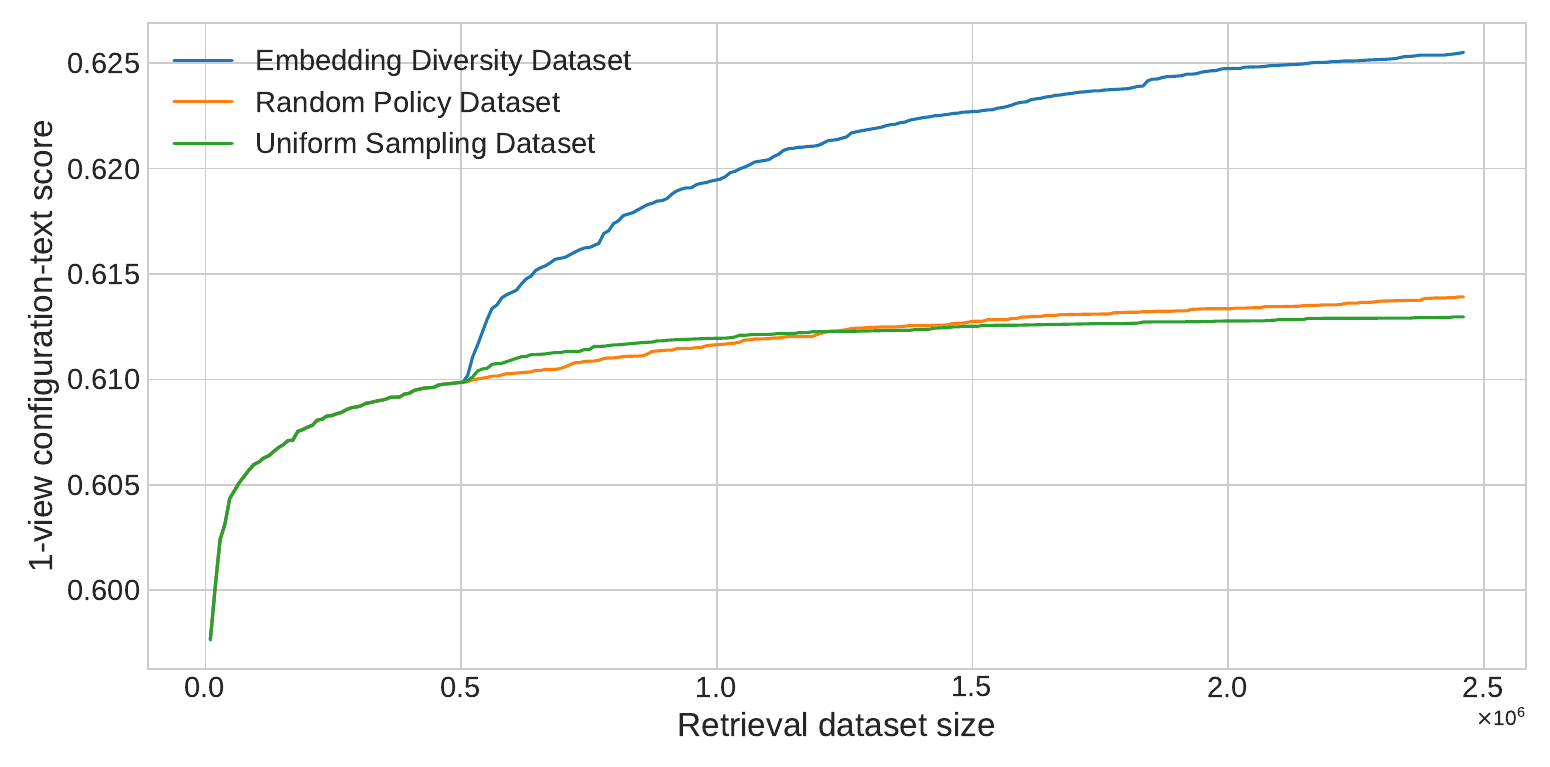}
\includegraphics[width=.5\linewidth]{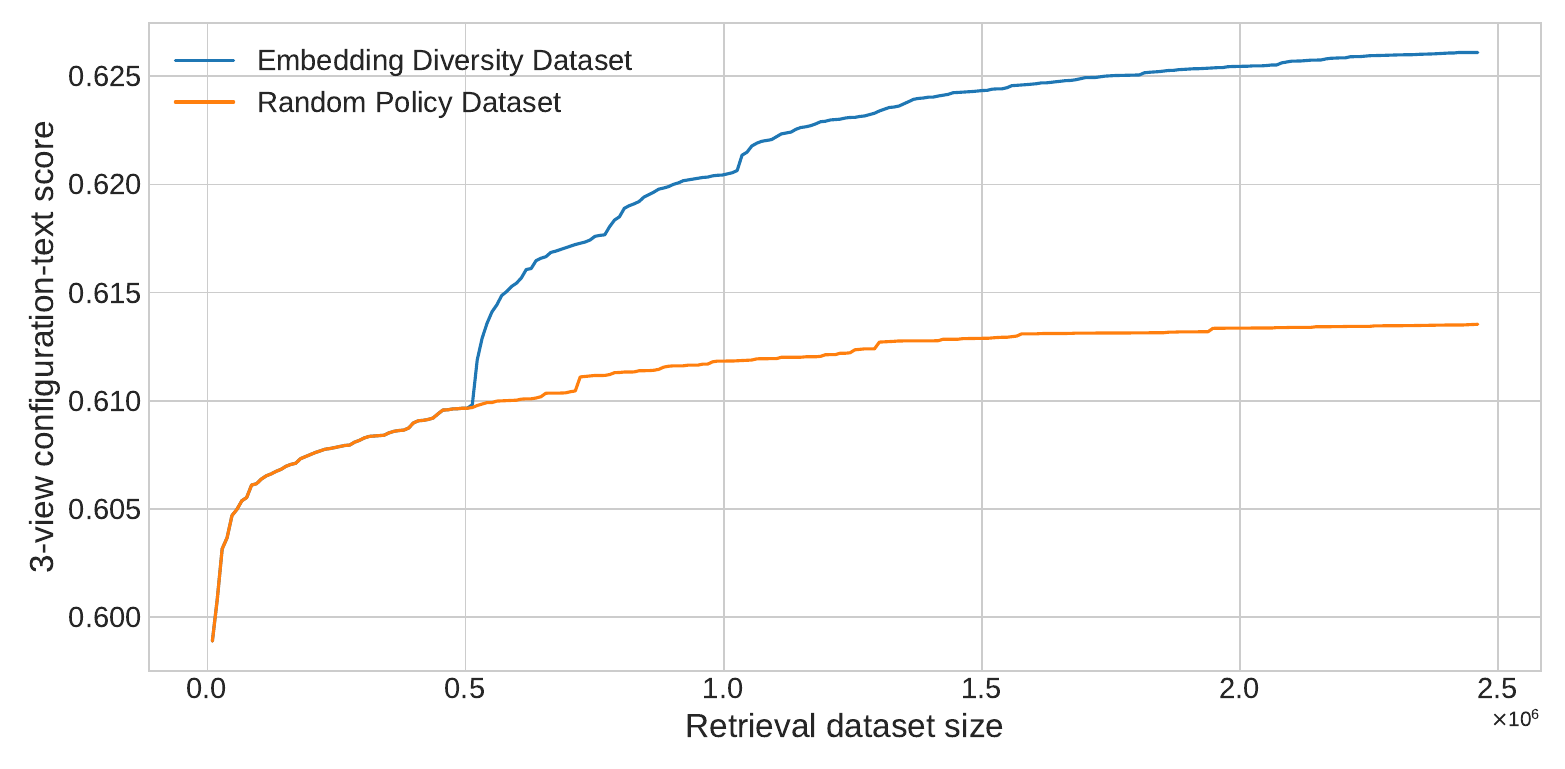}
}
\caption{Single-view (left) and multiview (right) configuration-text scores evaluated by EVA02-E-14+ on the best configuration retrieved from each of the three types of dataset, with respect to the dataset size, averaged over the 256 texts. (Results for uniform sampling with multiview scores are not shown due to a lack of time.)}
\label{fig:score_vs_size}
\end{center}
\end{figure}

Figure~\ref{fig:score_vs_size} shows the average configuration-text scores, for one- and three-view versions of those scores, as a function of the dataset size.
Although the embedding-diversity dataset-generation process extends the dataset in batches of size $500\, 000$ or $256\, 000$, this plot shows scores for intermediate dataset sizes, generated by randomly sampling an appropriate number of configurations from each new batch.
The plots show that an embedding-diversity dataset of size $600\, 000$ outperforms a random policy dataset that is $4\times$ larger. 
The behaviour of the uniform sampling dataset is similar to that of the random policy dataset for single-view scores. 

The shape of the curves for the embedding-diversity method can be explained as follows. For datasets smaller than $500\, 000$ samples, the embedding diversity samples are all generated by the random policy. Therefore, the curves for random policy and embedding diversity coincide. But at $500\, 000$ samples, there is a cusp in the embedding diversity curve, as new samples now result from minimizing pairwise configuration-embedding similarities. There is a similar cusp every $n = 256\, 000$ samples thereafter, corresponding to the retraining of the distilled model and the optimization of a new set of configurations, but those cusps are less visible as there is an overall diminishing return on investment.

\paragraph{Retrieval time and memory.} The cost of retrieval of the highest-scoring configuration from a configuration dataset, using na\"{i}ve brute-force evaluation, increases linearly with the size of that dataset, as long as the collection of precomputed embeddings for that dataset remains small enough to fit in the GPU memory. With an NVIDIA RTX A6000 GPU, the time to retrieve from a dataset of size $2.5 \times 10^6$ is 15 milliseconds. With 16-bit floating-point embeddings of dimension 1024, this dataset requires 4.77 GB of VRAM, whereas the GPU has 48 GB of VRAM. Were the retrieval time to become a bottleneck, fast nearest neighbour methods could be employed~\citep{wang2021comprehensive}.

\clearpage

\section{Multiview Scores}
\label{sec:mv_variant_appendix}



Table~\ref{tab:mv_eval_full} compares configurations retrieved using front-view configuration-text scores with configurations retrieved using multiview configuration-text scores from the embedding-diversity dataset, for the 256 tasks. These two sets of configurations are evaluated with single-view configuration-text scores from five different viewpoints, extending Table~\ref{tab:mv_eval} of the main paper to include two additional viewpoints: mid-right and right views. 
As would be expected, the scores for right and mid-right views are very similar to those for left and mid-left views respectively. 
The scores of configurations retrieved using the front-view tail off more rapidly with viewing angle than those retrieved using multiview.
Therefore multiview configuration evaluation leads to configurations that are more robust to changes in the viewpoints used by the rendering functions.

\begin{table}[h]
  \centering
  \setlength{\tabcolsep}{3pt}
  \small 
  \begin{tabular}{@{} l c S[table-format=1.4] S[table-format=1.4] S[table-format=1.4] S[table-format=1.4] S[table-format=1.4]}
    \toprule
    {Evaluator Model} & {\thead{Retrieval \\ \# Views}} & {\thead{Front \\ view}} & {\thead{Left \\ view}} & {\thead{Right \\ view}} & {\thead{Mid-left \\ view} }& {\thead{Mid-right \\ view}} \\
    \midrule
    \multirow{2}{*}{EVA02-E-14+ (LAION2B)} & 1 & \textbf{0.6260} & 0.5835 & 0.5786 & 0.6050 & 0.6030 \\
    & 3  & 0.6147 & \textbf{0.6104} & \textbf{0.6123} & \textbf{0.6138} & \textbf{0.6143} \\
    \midrule
    \multirow{2}{*}{EVA02-E-14 (LAION2B)} & 1 & \textbf{0.3496} & 0.3164 & 0.3120 & 0.3340 & 0.3330 \\
    & 3  & 0.3442 & \textbf{0.3386} & \textbf{0.3423} & \textbf{0.3435} & \textbf{0.3445} \\
    \midrule
    \multirow{2}{*}{ViT-H-14 (DFN5B))} & 1 & \textbf{0.4102} & 0.3816 & 0.3813 & 0.3992 & 0.4006 \\
    & 3  & 0.4092 & \textbf{0.3999} & \textbf{0.4043} & \textbf{0.4053} & \textbf{0.4082} \\
    \midrule
    \multirow{2}{*}{ViT-H-14 (MetaCLIP))} & 1 & \textbf{0.4197} & 0.4021 & 0.3967 & 0.4131 & 0.4121 \\
    & 3  & 0.4194 & \textbf{0.4172} & \textbf{0.4106} & \textbf{0.4202} & \textbf{0.4202} \\
    \midrule
    \multirow{2}{*}{ViT-bigG-14 (DataComp1B)} & 1 & \textbf{0.2771} & 0.2554 & 0.2498 & 0.2661 & 0.2634 \\
    & 3  & 0.2769 & \textbf{0.2722} & \textbf{0.2698} & \textbf{0.2734} & \textbf{0.2729} \\
    \bottomrule
  \end{tabular}
  \caption{Comparison of single-view configuration-text scores of configurations retrieved with EVA02-E-14+ using one or three views, evaluated from different viewpoints with various VLMs.}
  \label{tab:mv_eval_full}
\end{table}



\section{Distilled Model}
\label{app:distilled}
This section discusses the architecture, training,  accuracy and smoothness of the distilled model.

\begin{figure}[h]
\begin{center}
\centerline{\includegraphics[width=.7\linewidth]{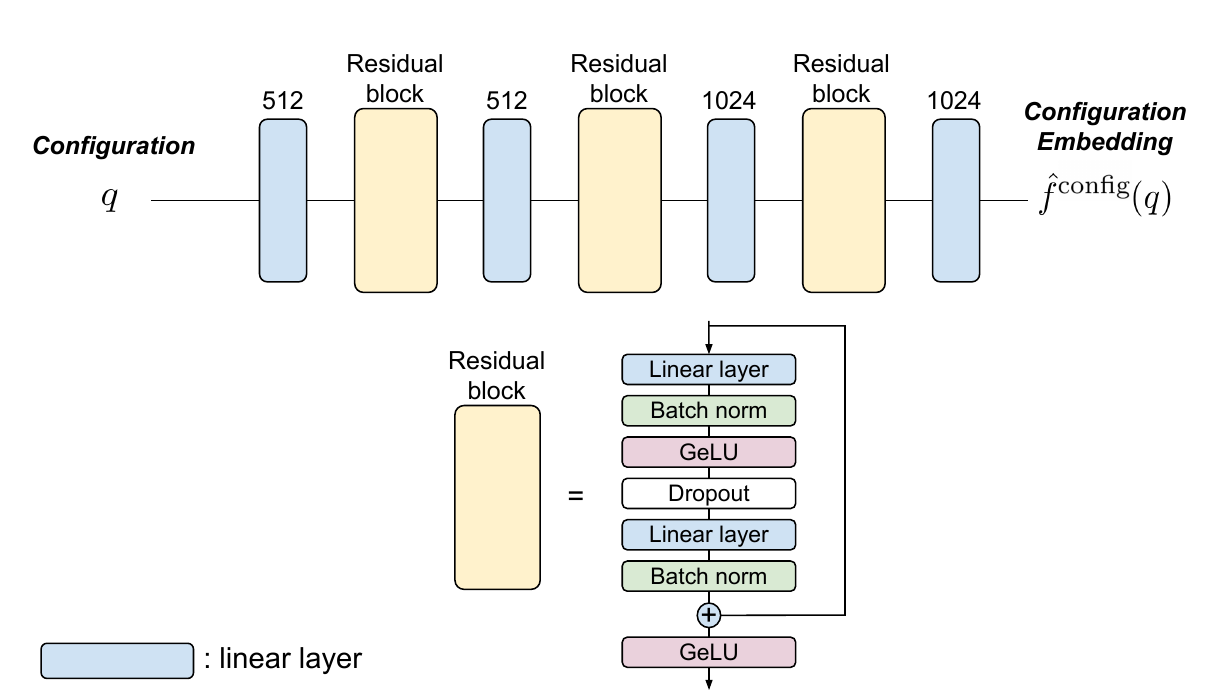}}
\caption{Architecture of the distilled model.}
\label{fig:dm_archi}
\end{center}
\vskip -0.1in
\end{figure}

\paragraph{Architecture.} Figure~\ref{fig:dm_archi} illustrates the architecture used for distilled models. This architecture is an MLP, augmented with residual connections and batch normalization inspired by the  ResNet architecture~\cite{he2016deep}. It uses GeLU~\cite{hendrycks2016gaussian} as activation function and has about $5.0 \times 10^6$ learnable parameters.

\paragraph{Training.} Distilled models are optimized with AdamW~\cite{loshchilov2017decoupled} with a weight decay of 0.01 and a learning rate of 0.001.
We use a mean-squared error (MSE) loss on predicted configuration embeddings, with random batches of size 256, and a dropout of 0.01.

\begin{figure}[h]
\vskip 0.2in
\begin{center}
\centerline{\includegraphics[width=.75\linewidth]{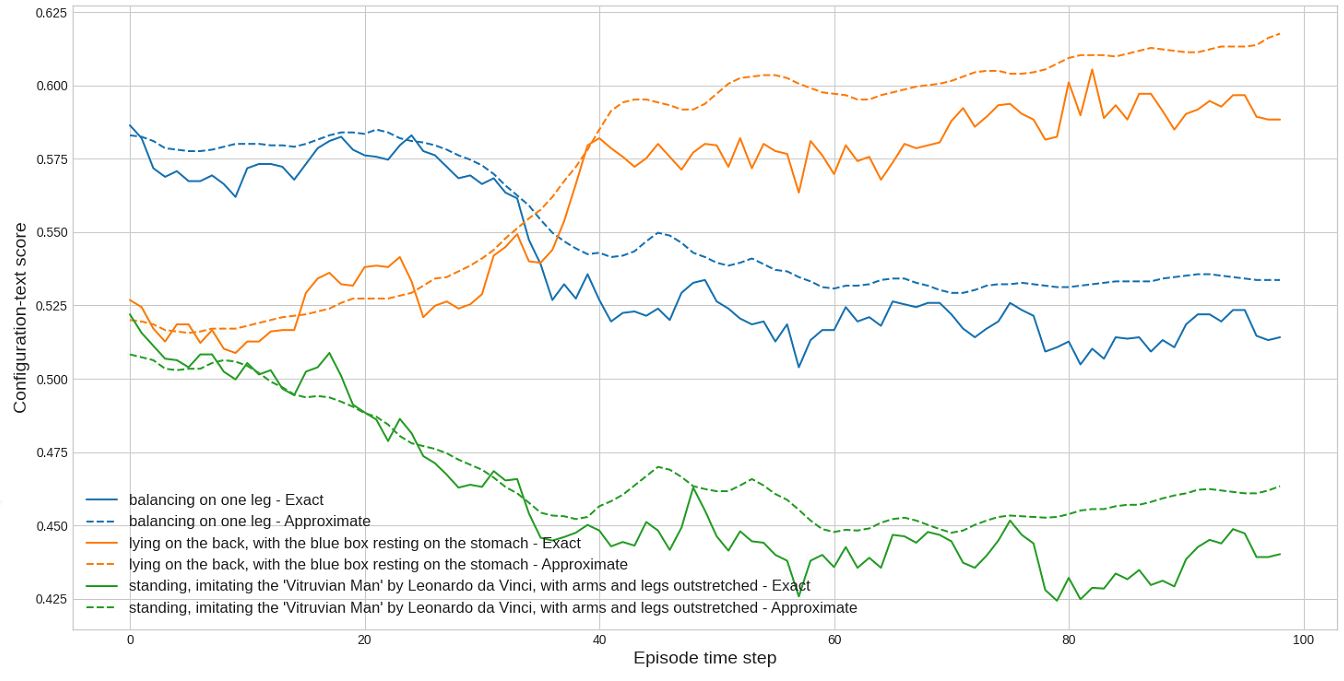}}
\caption{Comparison of exact and approximate configuration-text scores, for a trajectory resulting from the random policy, and for three representative texts.}
\label{fig:dm_smoothness}
\end{center}
\end{figure}

\paragraph{Accuracy.} 
To assess accuracy, we train a distilled model on the $2.5\times 10^6$ three-view configuration embeddings of the embedding-diversity dataset. 
Then we use it to predict the embeddings of the last $2.0\times 10^6$ configurations of the random policy dataset, noting that the datasets have their first $0.5\times 10^6$ configurations in common.
We find that the distilled model predicts such configuration embeddings with an RMSE of $4.1\times 10^{-3}$ (the normalization used in this RMSE involves division both by the number of components and the number of predictions).
The RMSE in configuration-text scores predicted by the distilled model is $9.0\times 10^{-3}$, with the mean taken over the configurations of the random policy dataset and the texts of the 256 tasks.
This RMSE is only slightly larger than the amplitude of the `noisy' oscillations of the VLM score (see Figure~\ref{fig:dm_smoothness} and Section~\ref{sec:oscillatory-nature}).\footnote{By using an ensemble of distilled models, we were able to slightly reduce these RMSEs, but the reductions were limited by this oscillatory behaviour, so we do not discuss such ensembles any further in this paper.} 

\paragraph{Smoothness.} Figure~\ref{fig:dm_smoothness} demonstrates the `smoothness' of the distilled model relative to the exact configuration-text scores. This plot was generated by rolling out a trajectory with the random policy and evaluating the exact and (distilled-model-based) approximate multiview configuration-text scores of the configurations encountered, using the text embeddings of three representative tasks.
We argue that this `smooth’ behaviour of the distilled model makes it more suitable for gradient-based finetuning of configurations than the exact VLM score.




\clearpage
\section{Configuration Finetuning}
\label{sec:appendix_finetuning}
The process used to finetune a single retrieved configuration is presented in Algorithm~\ref{alg:finetuning}. In practice, it is applied for a batch of 256 retrieved configurations in parallel, as the configuration-text score is a highly multimodal function of the configuration.
First, we precompute the text embedding (Line~2).
Then we repeatedly perform gradient ascent, with step-size $\alpha$, on the configuration-text score approximated with the distilled model (Line~4).
Each such gradient step is followed by a projection onto the set of admissible configurations (Line~5), for instance, to ensure that the objects of the environment do not interpenetrate.
We repeat this process for 80 finetuning steps. 

\begin{algorithm}[h]
   \caption{Gradient-based configuration finetuning}
   \label{alg:finetuning}
\begin{algorithmic}[1]
\STATE {\bfseries Input:} Initial configuration $q\in \configset$, text $\txt \in \textset$, VLM text encoder $\ftext$, distilled model $\fdist$ and projection $\proj$.
\STATE $z_\mathrm{t} \leftarrow \ftext(x)$ 
\WHILE{stopping criterion not met}
    \STATE $q \leftarrow  q + \alpha  \nabla_{\bq} (z_\mathrm{t} \cdot \fdist(q)) \qquad$ \COMMENT{In practice, we use Adam~\citep{kingma2014adam} for this step.}
    \STATE $q \leftarrow \proj(q)$ 
\ENDWHILE
\STATE \textbf{return} $q$
\end{algorithmic}
\label{alg:fineuning}
\end{algorithm}

\begin{figure}[h]
\begin{center}
\centerline{\includegraphics[width=.9\linewidth]{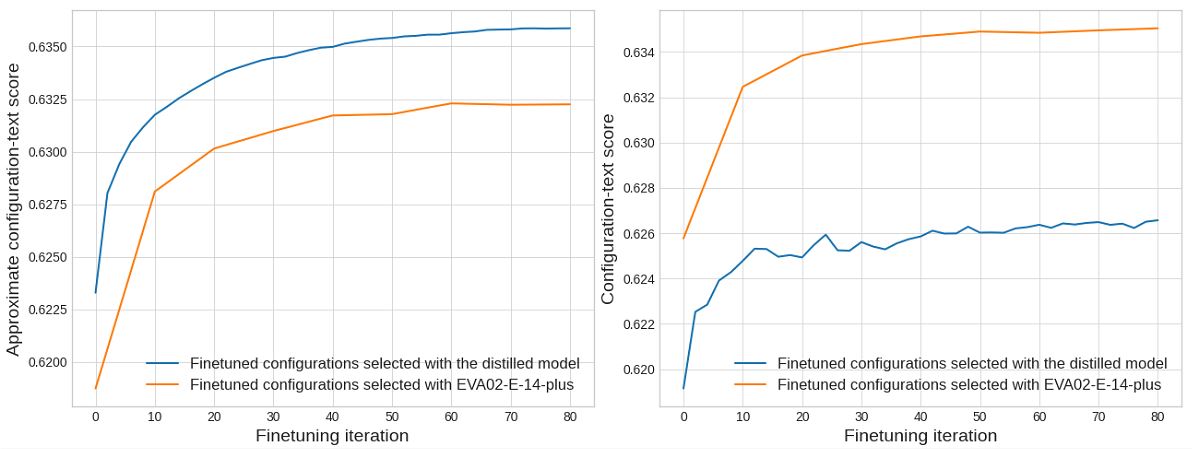}}
\caption{Evolution of the approximate configuration-text score (left) and the exact configuration-text score (right) of configurations selected by the distilled model (blue) and the configurations selected by EVA02-E-14+ (orange) during finetuning, averaged over the 256 tasks. Both scores plateau around 60 finetuning iterations. All scores shown are multiview.}
\label{fig:finetuning_it}
\end{center}
\end{figure}

This choice of the number of finetuning steps is motivated by Figure~\ref{fig:finetuning_it}, which shows that the configuration-text score for the best-in-batch configuration plateaus around 60 steps.
This figure shows that the plateauing behaviour is observed for both exact and approximate scores, when configurations are selected from the batch using both the exact and approximate scores: there is no sign of overfitting to the distilled model (that is, the exact score does not start decreasing after some number of finetuning iterations).

\clearpage

\section{LCA Architectures, Hyperparameters and Training}
\label{appendix:architectures-and-ppo}

This section discusses the architectures of the LCAs and their training, and  illustrates the performance of the MTRL baseline during training.

\subsection{LCA Architectures}
\begin{figure}[h]
\vskip -0.1in
\begin{center}
\centerline{\includegraphics[width=.65\linewidth]{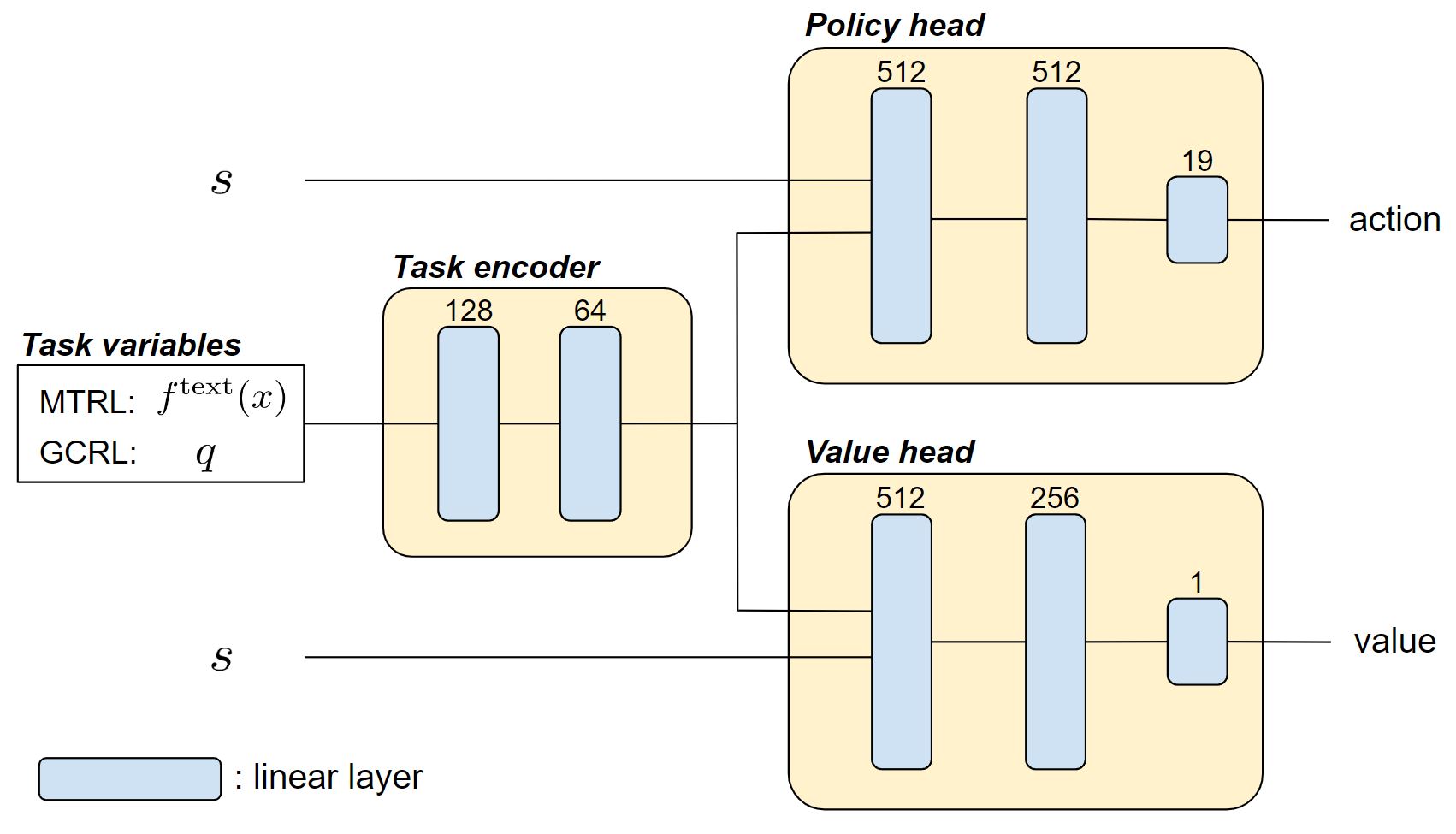}}
\caption{Architecture of MTRL and GCRL agents. Both receive state $s$ as well as task variables as input. The architecture is then composed of three MLPs: (1) the \emph{task encoder}, which embeds the task variables into a vector in $\R^{64}$; (2) the \emph{policy head}, which predicts the action from the concatenation of the state with the embedded task variables; and (3) the \emph{value head}, which predicts the value from the concatenation of the state with the embedded task variables.}
\label{fig:LCA_architecture}
\end{center}
\vskip -0.2in
\end{figure}

To ensure a fair comparison between the STRL, MTRL and GCRL agents, they share the same architectures as far as possible. 
As shown in Figure~\ref{fig:LCA_architecture}, the MTRL and GCRL agents have a task encoder shared between the policy and the value heads. The agents only differ in their inputs: the MTRL agent receives a text embedding of dimension 1024 as task variable; whereas the GCRL agent receives a target configuration of dimension 31. The STRL agent has neither task variables nor a task encoder. 
The MTRL, GCRL and STRL agents have about $1.1\times 10^6$, $0.9\times 10^6$ and $0.6\times 10^6$ learnable parameters respectively.

\subsection{LCA Training}
\label{appendix:lca_training}

We train the STRL, MTRL and GCRL agents to maximize the finite-horizon discounted sums of their respective rewards~(\ref{eq:mtrl_reward}) and~(\ref{eq:gcrl_reward}), using the proximal policy optimization (PPO) algorithm~\citep{schulman2017proximal}, with the hyperparameters specified in Table~\ref{tab:hp_ppo_shared}.

\begin{table}
\centering
\caption{Hyperparameters used when training the STRL, MTRL and GCRL agents with PPO.}
\label{tab:hp_ppo_shared}
\vskip 0.15in
\begin{tabular}{ l | c }
\hline
\textbf{Hyperparameter}           & \textbf{Value} \\
\hline
 Clipping             & 0.2 \\
 Discount factor, $\gamma$       & 0.999 \\
 GAE parameter, $\lambda$      & 0.95 \\ 
 Update time-step        &  $204\,800$ (MTRL and GCRL), $25\,600$ (STRL) \\
 Batch size         &  $102\,400$ (MTRL and GCRL), $12\,800$ (STRL) \\
 Epochs              &  10 \\
 Learning rate        & 5e-4 \\
 Learning-rate schedule & linear annealing \\
 Gradient norm clipping & 0.5 \\
 Value clipping & no \\
 Entropy coefficient & 2.5e-2 \\
 Value coefficient & 0.5 \\
 Activation function & $\text{GeLU}$ \cite{hendrycks2016gaussian} \\
 Optimizer & AdamW \cite{loshchilov2017decoupled} \\
 Weight decay & 0.01 \\
\hline
\end{tabular}
\\
\end{table}

The training times of our methods are as follows, using an NVIDIA RTX A6000 GPU and a 40-core Intel Xeon w7-2475X:
we train GCRL for $2 \times 10^9$ samples, which takes 149 hours;
we train MTRL for $4\times 10^8$ samples, which takes 30 hours;
and for each task, we train STRL for $5\times 10^7$ samples, which takes 35 minutes. 
More training samples are required for the training performance of GCRL to plateau than are required for MTRL. 
We think this is because the number of goal configurations that GCRL is trained to reach is far greater than the number of tasks that MTRL is trained to perform ($2.5\times 10^6$ versus $128$), and because the goal configurations of GCRL are diverse.

MTRL and STRL trainings are feasible because we leverage the distilled model: if we had trained the MTRL agent on the exact VLM reward derived from EVA02-E-14+ for the same number of samples, then the reward calculation alone would have taken about 3000 hours (125 days) for the single-view version and about 9000 hours (475 days) for the multiview version. The distilled model reduce these durations by a factor of $40\,000 \times$, allowing to spend most training time on data collection and PPO updates.


\begin{figure}
\vskip -0.1in
\begin{center}
\centerline{\includegraphics[width=.9\linewidth]{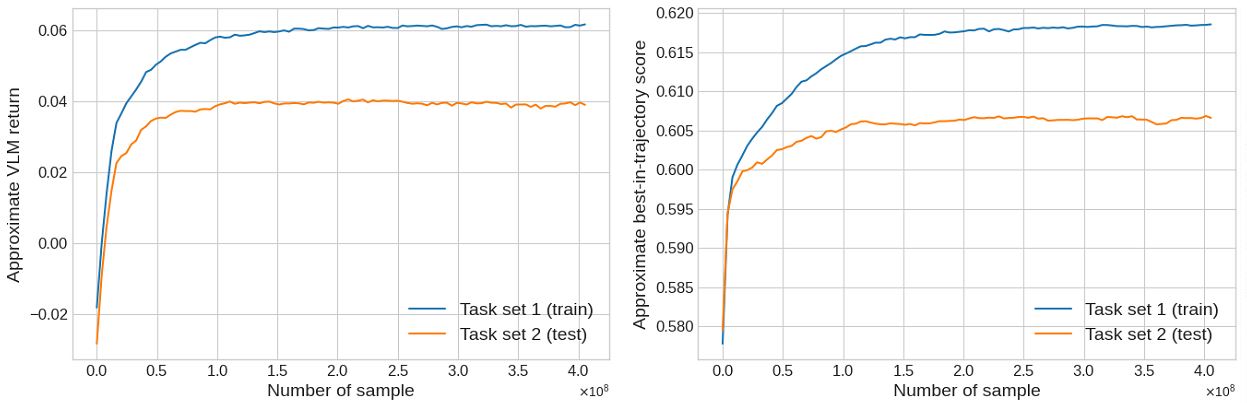}}
\caption{Approximate (multiview) VLM return and approximate best-in-trajectory configuration-text score of the MTRL baseline during training, averaged over the 128 training tasks (blue) and the 128 test tasks (orange), and averaged over 10 episodes per task.}
\label{fig:mtrl_training}
\end{center}
\vskip -0.2in
\end{figure}

Figure~\ref{fig:mtrl_training} shows the evolution of the performance of the MTRL agent on both training and test tasks during training. The performance has plateaued, and we do not observe any overfitting (that is, degradation of test-task performance), even after extended training.

\clearpage

\section{Evaluation of LCAs}
\label{sec:appendix_lca_eval}
This section provides additional evaluations of the proposed LCAs. We compare STRL, MTRL and GCRL variants using exact VLM scores, rather than relying on the distilled model (Section~\ref{app:LCA-exact}); and in terms of 
the number of tasks where each GCRL variant outperforms the STRL and MTRL baselines (Section~\ref{app:win-rates}).
Finally, we present task-by-task performance metrics as polar plots (Section~\ref{app:polar}).

\subsection{Evaluating LCAs with Exact VLM Scores}
\label{app:LCA-exact}
\paragraph{Definition.} 
Given trajectory  $(s_0, a_0), \dots, (s_{T-1}, a_{T-1})$, for an episode of length $T$, the \emph{VLM return} is the discounted sum
$\sum_{t=0}^{T-1} \gamma^t R_{\txt}(s_t, a_t)$, and the \emph{best-in-trajectory score} is $\max_{t=0, \dots, T-1} \qtextsim(\natproj(s_t), \txt),$
where the VLM reward $R_{\txt}(s,a)$ for state $s$, action $a$ and text $\txt$ is as defined in equation~(\ref{eq:mtrl_reward_exact}).

\begin{figure}[h]
\begin{center}
\centerline{\includegraphics[width=.92\linewidth]{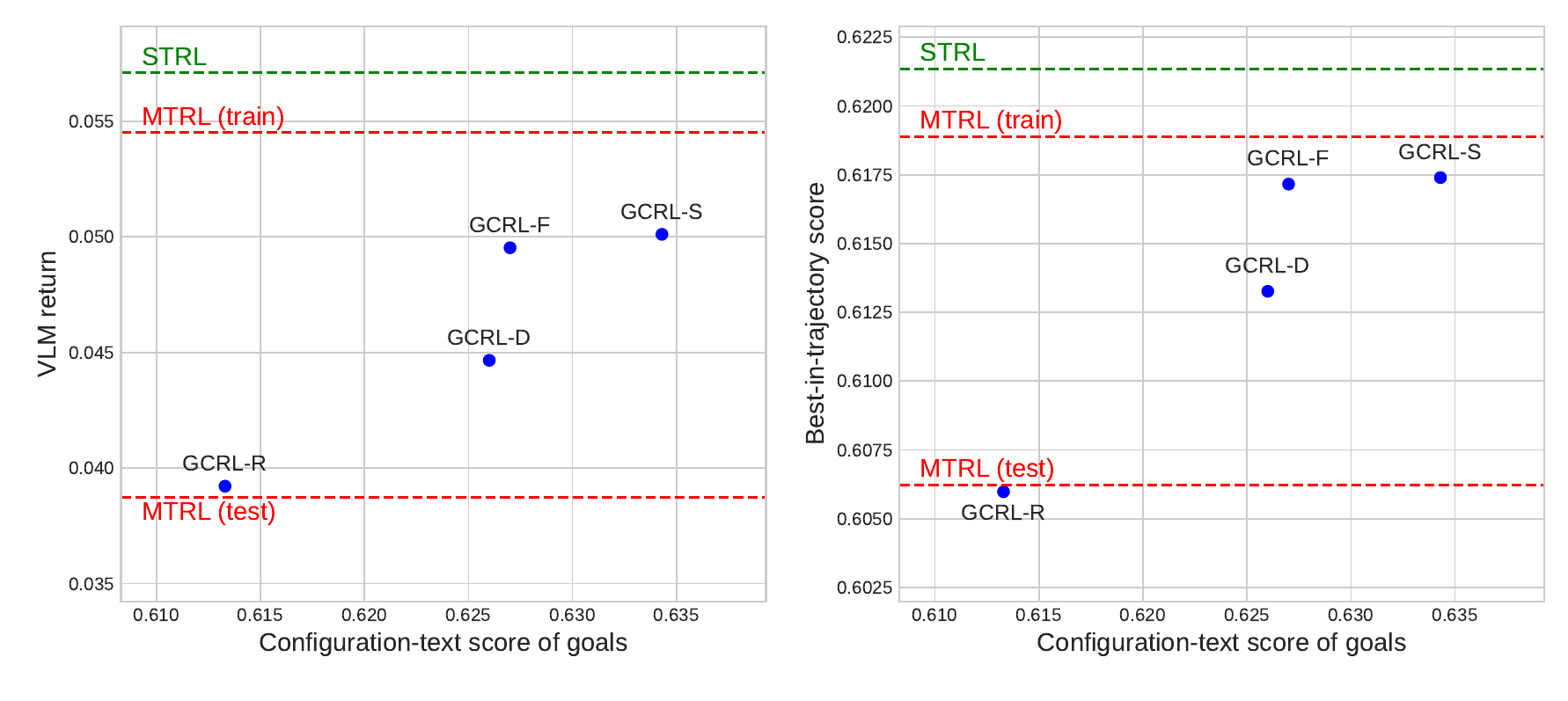}}
\caption{Returns and best-in-trajectory configuration-text scores evaluated by the VLM, for GCRL agents, and for STRL and MTRL baselines. The results are averaged over the 256 tasks.}
\label{fig:LCA_eval}
\end{center}
\end{figure}

\paragraph{Results.} Figure~\ref{fig:LCA_eval} compares the GCRL, STRL and MTRL agents in terms of their VLM return and best-in-trajectory scores.
This figure is thus the analogue of Figure~\ref{fig:LCA_eval_approx}, which was based on approximate scores.
We make the following observations:
\begin{enumerate}[topsep=0pt,itemsep=1pt,parsep=0pt]
 \item GCRL-D, GCRL-F and GCRL-S outperform MTRL (test): thus, decoupling goal-generation and goal-reaching can result in better generalization. 
 \item Providing goal configurations with higher configuration-text scores to the GCRL agent consistently improves its VLM return and best-in-trajectory score. Indeed, GCRL-F and GCRL-S, which use the highest-quality goals, are competitive with MTRL (train) in terms of best-in-trajectory score. 
\item GCRL-S outperforms GCRL-F, in contrast with Figure~\ref{fig:LCA_eval_approx}. 
This is to be expected as GCRL-S uses configurations selected by the VLM, and Figure~\ref{fig:LCA_eval} shows performance evaluated with the VLM, whereas GCRL-F and Figure~\ref{fig:LCA_eval_approx} are both based on approximate scores.
\item STRL outperforms MTRL (train), but only by a small margin relative to the gap between MTRL (test) and MTRL (train).
\end{enumerate}

\subsection{LCA Performance by Task Category}
\label{app:task_breakdown}

We now compare the performance of the LCAs on tasks involving Humanoid-cube interactions with those that do not. To do so, we split the set of 256 tasks (listed in Appendix~\ref{appendix:tasks}) into three groups:
\begin{itemize}
    \item \textbf{Tasks involving cube interaction.} This group consists of tasks that explicitly require Humanoid-cube interactions, such as \textit{``doing a push-up with both feet on the box''}. There are 36 such tasks, with the following IDs: 27, 35, 50, 51, 75, 76, 77, 94, 101, 102, 103, 104, 129, 132, 138, 139, 140, 141, 142, 143, 153, 156, 158, 162, 163, 165, 170, 173, 192, 215, 216, 225, 236, 240, 243 and 244.
    \item \textbf{Tasks potentially involving cube interaction.} This group consists of tasks that could potentially benefit from Humanoid-cube interaction, although they do not explicitly mention such interaction. For instance, when performing the task  \textit{``standing on tiptoes, arms reaching up as if trying to touch the ceiling''}, there is some possibility that standing on the cube might help. There are 32 such tasks, with the following IDs: 44, 61, 66, 91, 99, 106, 111, 113, 143, 115, 119, 120, 121, 122, 124, 135, 157, 166, 167, 168, 169, 171, 172, 180, 181, 182, 183, 184, 205, 218, 231 and 233.
    \item \textbf{Tasks involving no cube interaction.} These tasks do not require any form of Humanoid-cube interaction. This last group contains the remaining 188 tasks.
\end{itemize}

\begin{figure}[h]
\begin{center}
\centerline{\includegraphics[width=.92\linewidth]{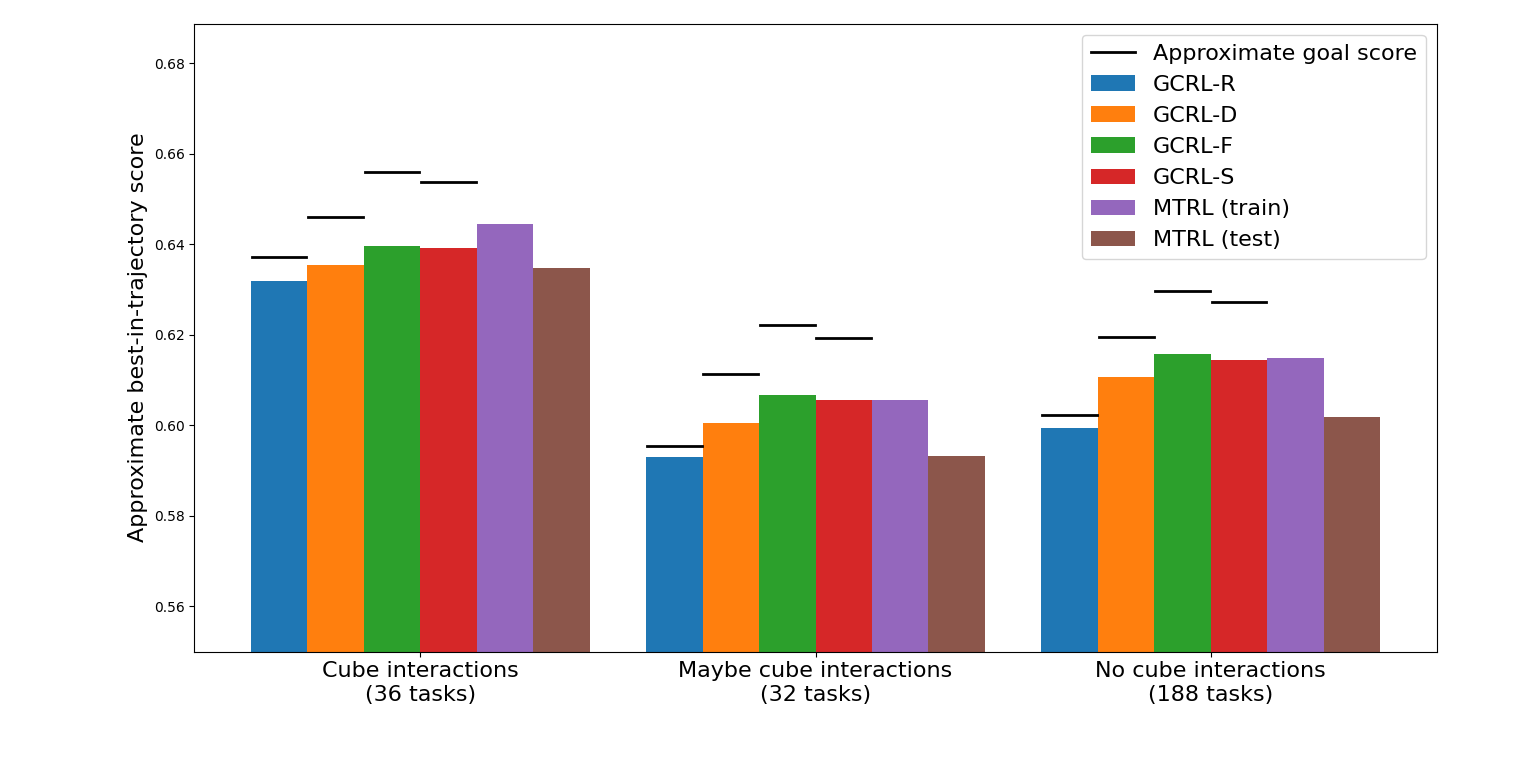}}
\caption{Approximate best-in-trajectory configuration-text scores for GCRL agents and for MTRL baselines averaged over tasks involving Humanoid-cube interactions (left), tasks potentially involving Humanoid-cube interactions (middle) and tasks involving no  Humanoid-cube interactions (right). For those LCAs using a goal, the average of the approximate score for the goal is shown as a black line segment.}
\label{fig:LCA_eval_task_breakdown}
\end{center}
\end{figure}

Figure~\ref{fig:LCA_eval_task_breakdown} shows the approximate best-in-trajectory score of each LCA, averaged by task category. For each category, the main result holds: GCRL-D, GCRL-F and GCRL-S outperform MTRL (test) in zero-shot generalization. 

Interestingly, MTRL (train) is only better than GCRL-F on tasks requiring Humanoid-cube interactions, attaining a higher approximate best-in-trajectory score on 64\% ($\approx 23/36$) of such tasks; and MTRL (train) only beats GCRL-F on  36\% ($\approx 80/220$) of the remaining tasks. 
Upon further investigation, we find that on many tasks requiring Humanoid-cube interaction, neither GCRL-F nor MTRL (train) executes the task in a way that would be satisfactory to a human observer. Part of the problem is that such tasks involve relations between the humanoid and the cube, which are often difficult for VLMs to assess, as discussed in Appendix~\ref{app:failures}. 
In other cases, both MTRL and GCRL appear limited by their exploration: for instance, the VLM can assess if a box has been lifted up, but neither the MTRL nor the GCRL policy can reach that state; whereas STRL, which performs more exploration per task, can successfully lift the cube.

\subsection{Comparing Win Rates of LCAs}
\label{app:win-rates}

\begin{table}
\centering
\begin{tabular}{|l|c|c|c|c|}
\hline
 & \textbf{GCRL-R}     & \textbf{GCRL-D}     & \textbf{GCRL-F}     & \textbf{GCRL-S}     \\ \hline
 \textbf{vs. MTRL (test) on exact best-in-trajectory score}       & 120/256 & 178/256 & 210/256 & 210/256 \\ \hline

 \textbf{vs. MTRL (test) on exact VLM return}                    & 120/256 & 163/256 & 198/256 & 205/256 \\ \hline
\textbf{vs. MTRL (test) on approximate best-in-trajectory score}       & 99/256 & 185/256 & 214/256 & 207/256 \\ \hline
\textbf{vs. MTRL (test) on approximate VLM return}                    & 109/256 & 168/256 & 203/256 & 202/256 \\ \hline
\multicolumn{5}{c}{} \\[-1ex] 
\hline 
\textbf{vs. MTRL (train) on exact best-in-trajectory score}      & 4/256 & 83/256 & 124/256 & 128/256 \\ \hline
\textbf{vs. MTRL (train) on exact VLM return}                   & 13/256 & 64/256 & 91/256 & 94/256 \\ \hline
\textbf{vs. MTRL (train) on approximate best-in-trajectory score}      & 6/256 & 89/256 & 153/256 & 130/256 \\ \hline
\textbf{vs. MTRL (train) on approximate VLM return}                   & 5/256 & 59/256 & 100/256 & 91/256 \\ \hline
\multicolumn{5}{c}{} \\[-1ex] 
\hline 
\textbf{vs. STRL on exact best-in-trajectory score}      & 1/256 & 60/256 & 91/256 & 102/256 \\ \hline
\textbf{vs. STRL on exact VLM return}                   & 5/256 & 54/256 & 70/256 & 83/256 \\ \hline
\textbf{vs. STRL on approximate best-in-trajectory score}      & 0/256 & 65/256 & 118/256 & 99/256 \\ \hline
\textbf{vs. STRL on approximate VLM return}                   & 0/256 & 37/256 & 82/256 & 72/256 \\ \hline

\end{tabular}
\caption{Win rates of GCRL variants against the MTRL baselines on the MTRL test tasks (top 4 rows) and training tasks (middle 4 rows), and against the STRL baselines (bottom 4 rows). The win rates are given for the best-in-trajectory score and the VLM return, based on both the exact (using the VLM) and approximate (using the distilled model) multiview scores.}
\label{tab:winrates}
\end{table}

Table~\ref{tab:winrates} presents the number of tasks where GCRL variants outperform the STRL and MTRL baselines according to a variety of criteria. It shows that providing higher quality goal configurations to the GCRL agent not only increases its average VLM return (Figures~\ref{fig:LCA_eval_approx} and~\ref{fig:LCA_eval}) but also increases the number of tasks in which it outperforms the baselines. 
Not only do GCRL-D, -F and -S dominate MTRL (test) for all criteria shown, 
but also, the best-in-trajectory scores of GCRL-F and -S exceed those of MTRL (train) on at least 48\% ($\approx 124/256$) of tasks.
Moreover, although STRL was trained to optimize the approximate VLM return, STRL attains a lower (exact and approximate) VLM return than GCRL-F and -S on at least 27\% ($\approx 70/256$) of tasks. 

The win rates of GCRL-D, -F and -S for best-in-trajectory score, are consistently higher than the corresponding win rates for VLM return.  
We interpret this phenomenon as follows.
Let $(\sco{t})_{t=0}^{T-1}$ be the sequence of (approximate multiview) VLM scores encountered in an episode by some policy. Let score $\sco{T}$ be the \emph{postultimate} score defined as the score for the state predicted to follow the last state of the trajectory, assuming deterministic transitions for simplicity. 
Then the VLM return for that episode is
\begin{align*}
\sum_{t=0}^{T-1} \gamma^t (\sco{t+1}-\sco{t}) 
= \gamma^{T-1} \sco{T} - \sco{0} + (1-\gamma) \sum_{t=1}^{T-1} \sco{t}  + O\left( (1-\gamma)^2 \right) \qquad \text{as $\gamma\to 1$},
\end{align*}
where the equality follows from the binomial expansion of $(1-(1-\gamma))^t$.
As the MTRL agent attempts to maximize this return, it can be seen as maximizing a weighted sum of the postultimate score $\sco{T}$ and the average score $\sum_{t=0}^{T-1}\sco{t}/T$, noting that the agent has no control over $\sco{0}$. 
The MTRL policy is free to choose its final configuration as a function of its initial state, and in such a way that its trajectory maximizes this weighted sum. 
This choice need not result in near-optimal best-in-trajectory scores: maximizing a sum is not generally the same as maximizing the largest term of that sum.

In contrast, the GCRL agent has a fixed goal, and by the binomial expansion argument given above, it can be seen as minimizing a weighted sum of the distance to that goal in its postultimate configuration and the average distance to that goal. 
As the text-to-goal methods are designed to choose goals that score as highly as possible, if configurations near the goal are reachable and the score is not too rapidly varying near the goal, then by minimizing this weighted sum, the GCRL agent will necessarily attain near-optimal best-in-trajectory scores.

\begin{figure}
\begin{center}
\centerline{\includegraphics[width=.92\linewidth]{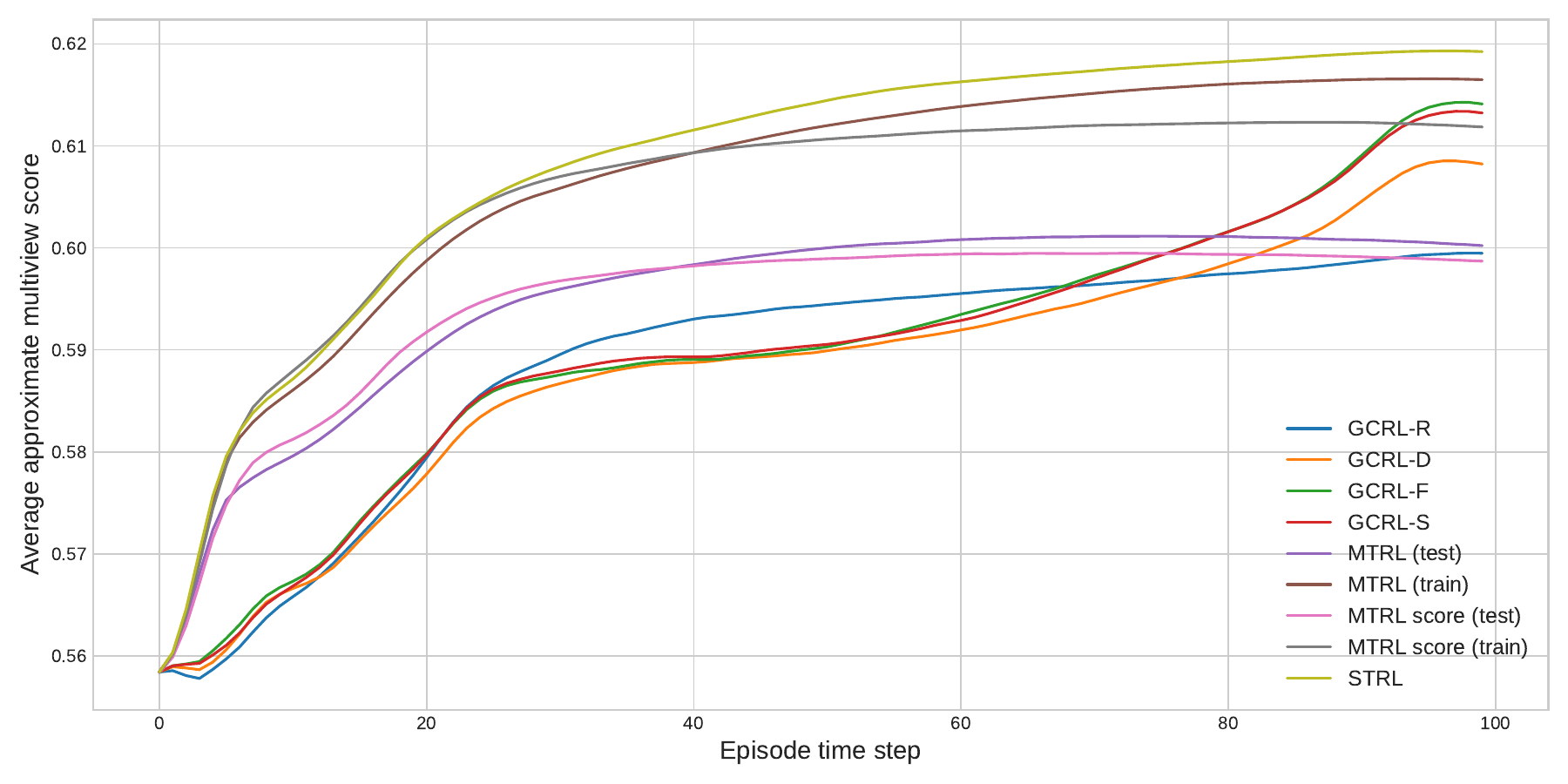}}
\caption{Evolution of the approximate multiview configuration-text score score of GCRL variants and MTRL baselines, averaged over the 256 tasks and 10 episodes per task, with respect to the episode time step.}
\label{fig:LCA_timestep}
\end{center}
\end{figure}

To provide further insight, Figure~\ref{fig:LCA_timestep} shows the approximate multiview configuration-text scores of different LCAs as a function of the time step of an episode. The scores of the MTRL agents increase more rapidly than those of the GCRL agents in the first 8 time steps and remain higher until around time step 80. This ensures higher average scores for the MTRL agents on that interval.  However, the average scores of GCRL-D, -F and -S increase rapidly in time steps 85--95. One hypothesis for this final burst is that some of the goals are configurations around which it is hard to stabilize. For such goals, the GCRL agent maximizes its trade-off between the distance-to-goal in the postultimate configuration and the average distance-to-goal by waiting in a more-stable-but-not-too-distant configuration until the last few time steps.  


\paragraph{Time-differenced versus raw VLM rewards.}
Figure~\ref{fig:LCA_timestep} also compares MTRL (train) in brown with \emph{MTRL score (train)} in grey, and MTRL (test) in purple with \emph{MTRL score (test)} in pink. 
The \emph{MTRL score} variants are trained with the reward
\begin{equation}
\hat{R}^{\text{past work}}_{\txt}(s, a) = \qtextsurr(\natproj(s), \txt).
\label{eq:reward_ablated}
\end{equation}
This is the reward used by previous works~\cite{mahmoudieh2022zero, fan2022minedojo, rocamonde2023vision, baumli2023vision}, except that it is based on multiview configuration-text scores, approximated with our distilled model. 
It can be seen as an ablated version of our reward function~(\ref{eq:mtrl_reward}), using the score instead of the time difference of scores.
We observe that MTRL score (train and test) attain higher scores than MTRL (train and test) respectively from time step 8 to 38. However, MTRL (train and test) dominate after time step 40, and achieve higher peak and final scores.
Other authors have used reward~(\ref{eq:reward_ablated}) to perform tasks defined in terms of the final configuration of the environment, such as the task \textit{`stacking a yellow block on top of a red block'} considered by~\citet{mahmoudieh2022zero}. This result shows that applying our time-differenced reward may be more suitable for such tasks.

\subsection{Task-by-Task Evaluation of LCAs}
\label{app:polar}

\begin{figure}
\vskip -0.1in
\begin{center}
\centerline{\hspace{26mm}\includegraphics[width=.61\linewidth]{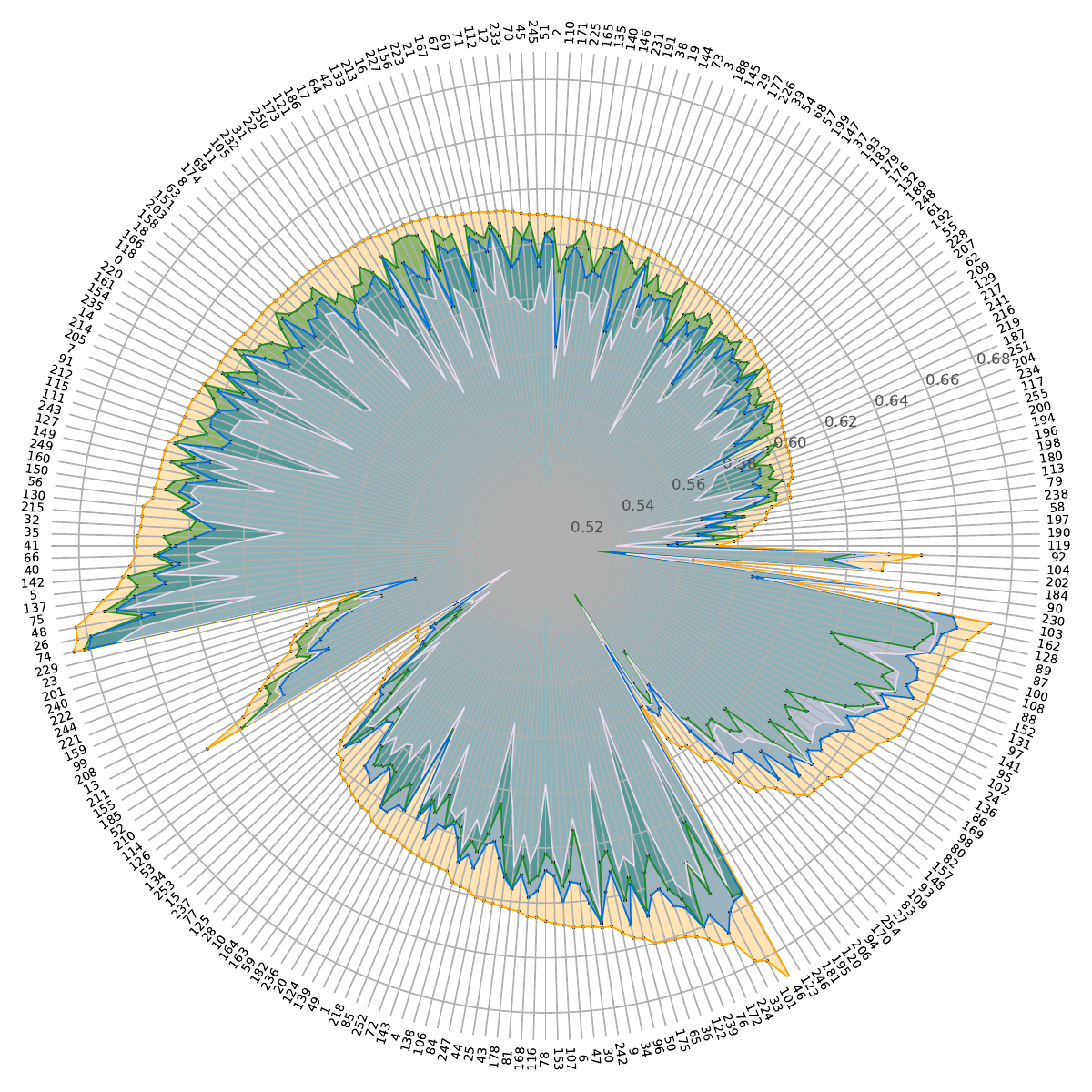}\includegraphics[width=.15\linewidth]{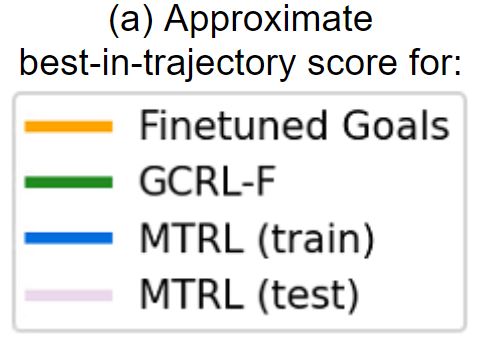}}
\vskip -0.1in
\centerline{\hspace{26mm}\includegraphics[width=.61\linewidth]{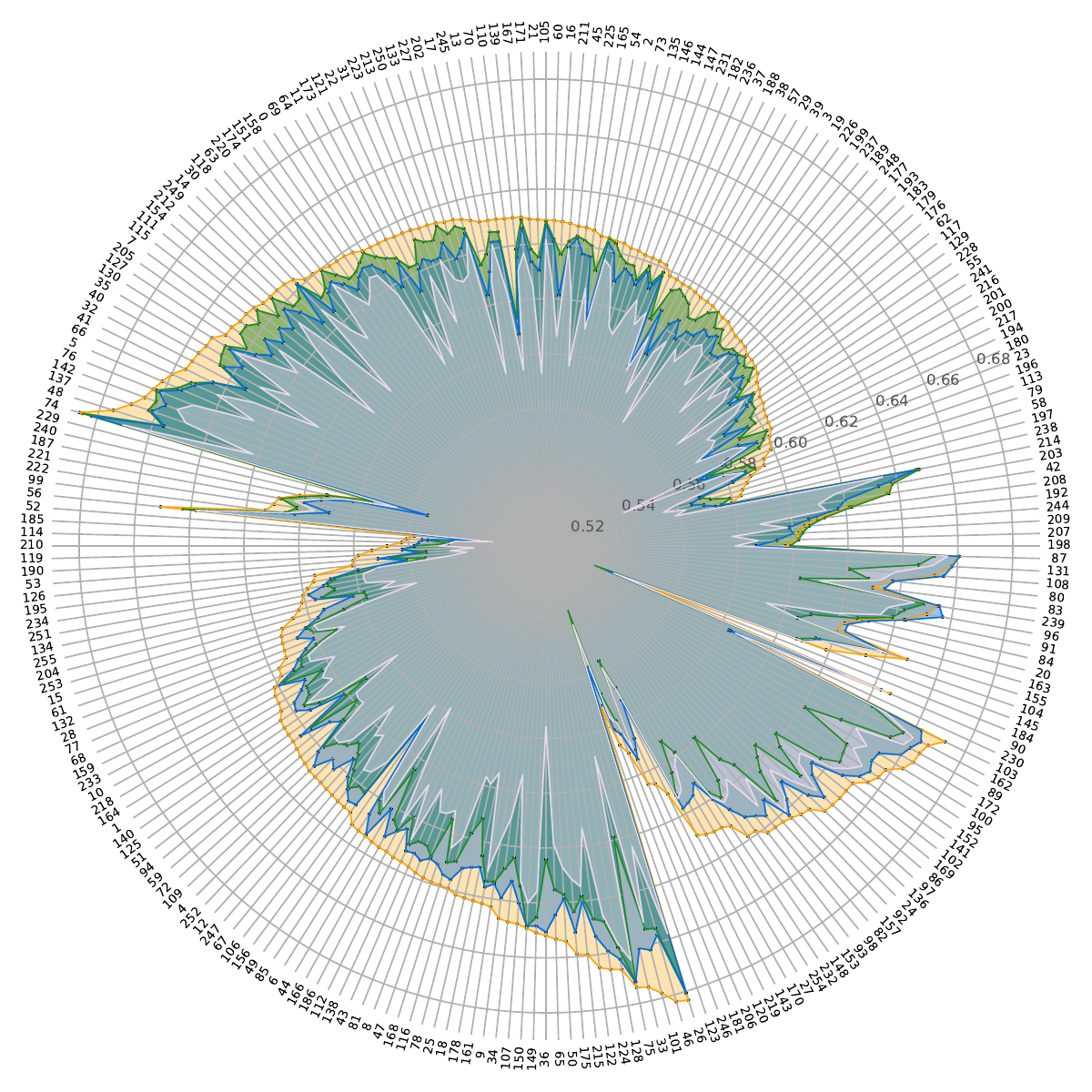}\includegraphics[width=.15\linewidth]{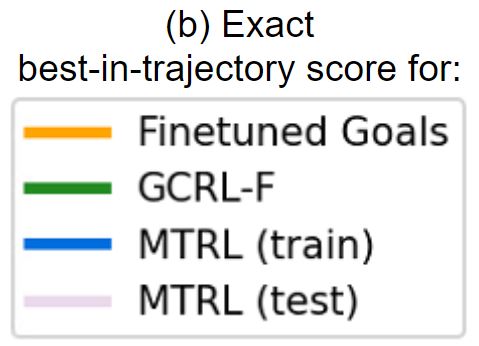}}
\caption{
Polar plots of approximate (top) and exact (bottom) multiview configuration-text scores for each task. Task numbers are shown in the outer circle. Scores are shown for: best-in-trajectory configurations of GCRL-F (green), MTRL (train) (blue) and MTRL (test) (pale grey magenta), averaged over 10 episodes per task; and for the finetuned configurations used as goals by GCRL-F (orange). The tasks are grouped by ranking of these scores. For instance, the top region of both plots has orange containing green containing blue containing pale grey magenta,  corresponding to the ranking $\text{Finetuned Goals} > \text{GCRL-F} > \text{MTRL (train)} > \text{MTRL (test)}$. Within each group, they are sorted by the finetuned goal score. 
}
\label{fig:LCA_taskbytask}
\end{center}
\vskip -0.2in
\end{figure}

Figure~\ref{fig:LCA_taskbytask} compares the approximate (top) and exact (bottom) task-by-task best-in-trajectory scores of the GCRL-F agent with MTRL (test) and MTRL (train), and with the scores of the finetuned configurations used as goals by GCRL-F. We make the following observations:

\begin{itemize}[leftmargin=5mm, itemsep=0.8mm, parsep=0mm, topsep=1mm, partopsep=0mm]
    \item 
    The finetuned goals have the highest approximate configuration-text scores for every task (in the top plot, the orange areas always contain the green, blue and grey magenta areas). 
    The fact that the GCRL agent sometimes fails to reach its goal may be because the goal is intrinsically hard to reach in the time limit of 100 steps, or due to a lack of training. 
    \item While the (exact) best-in-trajectory scores sometimes surpass the goal scores (bottom plot), this happens for $\le 10\%$ of tasks, and the improvement is always small. 
    \item 
    GCRL-F dominates MTRL (test) for both exact and approximate scores. Examining tasks where MTRL (test) outperforms GCRL-F, we observe a high resemblance between those tasks and the MTRL agent's training tasks.
    For instance, when the MTRL agent is tested on task 152: \emph{``sitting on the floor with straight legs and touching its head''}, it has been trained on task 151: \emph{``sitting on the floor with straight legs and one arm in the air''}.
\end{itemize}

\begin{figure}
\begin{center}
\centerline{\includegraphics[width=.9999\linewidth]{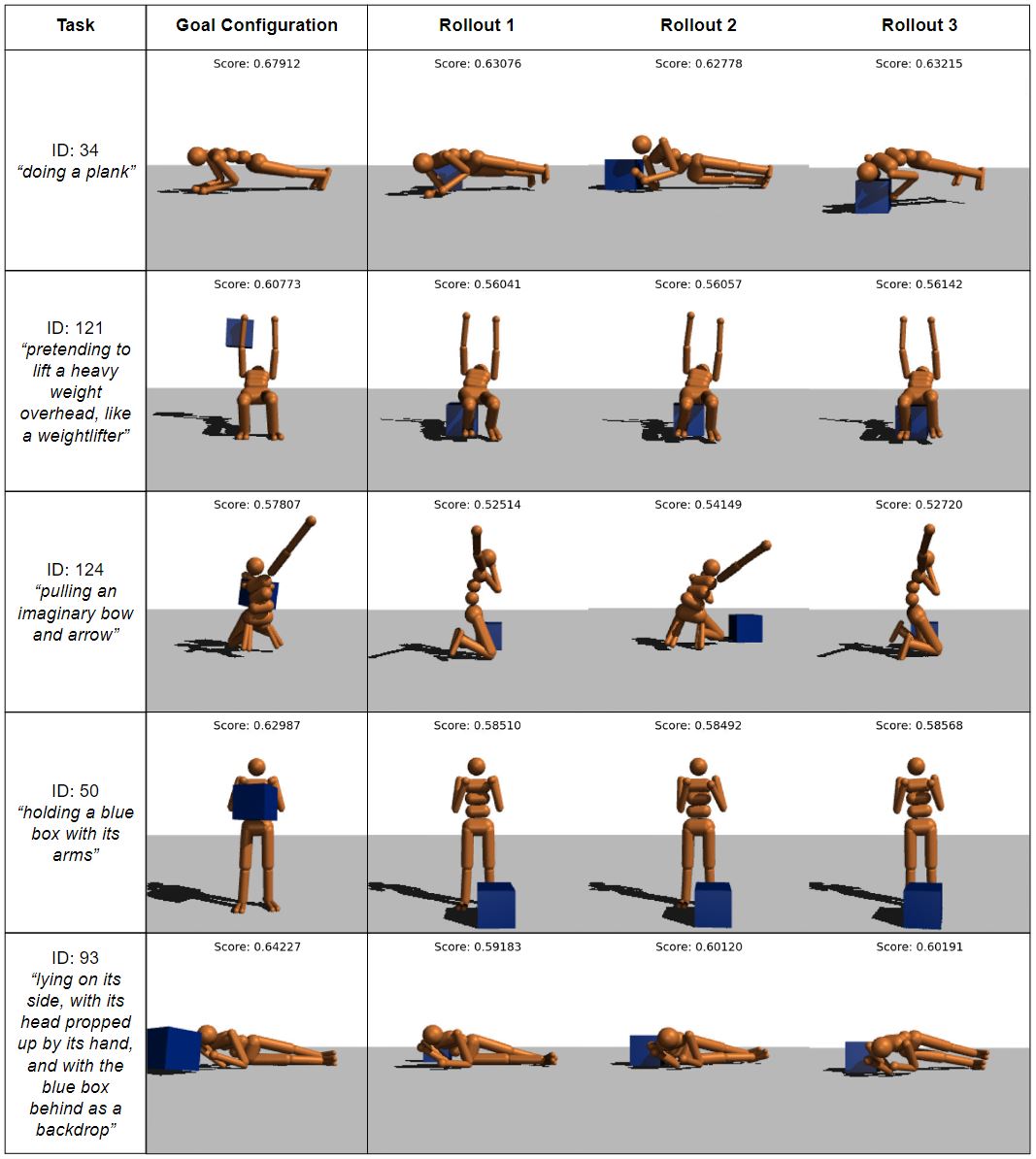}}
\caption{Front views of the (finetuned) goal configurations and the best-in-trajectory configurations reached by GCRL on its first three rollouts. The tasks shown are the `worst five failures' of GCRL, in the sense that the difference between the goal score and the best-in-trajectory score (averaged over episodes for that task) is largest.}
\label{fig:failure_1}
\end{center}
\end{figure}

\begin{figure}
\begin{center}
\centerline{\includegraphics[width=.999\linewidth]{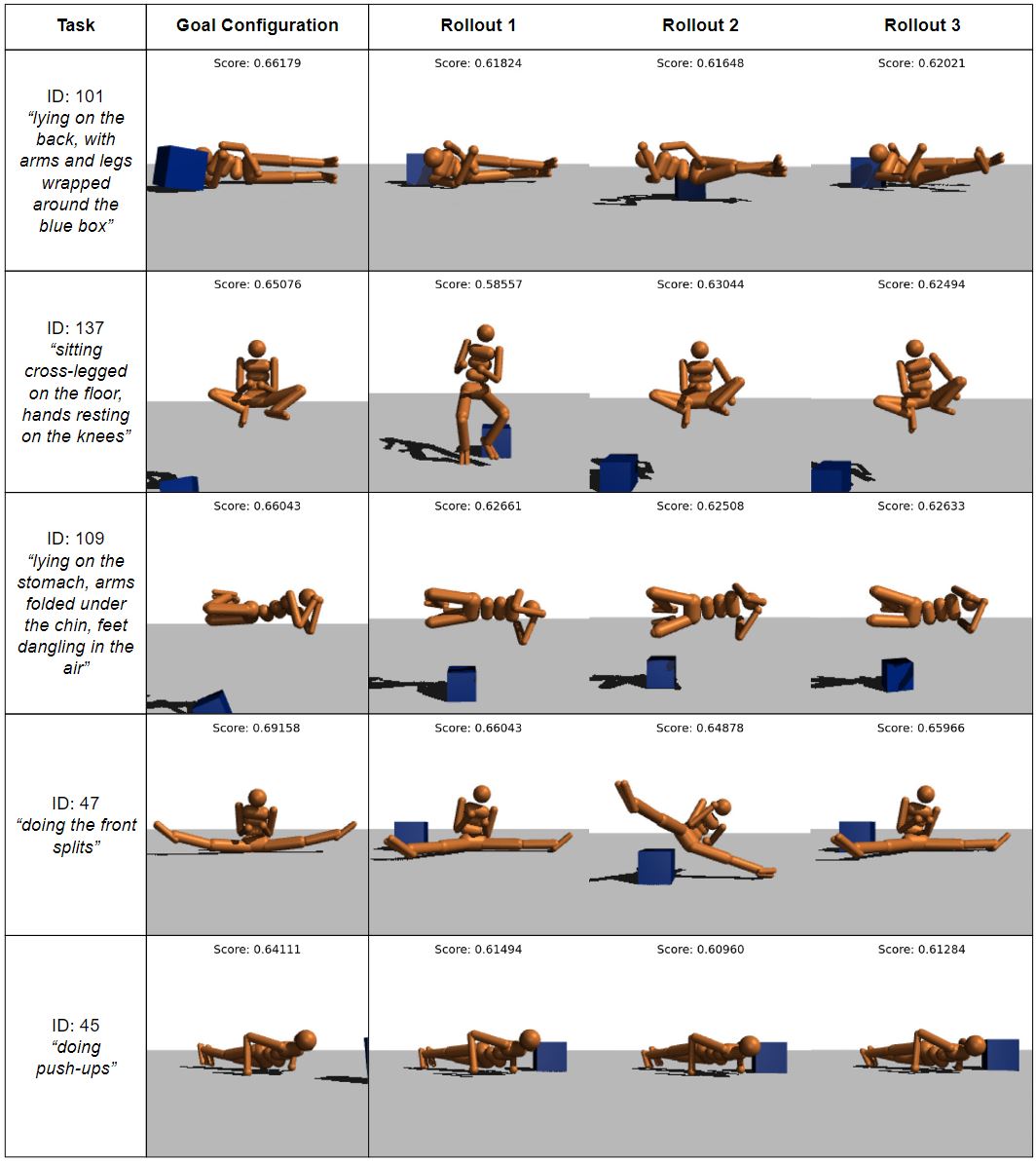}}
\caption{Front views of the (finetuned) goal configurations and the best-in-trajectory configurations reached by GCRL on its first three rollouts. The tasks shown are the `next worst five failures' of GCRL after those shown in Figure~\ref{fig:failure_1}. (If the tasks were ranked in order of decreasing difference between the goal score and the average best-in-trajectory score, then these tasks would be placed $6^\text{th}$ to $10^\text{th}$).}
\label{fig:failure_2}
\end{center}
\end{figure}

\clearpage
\subsection{Failure Study}
\label{app:failures}

We pursue our task-by-task evaluation by analysing failure modes of the GCRL agent.
To do so, we investigate the 10 tasks for which the difference between the score of the goal and the average best-in-trajectory score is largest.
Figures~\ref{fig:failure_1} and~\ref{fig:failure_2} illustrate these 10 tasks,
showing the goal and the best-in-trajectory configurations reached by GCRL in its first three rollouts. We interpret the failures as follows:
\begin{itemize}
    %
    %
    \item \textbf{Cube still visible (tasks 34, 47 and 45).} The GCRL agent approximately reaches the desired configuration, except that it does not move far enough relative to the cube that the cube goes out of view. 
    Although the cube is irrelevant for these tasks, its presence in the scene results in substantially lower scores.
    We observe from videos that in general the agent tends to move into a pose and then reposition itself (locomote) relative to the cube. When the agent is upright, it is able to locomote fast. However, in tasks 34, 47 and 45, the agent is lying down or seated and it locomotes by jittering. This mode of locomotion is too slow to travel far enough that the cube goes out of view, in the limited time span of one episode. 
    Perhaps this is a local optimum in policy space, and given more training the agent might learn to locomote then pose.
    \item \textbf{Cube in the way (tasks 93 and 101).} As just remarked, the agent tends to pose then locomote. In these tasks, the consequence is that the Humanoid lies down in front of the cube. From this position it is not able to jitter behind the cube in the limited time. Moreover, the visual difference between being in front of and behind the cube is large, hence the best-in-trajectory score is substantially worse than that of the goal.  
    \item \textbf{Horizon problem (tasks 137 and 109).} In the goal configurations of these tasks, the VLM misinterprets the horizon as a structure that the Humanoid can sit or lie on. This leads to  unstable goal configurations with the Humanoid floating in the air. Even though the GCRL agent does an impressive job of reaching nearby configurations by jumping, it cannot make such jumps accurately. In particular, attaining a precise apparent alignment between the legs and the horizon seems important to ensure the high score of the goal state. These failures suggest that the proposed text-to-goal method could be further improved by including views from different heights in the multiview score. 
    \item \textbf{Cube manipulation (tasks 121, 124 and 50).} In these tasks, the goal configurations require the GCRL agent to manipulate the cube. 
    In the goal configurations of tasks 121 and 124, the VLM seems to interpret the cube as a kettlebell and as a quiver respectively. These goals would require tremendous skill to be reached: in  further work, it would be interesting to select goals based on measures of ease of reachability. 
    However, task 50 exposes the fact that the GCRL agent has not yet learned the simple skill of holding a cube with its hands. We suspect this is because few training episodes encompass an increase in the cube's height, so the agent tends to ignore the cube's height. More training and additional goal-reaching training methods such as HER and ASP~\cite{andrychowicz2017hindsight,openai2021asymmetric} may help. 
    We also plan to investigate longer episodes.

\end{itemize}

\clearpage

\section{Study of VLM Scores}
\label{sec:study-scores}
This section studies some properties of VLM scores, and discusses how they impact our text-to-goal methods and the resulting LCAs. 
All results in this section use the EVA02-E-14+ VLM, which we chose based on its strong classification and retrieval performance~\citep{sun2023eva}, and based on the arguments of~\citet{baumli2023vision} and~\citet{rocamonde2023vision} that larger VLMs engender reward functions that agree better with handcrafted rewards. 
This VLM is an order of magnitude larger than those employed in previous studies of RL with VLM-based rewards. Its use is only practical thanks to our distilled model, as discussed in Appendix~\ref{appendix:lca_training}.
  
We begin by illustrating that VLM scores are highly \emph{multimodal}: they have multiple local maxima when seen as a function of the configuration. Then we discuss situations where VLM scores are \emph{unreliable} and misevaluate configurations. Finally, we show that the score is a highly  \emph{oscillatory} function of configuration: the score can increase then decrease rapidly in response to small configuration changes.

\subsection{Multimodality of VLM Scores}

\begin{figure}[h]
\begin{center}
\centerline{\includegraphics[width=.99\linewidth]{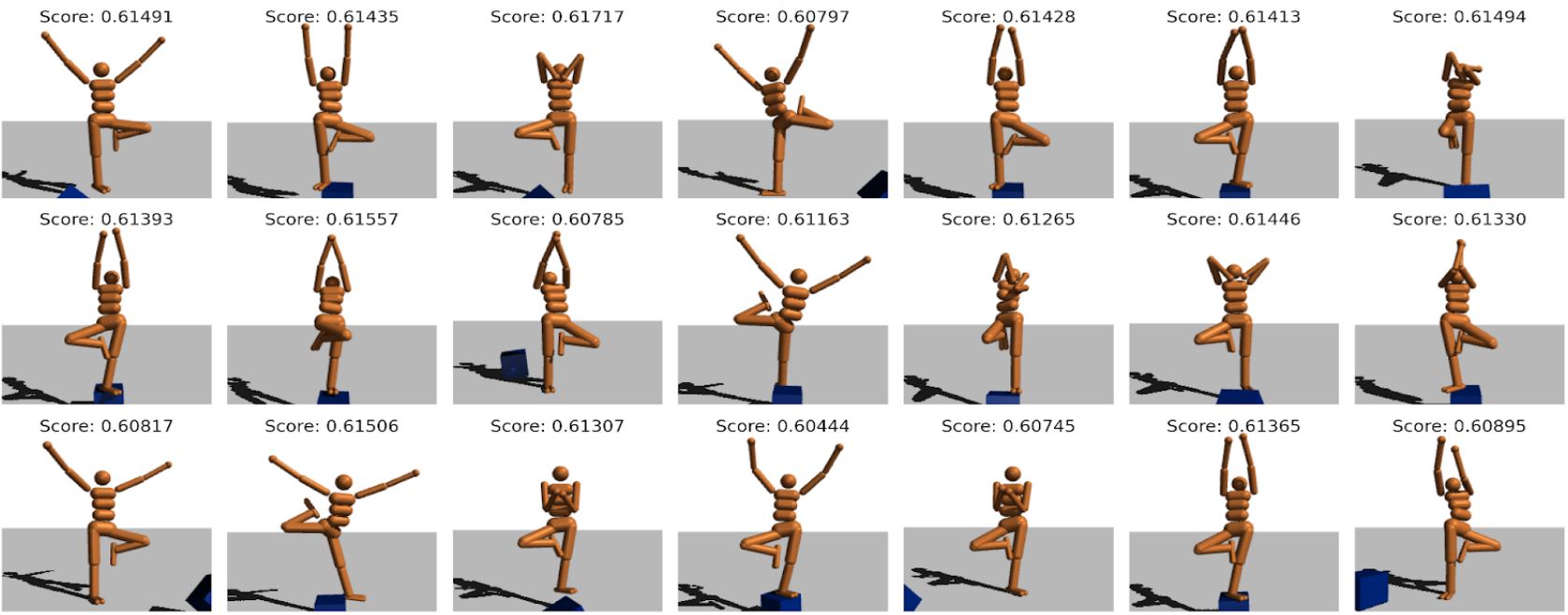}}
\caption{Top 21 configurations retrieved, using multiview scores, from the embedding-diversity dataset for task 39, \emph{``A 3D humanoid model doing a tree yoga pose (Vrkasana)''}.}
\label{fig:tree_yoga_multimodal_example}
\end{center}
\end{figure}

Figure~\ref{fig:tree_yoga_multimodal_example} shows front views of the top-21 configurations retrieved from the embedding-diversity dataset for task 39.  The diversity of these configurations demonstrates that the configuration-text score is a highly multimodal function of the configuration. 
One might provide such a diverse set to a (human or programmatic) user and let them choose which configuration they would like a GCRL agent to aim for, based on criteria other than VLM score alone. For instance, one might choose based on obstacle avoidance and reachability, perhaps using the value function of the GCRL agent, or based on aesthetics or predicted energy consumption. 
This is another potential advantage of the proposed approach over STRL or MTRL agents, as one has no direct control over which configuration these agents will aim for: one only has indirect control through the text, or through the addition of new terms to the agents' reward to limit energy consumption and so forth. 

\begin{figure}
\begin{center}
\centerline{\includegraphics[width=.85\linewidth]{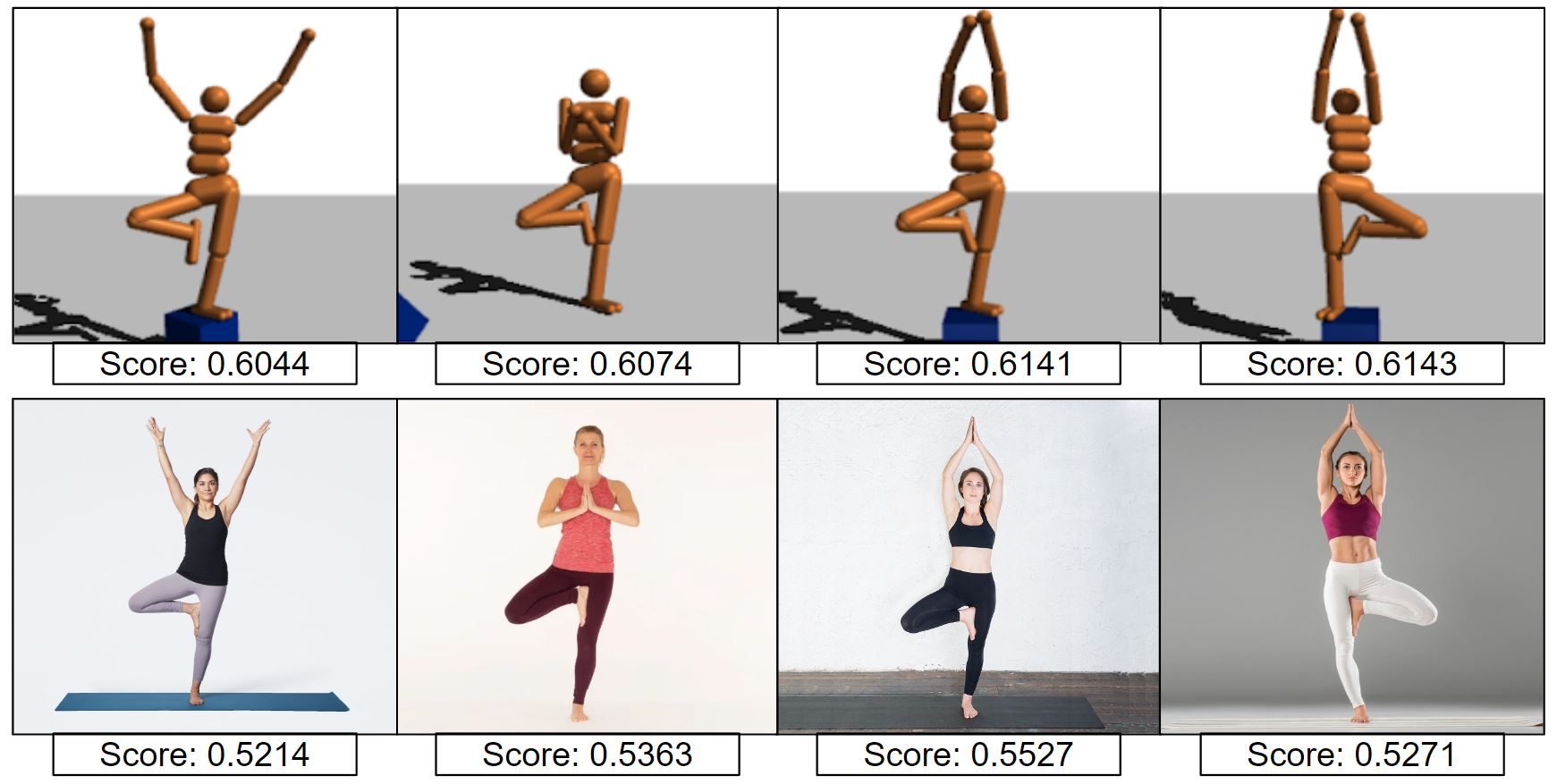}}
\caption{Four of the top-21 retrieved configurations for task 39, \emph{``A 3D humanoid model doing a tree yoga pose (Vrkasana)''} illustrated in  Figure~\ref{fig:tree_yoga_multimodal_example} matched with images from the internet.  The real-world images are evaluated for the text  \emph{``A lady doing a tree yoga pose (Vrkasana)''}.
Although the Humanoid and real-world images are clear examples of the requested pose, the image-text similarity scores are all less than $0.62$, whereas the theoretical maximum of the cosine similarity is $1$. 
}
\label{fig:tree_yoga_variants}
\end{center}
\end{figure}

Figure~\ref{fig:tree_yoga_variants} compares four of the top-21 configurations retrieved for task 39 with images from the internet. The correspondence between these images shows that our text-to-goal method finds diverse configurations that represent naturally occurring variants of a task.

\subsection{Unreliability of VLM Scores}
\label{app:compositionality}

\begin{figure}
\begin{center}
\centerline{\includegraphics[width=.5\linewidth]{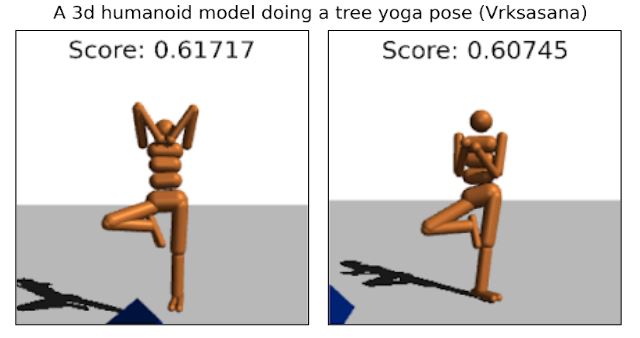}}
\caption{Top-scoring configuration (left) and another top-21 configuration (right) retrieved for task~39, \emph{``A 3D humanoid model doing a tree yoga pose (Vrkasana)''} shown in Figure~\ref{fig:tree_yoga_multimodal_example}. While the image on the right is often preferred by humans, the VLM gives a significantly higher score to the left image.}
\label{fig:tree_yoga_error}
\end{center}
\end{figure}

\paragraph{Misevaluation.} 
Figure~\ref{fig:tree_yoga_error} illustrates that VLMs sometimes misevaluate configurations: configurations that might be subjectively preferred by human viewers occasionally receive substantially lower VLM scores than other configurations.  

\paragraph{Compositionality.}
Today's VLMs are known to struggle with word order and compositionality~\cite{lewis2022does}. 
To understand this limitation, \citet{yuksekgonul2022and} explore the behaviour of VLMs given shuffled text inputs. They argue that this deficiency stems from the training of VLMs with contrastive losses on retrieval tasks, which provides little incentive for the model to learn to encode order and compositionality cues.

\begin{figure}[h]
\vskip 0.1in
\begin{center}
\centerline{\includegraphics[width=.999\linewidth]{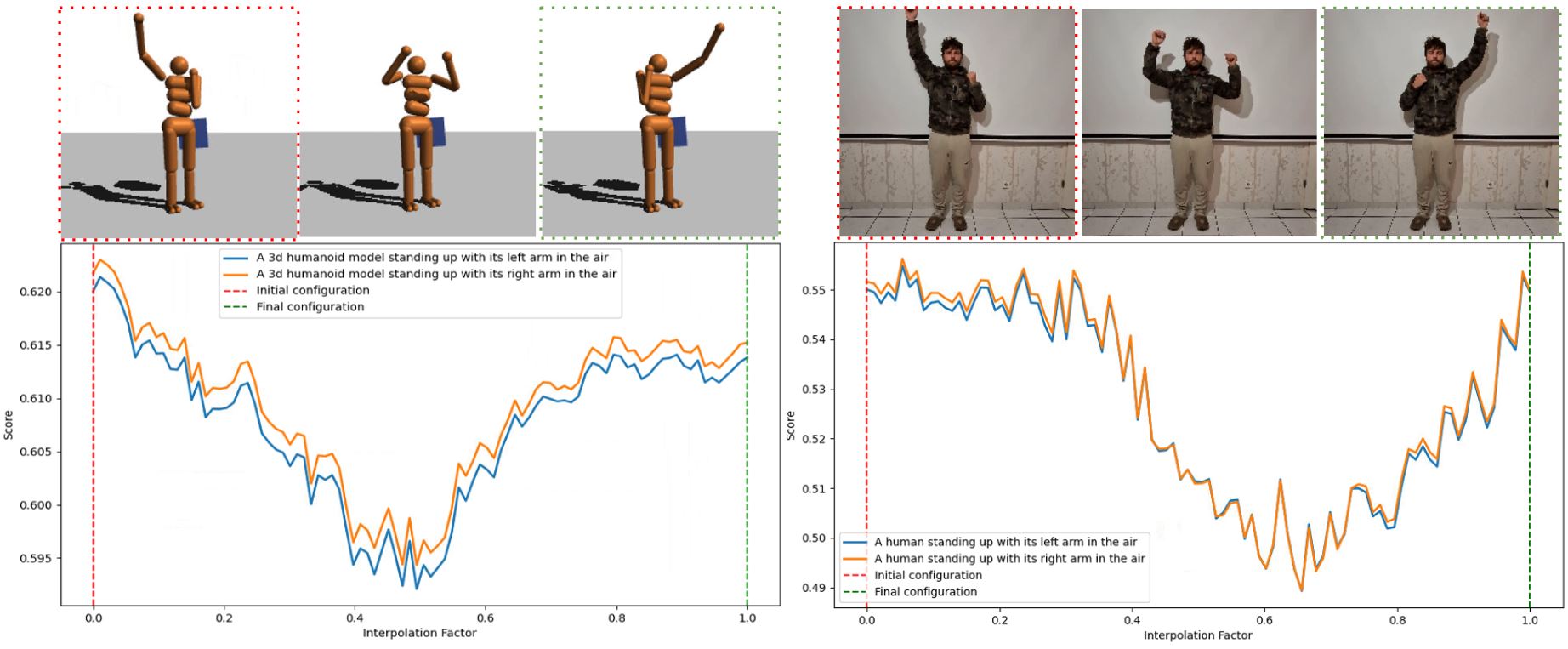}}
\caption{Evolution of VLM score for the text \emph{``A 3D humanoid model standing up with left hand in the air''} and \emph{``A 3D humanoid model standing up with right hand in the air''} on both rendered and real images interpolating between the two described positions.}
\label{fig:vlm_compositionality}
\end{center}
\end{figure}

Figure~\ref{fig:vlm_compositionality} illustrates the failure of VLMs to distinguish between `left' and `right'. To generate the left-hand plot, we interpolated configurations linearly between a configuration with the right hand in the air and a configuration with the left hand in the air. As we move through these interpolated configurations, the VLM scores for \textit{``A 3D humanoid model standing up with \textbf{left} hand in the air''} and for \textit{``A 3D humanoid model standing up with \textbf{right} hand in the air''} are highly correlated. This failure is not due to our choice of rendering functions: we observe the same phenomenon on the right-hand side of Figure~\ref{fig:vlm_compositionality}, which was generated from a 100-frame real-world video. Rather, this failure is due to the VLM's text embedding: the cosine similarity between the text embeddings of these two sentences is $0.9978$, thus the VLM scores for these two sentences are almost identical for \emph{any} image embedding. 

These failings, along with other known weaknesses such as an inability to count small numbers of objects, limit the diversity of tasks that LCAs derived from VLMs can currently tackle.
There have been recent efforts to overcome such limitations,  for instance by leveraging reconstruction losses~\cite{tschannen2023image}, generating hard negative image-text pairs~\cite{radenovic2023filtering} and building new benchmarks~\cite{hsieh2023sugarcrepe}.
Such work promises to enable future LCAs to successfully complete tasks involving compositional relationships between diverse sets of objects.

\clearpage 

\subsection{Oscillatory Nature of VLM Scores}
\label{sec:oscillatory-nature}
\begin{figure}[h]
\begin{center}
\centerline{\includegraphics[width=.65\linewidth]{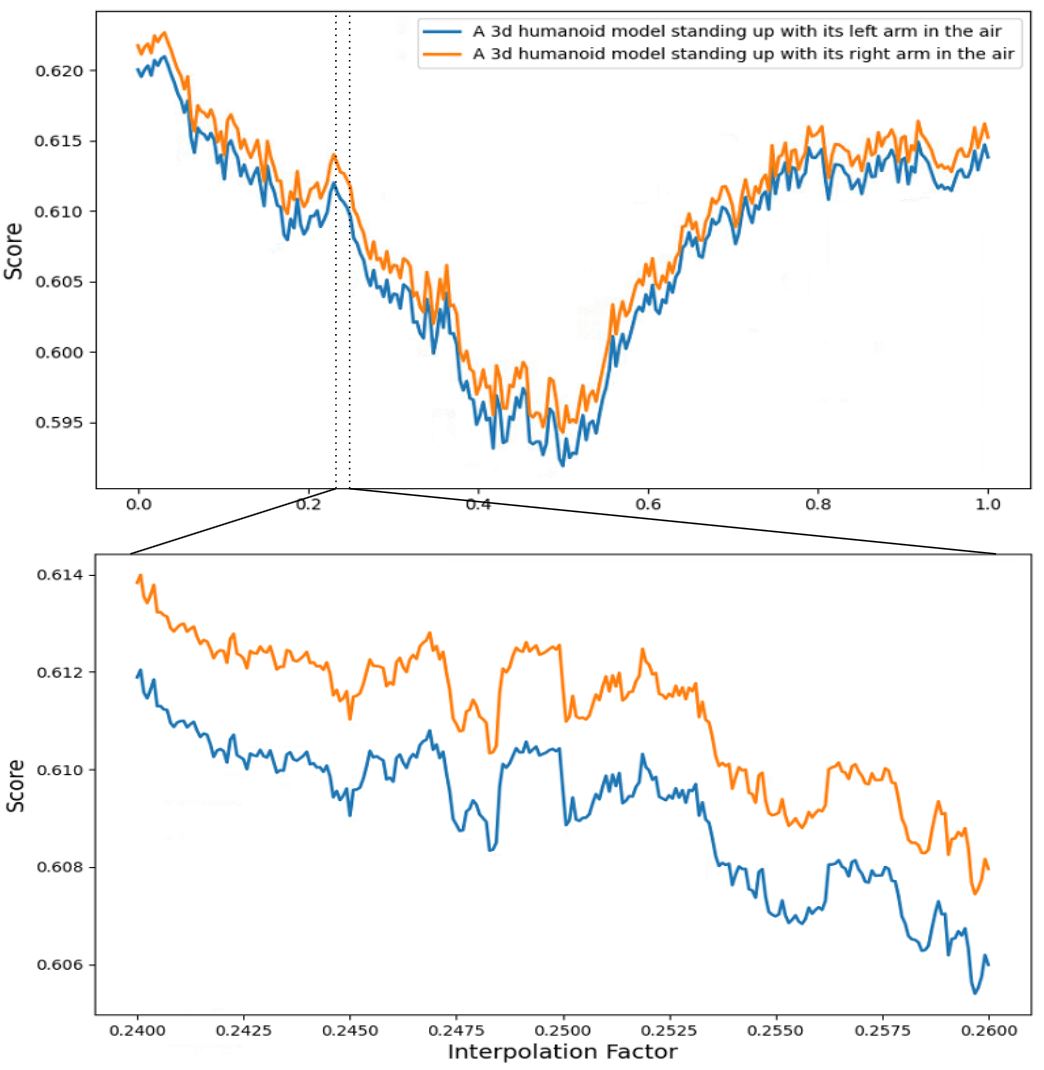}}
\caption{Zoomed-in and more densely sampled version of the left part of Figure~\ref{fig:vlm_compositionality}, illustrating the oscillatory nature of the configuration-text score as a function of configuration.}
\label{fig:noisiness}
\end{center}
\end{figure}

VLM scores vary rapidly with configuration, exhibiting multiple (aperiodic) oscillations as the configuration varies by a small amount.  Figure~\ref{fig:noisiness} shows a zoomed-in and more densely sampled version of Figure~\ref{fig:vlm_compositionality} (left). As the interpolation factor goes from 0.0 to 1.0, the angles made by the links of the Humanoid vary by at most $120^\circ$. Thus, a change in interpolation factor from 0.2400 to 0.2450 corresponds to a change in angle of less than $1^\circ$. Yet, over this small range of interpolation factors, at least 7 local maxima of the VLM score (as a function of the interpolation factor) are visible in the lower part of Figure~\ref{fig:noisiness}. 
Therefore, even if the configuration-text score could be differentiated with respect to the configuration, for instance using finite differences or differentiable rendering~\citep{kato2020differentiable}, we argue that the derivatives would be of less value for finetuning than the derivatives of a smoother distilled model. 

\clearpage
\section{Task Descriptions}
\label{appendix:tasks}
This section describes the process used to generate the 256 tasks and then lists those tasks. The tasks are descriptions of desired environment configurations in natural language, which we use to evaluate text-to-goal methods and LCAs. When specifying a task to the agents, we prepend the tasks with \textit{``A 3D humanoid model''}. This choice has been motivated by the remark in the Appendix E.1 of \citet{radford2021learning} indicating that prepending \textit{``a photo of''}
to the description of each image ``boosts CLIP's zero-shot R@1 performance between 1 and 2 points.''

\paragraph{Task-generation process.}
The tasks were generated as follows. We applied ChatGPT~\citep{achiam2023gpt} with a prompt requesting creative descriptions of a static scene containing a 3D humanoid and a cube, and giving 30 example task descriptions. We deduplicated the resulting tasks and eliminated those texts that referred to objects other than the cube. Finally, we manually edited about 10\% of the tasks, as they did not make sense or could have been expressed more naturally, for instance:
\begin{itemize}
\item ChatGPT proposed the task \textit{``A 3D humanoid model seated on the floor, one knee raised, chin resting on the knee, echoing the pose of `The Little Mermaid’ statue’’}. This does not make sense as the Little Mermaid statue does not have its chin on its knee. Therefore, it was edited to \textit{``A 3D humanoid model seated on the floor, with its knees to the side, echoing the pose of `The Little Mermaid’ statue’’}.
\item ChatGPT proposed the task \textit{``A 3D humanoid model doing a sidekick stance, like a martial artist’’}. A “sidekick” is distinct from a “side kick”. Moreover, while there are many named stances in the martial arts such as “horse stance” and “crane stance”, we did not identify one named “side kick stance”. Therefore, this was edited to \textit{``A 3D humanoid model doing a side kick, like a martial artist’’}.
\end{itemize}
There was no filtering to favour tasks that on which the proposed methods perform well.

\paragraph{List of tasks.} The tasks were divided into two sets, with one set used for the training and the other set for the testing of two MTRL agents. The task IDs of these sets are given below, followed by the task descriptions. 

Task set 1 IDs: 1, 2, 5, 8, 9, 13, 14, 16, 17, 18, 19, 20, 21, 24, 25, 26, 29, 32, 33, 34, 35, 36, 38, 39, 43, 44, 47, 50, 51, 53, 54, 55, 56, 58, 60, 61, 63, 66, 68, 69, 73, 74, 76, 77, 80, 81, 83, 85, 86, 87, 88, 90, 93, 97, 98, 99, 102, 103, 104, 105, 107, 109, 112, 114, 115, 117, 119, 121, 122, 124, 126, 129, 131, 138, 139, 140, 142, 143, 146, 147, 150, 152, 153, 154, 158, 159, 161, 163,  164, 169, 170, 172, 174, 176, 178, 183, 184, 189, 190, 192, 193, 195, 198, 201, 203, 204, 205, 206, 209, 212, 215, 216, 219, 223, 225, 227, 228, 233, 234, 236, 237, 238, 242, 245, 246, 250, 254, 255.

Task set 2 IDs: 0, 3, 4, 6, 7, 10, 11, 12, 15, 22, 23, 27, 28, 30, 31, 37, 40, 41, 42, 45, 46, 48, 49, 52, 57, 59, 62, 64, 65, 67, 70, 71, 72, 75, 78, 79, 82, 84, 89, 91, 92, 94, 95, 96, 100, 101, 106, 108, 110, 111, 113, 116, 118, 120, 123, 125, 127, 128, 130, 132, 133, 134, 135, 136, 137, 141, 144, 145, 148, 149, 151, 155, 156, 157, 160, 162, 165, 166, 167, 168, 171, 173, 175, 177, 179, 180, 181, 182, 185, 186, 187, 188, 191, 194, 196, 197, 199, 200, 202, 207, 208, 210, 211, 213, 214, 217, 218, 220, 221, 222, 224, 226, 229, 230, 231, 232, 235, 239, 240, 241, 243, 244, 247, 248, 249, 251, 252, 253.

{\small \begin{enumerate}[label=Task \arabic*:, leftmargin=*, align=left, itemsep=-0.7mm, parsep=0mm]
    \item \textit{``balancing on one leg''}
    \item \textit{``balancing on one leg and waving''}
    \item \textit{``balancing on one leg with both hands up''}
    \item \textit{``balancing on one leg with folded arms''}
    \item \textit{``balancing on one leg with the other leg bent''}
    \item \textit{``bowing''}
    \item \textit{``crawling''}
    \item \textit{``crouching, hands on the floor between feet, head tilted up looking forward''}
    \item \textit{``crouching, one hand on the ground for support, looking over the shoulder''}
    \item \textit{``crouching, one hand touching the ground, and the other hand shielding its eyes as if looking for something small''}
    \item \textit{``curtsying''}
    \item \textit{``demonstrating a bridge yoga pose''}
    \item \textit{``demonstrating a child yoga pose''}
    \item \textit{``demonstrating a cow yoga pose''}
    \item \textit{``demonstrating a downward-facing dog yoga pose''}
    \item \textit{``demonstrating a happy baby yoga pose''}
    \item \textit{``demonstrating a headstand yoga pose''}
    \item \textit{``demonstrating a lotus yoga pose''}
    \item \textit{``demonstrating a lunge yoga pose''}
    \item \textit{``demonstrating a mountain yoga pose''}
    \item \textit{``demonstrating a triangle yoga pose''}
    \item \textit{``demonstrating a warrior yoga pose''}
    \item \textit{``demonstrating a wheel yoga pose''}
    \item \textit{``doing a ``talk to the hand'' pose''}
    \item \textit{``doing a `victory' pose (V shape with arms)''}
    \item \textit{``doing a bird dog pose''}
    \item \textit{``doing a box step-up''}
    \item \textit{``doing a camel yoga pose''}
    \item \textit{``doing a cobra yoga pose''}
    \item \textit{``doing a dab pose''}
    \item \textit{``doing a goosestep''}
    \item \textit{``doing a half-moon pose''}
    \item \textit{``doing a handstand''}
    \item \textit{``doing a plank''}
    \item \textit{``doing a push-up with both feet on the box''}
    \item \textit{``doing a side kick, like a martial artist''}
    \item \textit{``doing a side plank''}
    \item \textit{``doing a taekwondo back kick''}
    \item \textit{``doing a tree yoga pose (Vrksasana)''}
    \item \textit{``doing a warrior 3 yoga pose (Virabhadrasana III)''}
    \item \textit{``doing an arabesque''}
    \item \textit{``doing an arabesque penchée''}
    \item \textit{``doing an extended hand-to-big-toe pose''}
    \item \textit{``doing crunches''}
    \item \textit{``doing push-ups''}
    \item \textit{``doing squats''}
    \item \textit{``doing the front splits''}
    \item \textit{``doing the limbo''}
    \item \textit{``doing the side splits''}
    \item \textit{``holding a blue box with its arms''}
    \item \textit{``holding the blue box so that one corner is in contact with the floor''}
    \item \textit{``in a ``Saturday Night Fever'' dance pose''}
    \item \textit{``in a `superhero' stance, hands on hips, chest out, looking confident''}
    \item \textit{``in a Bruce Lee fighting stance''}
    \item \textit{``in a Namaste pose''}
    \item \textit{``in a boxing guard pose''}
    \item \textit{``in a crouching position, hands placed on the floor between feet, looking down''}
    \item \textit{``in a fencing `en garde' position''}
    \item \textit{``in a front double biceps pose''}
    \item \textit{``in a front lat spread pose''}
    \item \textit{``in a pose reminiscent of Michelangelo's `The Crouching Boy{'}''}
    \item \textit{``in a side chest pose''}
    \item \textit{``in a side triceps pose''}
    \item \textit{``in a sprint starting pose''}
    \item \textit{``in the middle of a `moonwalk' dance step''}
    \item \textit{``jumping''}
    \item \textit{``kneeling''}
    \item \textit{``kneeling and waving''}
    \item \textit{``kneeling down and pretending to propose with an imaginary ring''}
    \item \textit{``kneeling with both arms in the air''}
    \item \textit{``kneeling with both hands up''}
    \item \textit{``kneeling with folded arms''}
    \item \textit{``kneeling with its hands behind its back''}
    \item \textit{``kneeling with its hands on its head''}
    \item \textit{``kneeling, one hand on the blue box, looking up as if observing something above''}
    \item \textit{``kneeling, one hand on the blue box, the other hand raised above the head, palm facing forward''}
    \item \textit{``kneeling, one hand resting on an upright blue box, the other hand on the hip''}
    \item \textit{``leaning against an imaginary wall with one arm raised''}
    \item \textit{``leaning backward, holding its ankle, in a quad stretch pose''}
    \item \textit{``looking for someone''}
    \item \textit{``lying down on its side''}
    \item \textit{``lying down on the floor''}
    \item \textit{``lying down on the floor and waving''}
    \item \textit{``lying down on the floor with both hands up''}
    \item \textit{``lying down on the floor with folded arms''}
    \item \textit{``lying down on the floor with hands on its head''}
    \item \textit{``lying face down with arms and legs stretched out, like a     `starfish{'}''}
    \item \textit{``lying face up, hands under the head as a pillow, one leg crossed over the other''}
    \item \textit{``lying flat on the back, both legs raised straight up''}
    \item \textit{``lying flat on the back, one leg raised straight up''}
    \item \textit{``lying on its back with knees bent, in a sit-up position''}
    \item \textit{``lying on its back with one knee bent''}
    \item \textit{``lying on its back, forming an `X' shape with spread arms and legs''}
    \item \textit{``lying on its side, with its head propped up by its hand, and with the blue box behind as a backdrop''}
    \item \textit{``lying on its stomach, propping up its head with its hands''}
    \item \textit{``lying on the back, arms and legs slightly spread apart, in a relaxed pose''}
    \item \textit{``lying on the back, hands under the head, one leg crossed over the other, ankle on knee''}
    \item \textit{``lying on the back, imitating a snow angel movement, arms and legs moving outward and inward''}
    \item \textit{``lying on the back, in a `fainting' pose reminiscent of Victorian paintings''}
    \item \textit{``lying on the back, one hand on the forehead, the other resting on the stomach''}
    \item \textit{``lying on the back, with arms and legs wrapped around the blue box''}
    \item \textit{``lying on the back, with the blue box placed under the head like a pillow''}
    \item \textit{``lying on the back, with the blue box resting on the stomach''}
    \item \textit{``lying on the side, hugging the blue box, with one leg bent''}
    \item \textit{``lying on the side, propped up on one elbow''}
    \item \textit{``lying on the side, propped up on one elbow, legs crossed at the ankles, in a classic `Reclining Venus' pose''}
    \item \textit{``lying on the side, propping head up with one hand, the other arm resting along the body''}
    \item \textit{``lying on the stomach, arms and legs outstretched, face turned to one side''}
    \item \textit{``lying on the stomach, arms folded under the chin, feet dangling in the air''}
    \item \textit{``lying on the stomach, chin resting on hands, legs crossed at the ankles behind''}
    \item \textit{``meditating''}
    \item \textit{``mimicking the ``Karate Kid'' crane kick pose''}
    \item \textit{``mimicking the ``Rodin's The Thinker'' pose''}
    \item \textit{``mimicking the iconic `Statue of Liberty' pose''}
    \item \textit{``mimicking the pose of Leonardo da Vinci's `The Vitruvian Man{'}''}
    \item \textit{``on all fours, one arm extended forward, the opposite leg extended back''}
    \item \textit{``on one knee, other leg bent with foot flat on the floor, and hands on the raised knee''}
    \item \textit{``playing an imaginary violin''}
    \item \textit{``praying''}
    \item \textit{``pretending to be a conductor leading an orchestra''}
    \item \textit{``pretending to lift a heavy weight overhead, like a weightlifter''}
    \item \textit{``pretending to shoot a basketball''}
    \item \textit{``prostrating itself''}
    \item \textit{``pulling an imaginary bow and arrow''}
    \item \textit{``running''}
    \item \textit{``running in the style of Naruto ninjas''}
    \item \textit{``saluting like a soldier''}
    \item \textit{``seated on the floor, bending forward towards the feet, arms extended forward''}
    \item \textit{``seated on the floor, one hand resting on the blue box, looking thoughtful''}
    \item \textit{``seated on the floor, with its knees to the side, echoing the pose of `The Little Mermaid' statue''}
    \item \textit{``seated on the ground, leaning sideways with one hand supporting the body''}
    \item \textit{``seated, leaning against the blue box, legs crossed, one hand supporting the head''}
    \item \textit{``seated, leaning forward with its elbows on its knees and its chin resting on its hands''}
    \item \textit{``seated, legs stretched out, leaning back on its hands, gazing upward''}
    \item \textit{``singing''}
    \item \textit{``sitting back on heels, hands resting on thighs''}
    \item \textit{``sitting cross-legged on the floor, hands resting on the knees''}
    \item \textit{``sitting cross-legged on the floor, one hand on the blue box, the other hand gesturing in mid-air''}
    \item \textit{``sitting on the blue box, hands on knees, looking straight ahead''}
    \item \textit{``sitting on the blue box, leaning forward with its chin resting on hands''}
    \item \textit{``sitting on the blue box, leaning forward, in a pose reminiscent of `The Old Guitarist' by Picasso''}
    \item \textit{``sitting on the blue box, legs crossed, one hand on the knee, the other resting on the box''}
    \item \textit{``sitting on the blue box, legs dangling''}
    \item \textit{``sitting on the floor''}
    \item \textit{``sitting on the floor and waving''}
    \item \textit{``sitting on the floor in a mermaid pose''}
    \item \textit{``sitting on the floor with both hands up''}
    \item \textit{``sitting on the floor with folded arms''}
    \item \textit{``sitting on the floor with straight legs''}
    \item \textit{``sitting on the floor with straight legs and both arms in the air''}
    \item \textit{``sitting on the floor with straight legs and one arm in the air''}
    \item \textit{``sitting on the floor with straight legs and touching its head''}
    \item \textit{``sitting on the floor with the blue box in front, arms resting on the box''}
    \item \textit{``sitting on the floor, back straight, with its hands resting on its knees, as if meditating''}
    \item \textit{``sitting on the floor, body twisted to one side, looking over the shoulder''}
    \item \textit{``sitting on the floor, hugging the blue box to the chest''}
    \item \textit{``sitting on the floor, imitating Rodin's `The Thinker' pose, with chin resting on hand''}
    \item \textit{``sitting on the floor, leaning against the blue box with arms crossed''}
    \item \textit{``sitting on the floor, leaning back on hands with legs stretched out front''}
    \item \textit{``sitting on the floor, leaning on one hand, legs casually stretched out, in a relaxed Bohemian style''}
    \item \textit{``sitting on the floor, one arm resting on the raised knee, the other hand supporting the body behind''}
    \item \textit{``sitting on the floor, one hand on the blue box, the other raised as if it wished to ask a teacher a question''}
    \item \textit{``sitting on the ground, back against the blue box, legs stretched out''}
    \item \textit{``sitting on the ground, leaning back on its hands, with its legs bent, in a relaxed `La Dolce Vita' style''}
    \item \textit{``sitting on top of a box''}
    \item \textit{``sitting with legs crossed and one hand on the chin''}
    \item \textit{``sitting with legs crossed, one hand under the chin, and a pensive look, emulating `The Thinker{'}''}
    \item \textit{``sitting with legs tucked under, resting hands on knees''}
    \item \textit{``sitting with legs wide apart, leaning forward with arms extended between legs''}
    \item \textit{``sitting with the blue box placed on its lap, hands resting on the box''}
    \item \textit{``sitting, knees drawn up, chin resting on its knees, arms wrapped around its legs''}
    \item \textit{``sitting, leaning forward with its elbows on its knees, gazing downward thoughtfully''}
    \item \textit{``squatting next to the blue box, resting one elbow on the box''}
    \item \textit{``squatting with arms extended forward''}
    \item \textit{``squatting, arms wrapped around the knees, head bowed''}
    \item \textit{``squatting, with its hands touching the ground, in a sprinter's starting position''}
    \item \textit{``standing and gazing upwards, hands clasped behind the back''}
    \item \textit{``standing in a `goalkeeper' stance, arms outstretched to the sides, legs slightly bent''}
    \item \textit{``standing in a T-pose (arms extended horizontally)''}
    \item \textit{``standing in a contrapposto pose, weight shifted onto one leg, reminiscent of classical Greek statues''}
    \item \textit{``standing in a surfing stance, pretending to ride a wave''}
    \item \textit{``standing in the `Atlas' pose, as if holding an invisible globe on its shoulders''}
    \item \textit{``standing in the `Discobolus' (Discus Thrower) pose, as if ready to throw an imaginary discus''}
    \item \textit{``standing knock kneed and pigeon toed''}
    \item \textit{``standing like a captain, with hands akimbo (on the hips)''}
    \item \textit{``standing like a mime artist pulling an imaginary rope''}
    \item \textit{``standing on tiptoes, arms reaching up as if trying to touch the ceiling''}
    \item \textit{``standing up''}
    \item \textit{``standing up and reaching down to touch the toes''}
    \item \textit{``standing up and waving''}
    \item \textit{``standing up in Usain Bolt's celebration pose''}
    \item \textit{``standing up on top of the blue box''}
    \item \textit{``standing up with bent knees''}
    \item \textit{``standing up with both hands up''}
    \item \textit{``standing up with folded arms''}
    \item \textit{``standing up with hands behind its back''}
    \item \textit{``standing up with its hands on its head''}
    \item \textit{``standing up with its hands on its hips''}
    \item \textit{``standing up with its hands touching above its head''}
    \item \textit{``standing up with one hand in the air and one hand on its head''}
    \item \textit{``standing up with one hand on its hip and one hand on its head''}
    \item \textit{``standing with arms crossed over the chest''}
    \item \textit{``standing with arms extended upward, forming a `Y' shape''}
    \item \textit{``standing with arms folded behind the head, elbows out, in a relaxed stance''}
    \item \textit{``standing with arms outstretched, mimicking the Christ the Redeemer statue in Rio de Janeiro''}
    \item \textit{``standing with arms raised high, one foot forward, in a ballet `Arabesque' position''}
    \item \textit{``standing with arms wide open, as if ready to give a hug''}
    \item \textit{``standing with back arched and hands reaching towards the sky''}
    \item \textit{``standing with feet apart, hands clasped together in front at waist level''}
    \item \textit{``standing with hands together in front, as if holding an invisible object delicately''}
    \item \textit{``standing with one arm bent, fist near the cheek, emulating the iconic `Rosie the Riveter' pose''}
    \item \textit{``standing with one arm extended horizontally, pointing to the side''}
    \item \textit{``standing with one arm extended outward as if signaling `stop{'}''}
    \item \textit{``standing with one arm extended upwards, pointing''}
    \item \textit{``standing with one foot on the blue box, hands on hips, looking down at the box''}
    \item \textit{``standing with one hand extended as if presenting the blue box''}
    \item \textit{``standing with one hand on the chest and the other hand raised in a `stop' gesture''}
    \item \textit{``standing with one hand on the chin, in a thoughtful pose''}
    \item \textit{``standing with one leg crossed over the other, arms hanging loosely at the sides''}
    \item \textit{``standing, arms akimbo, with a tilted head, emulating the iconic `Superman' stance''}
    \item \textit{``standing, arms interlocked above the head, stretching upward''}
    \item \textit{``standing, arms outstretched to the sides, palms facing forward, in a welcoming stance''}
    \item \textit{``standing, arms reaching out to the sides at shoulder height, palms up''}
    \item \textit{``standing, bending forward at the waist, hands reaching towards the feet''}
    \item \textit{``standing, bending one knee, foot on the blue box, mimicking a stretching pose''}
    \item \textit{``standing, bending slightly forward with hands on its lower back''}
    \item \textit{``standing, bending slightly to one side, one hand on the hip, the other hand touching the side of the head''}
    \item \textit{``standing, body twisted, one arm reaching across to touch the opposite shoulder''}
    \item \textit{``standing, head tilted back, hands clasped behind the neck''}
    \item \textit{``standing, head tilted to one side, hands forming a heart shape at chest level''}
    \item \textit{``standing, imitating the `Vitruvian Man' by Leonardo da Vinci, with arms and legs outstretched''}
    \item \textit{``standing, leaning slightly forward, as if peering into the distance''}
    \item \textit{``standing, mimicking the act of holding a large object above the head with both hands''}
    \item \textit{``standing, mimicking the act of looking through an imaginary telescope''}
    \item \textit{``standing, one arm bent with hand on waist, the other arm extended, palm up''}
    \item \textit{``standing, one arm draped over the blue box, the other hand touching the chin, in a contemplative pose''}
    \item \textit{``standing, one arm raised, hand shading the eyes as if looking into the distance''}
    \item \textit{``standing, one foot slightly forward, hands clasped at waist level, looking serene''}
    \item \textit{``standing, one hand on heart, the other extended outward in a welcoming gesture''}
    \item \textit{``standing, one hand on the blue box, as if leaning on it for support''}
    \item \textit{``standing, one hand touching the chin, the other arm hanging by the side''}
    \item \textit{``standing, one hand touching the temple as if deep in thought''}
    \item \textit{``standing, one leg raised and resting on the blue box, hands on hips, looking at the raised foot''}
    \item \textit{``standing, one leg slightly raised, foot resting on the blue box, hands on hips''}
    \item \textit{``standing, tilting the head to one side, hands relaxed by the sides''}
    \item \textit{``stretching its glutes on the floor''}
    \item \textit{``stretching its glutes while standing up''}
    \item \textit{``stretching its quads on the floor''}
    \item \textit{``stretching its quads while standing up''}
    \item \textit{``throwing a high kick''}
    \item \textit{``throwing an uppercut''}
    \item \textit{``touching its knees with its hands''}
    \item \textit{``walking''}
    \item \textit{``walking like an Egyptian''}
    \item \textit{``walking with both arms in the air''}
    \item \textit{``walking with both hands touching its head''}
\end{enumerate}} 
\clearpage

\section{Generated and Reached Configurations}
\label{sec:goal-configs}

This section provides frontal views of the configurations generated by our text-to-goal methods and then the configurations reached by our LCAs and the baseline LCAs for each of the 256 tasks listed in  Appendix~\ref{appendix:tasks}. 

\subsection{Generated Goal Configurations}

The following figures show frontal views of the highest-scoring configurations for each of the proposed text-to-goal methods for both single and multiview scores based on the EVA02-E-14+ VLM.

\begin{figure}
\begin{center}
\centerline{\includegraphics[width=.9999\linewidth]{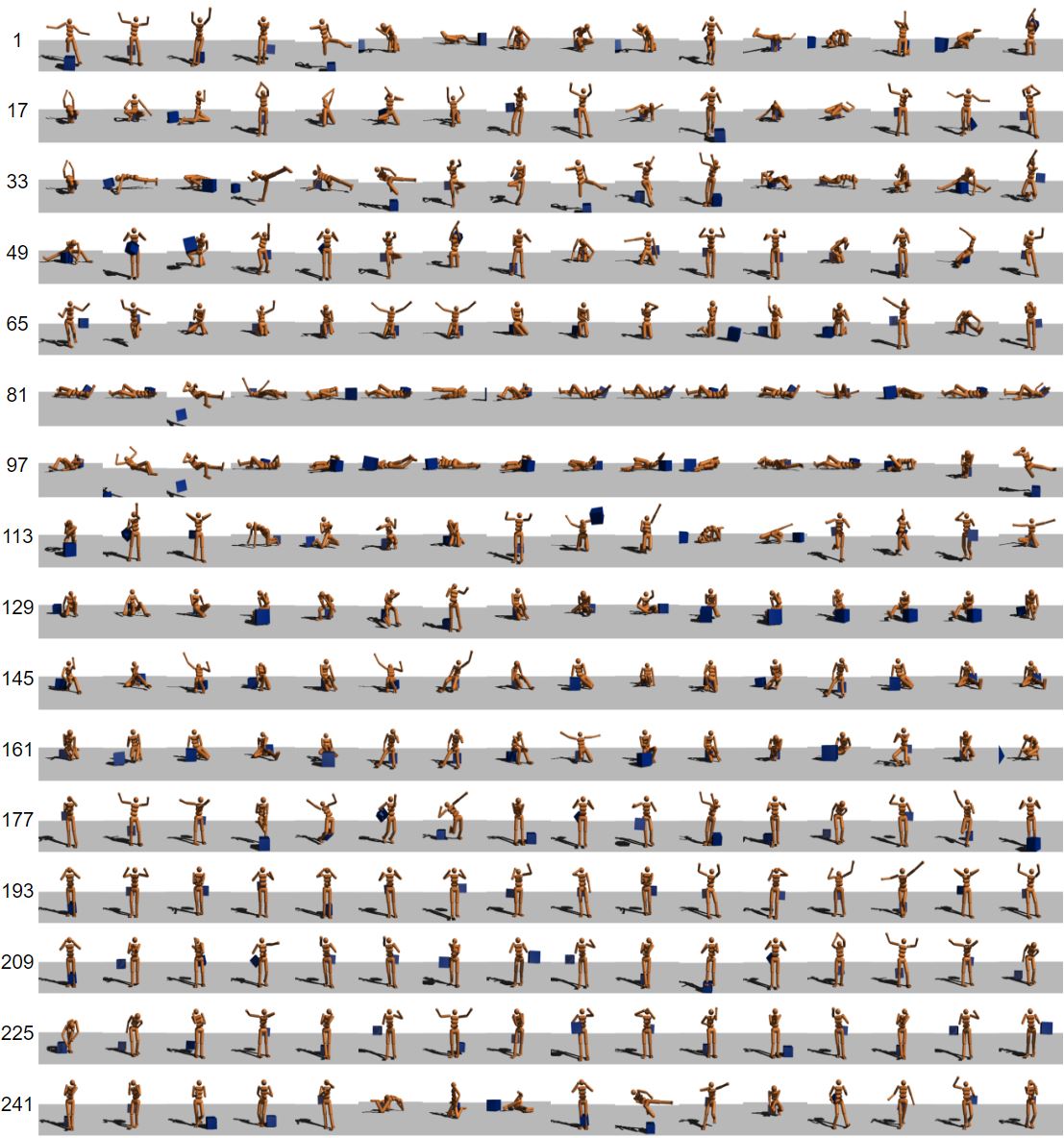}}
\caption{Front views of the highest-scoring configurations retrieved from the random-policy dataset with EVA02-E-14+, using single-view evaluations.}
\label{fig:retrieved_rand_1v}
\end{center}
\end{figure}

\begin{figure}
\begin{center}
\centerline{\includegraphics[width=.9999\linewidth]{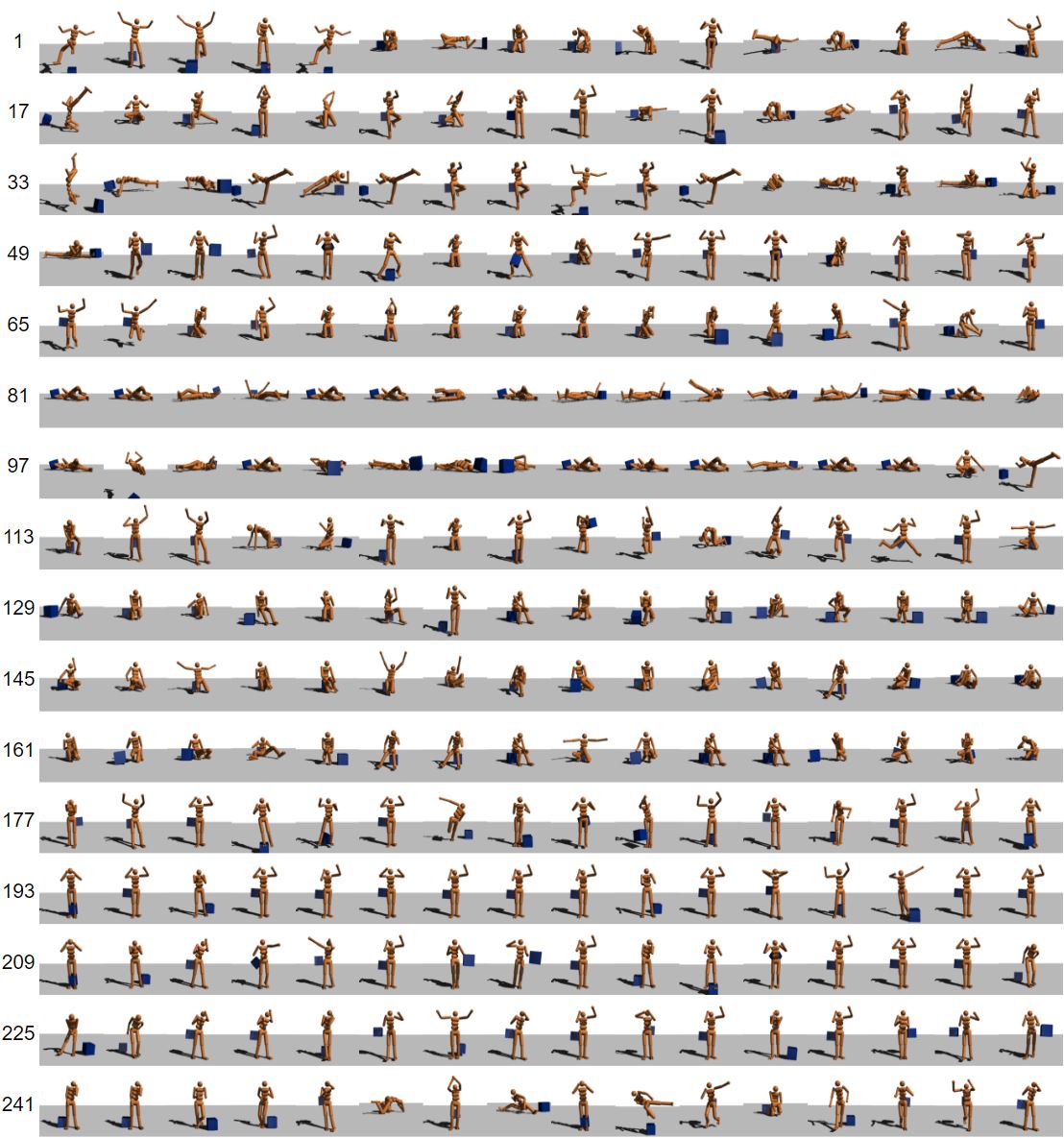}}
\caption{Front views of the highest-scoring configurations retrieved from the random-policy dataset with EVA02-E-14+, using multiview evaluations.}
\label{fig:retrieved_rand_3v}
\end{center}
\end{figure}

\begin{figure}
\begin{center}
\centerline{\includegraphics[width=.9999\linewidth]{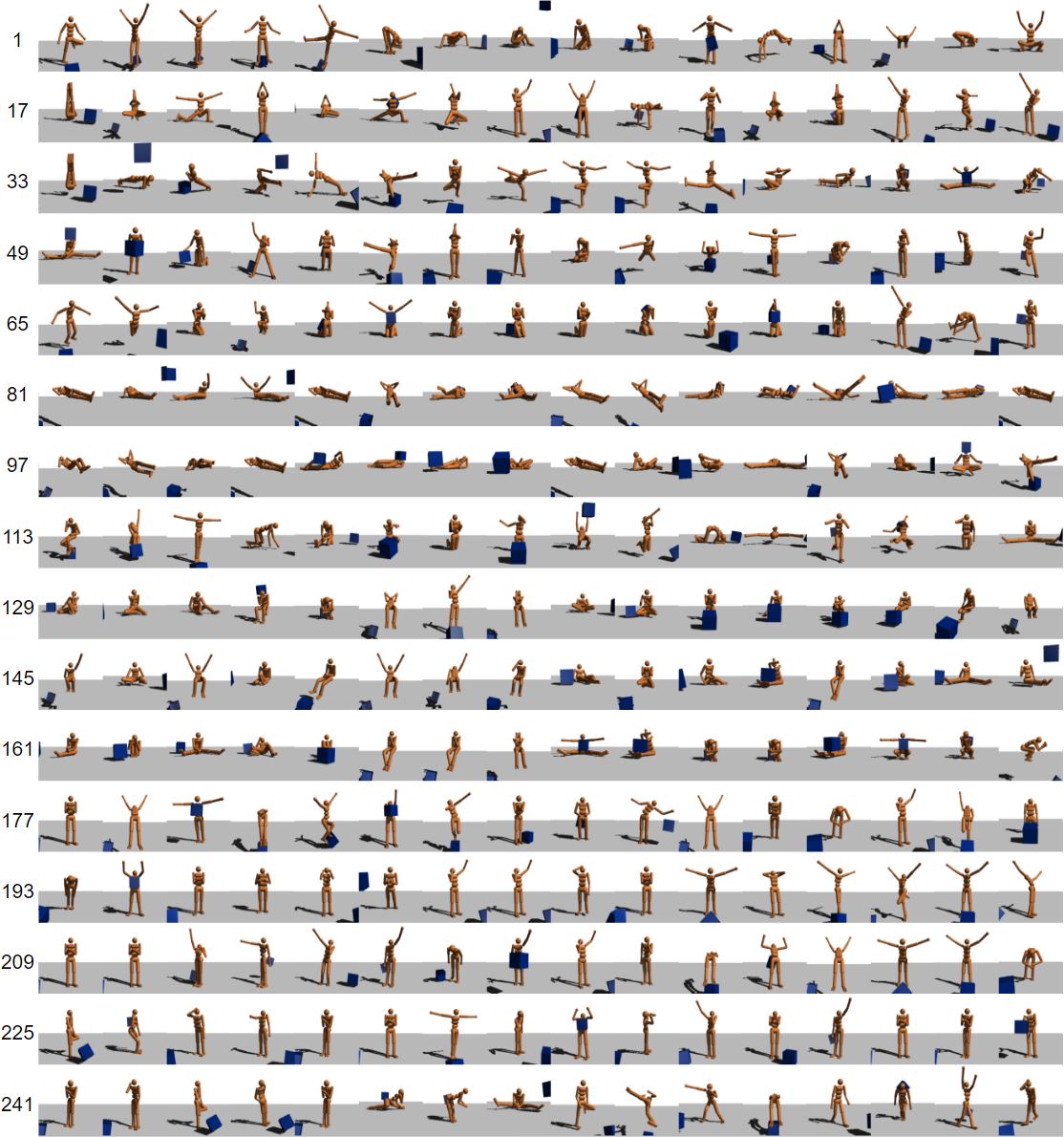}}
\caption{Front views of the highest-scoring configurations retrieved from the single-view embedding-diversity dataset with EVA02-E-14+, using single-view evaluations.}
\label{fig:retrieved_cov_1v}
\end{center}
\end{figure}

\begin{figure}
\begin{center}
\centerline{\includegraphics[width=.9999\linewidth]{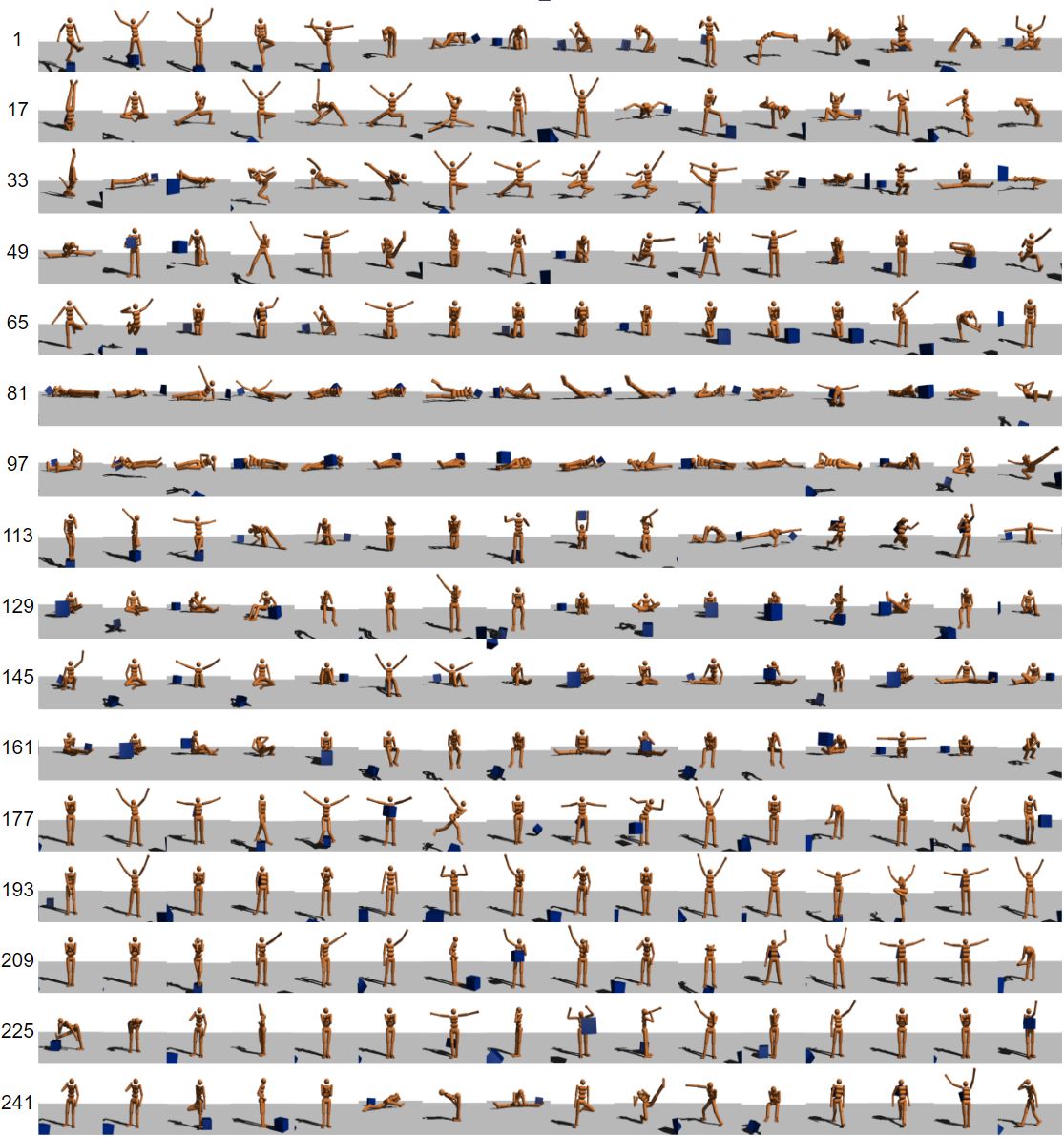}}
\caption{Front views of the highest-scoring configurations retrieved from the single-view embedding-diversity dataset with EVA02-E-14+, using multiview evaluations.}
\label{fig:retrieved_cov_3v}
\end{center}
\end{figure}

\begin{figure}
\begin{center}
\centerline{\includegraphics[width=.9999\linewidth]{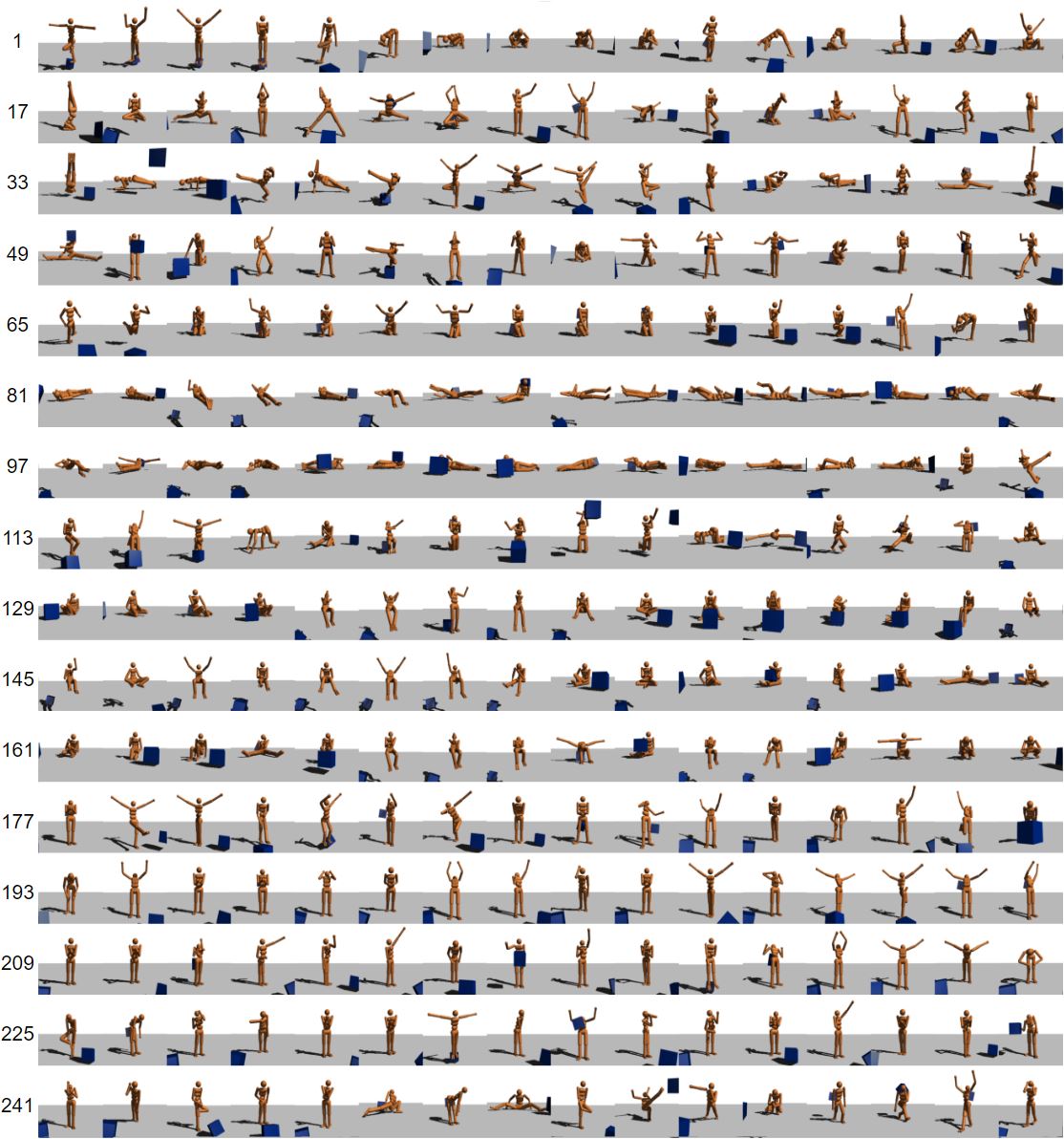}}
\caption{Front views of configurations retrieved from the single-view embedding-diversity dataset, then finetuned using a single-view distilled model.}
\label{fig:finetuned_cov_1v}
\end{center}
\end{figure}

\begin{figure}
\begin{center}
\centerline{\includegraphics[width=.9999\linewidth]{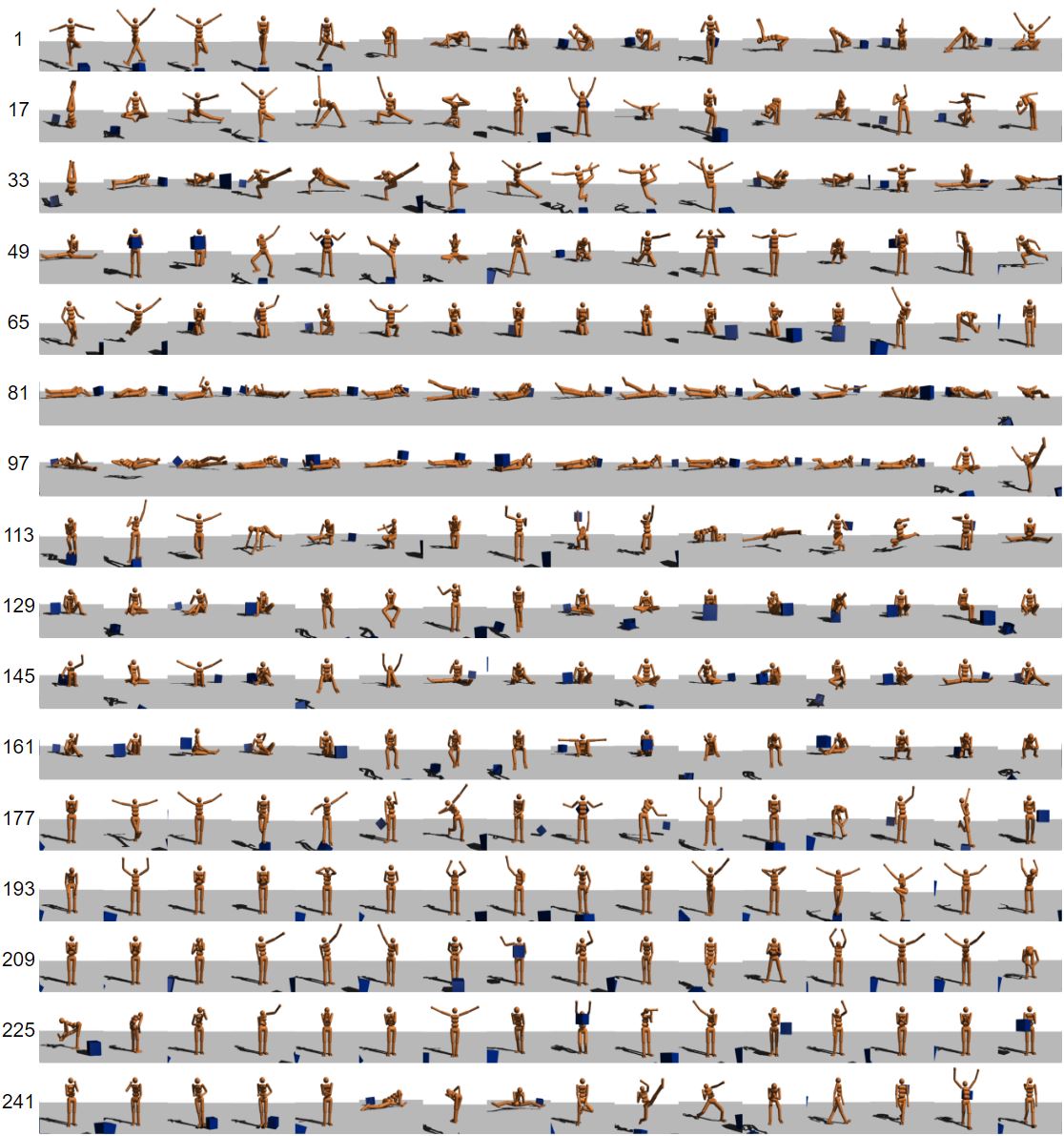}}
\caption{Front views of configurations retrieved from the multiview embedding-diversity dataset, then finetuned using a multiview distilled model.}
\label{fig:finetuned_cov_3v}
\end{center}
\end{figure}

\begin{figure}
\begin{center}
\centerline{\includegraphics[width=.9999\linewidth]{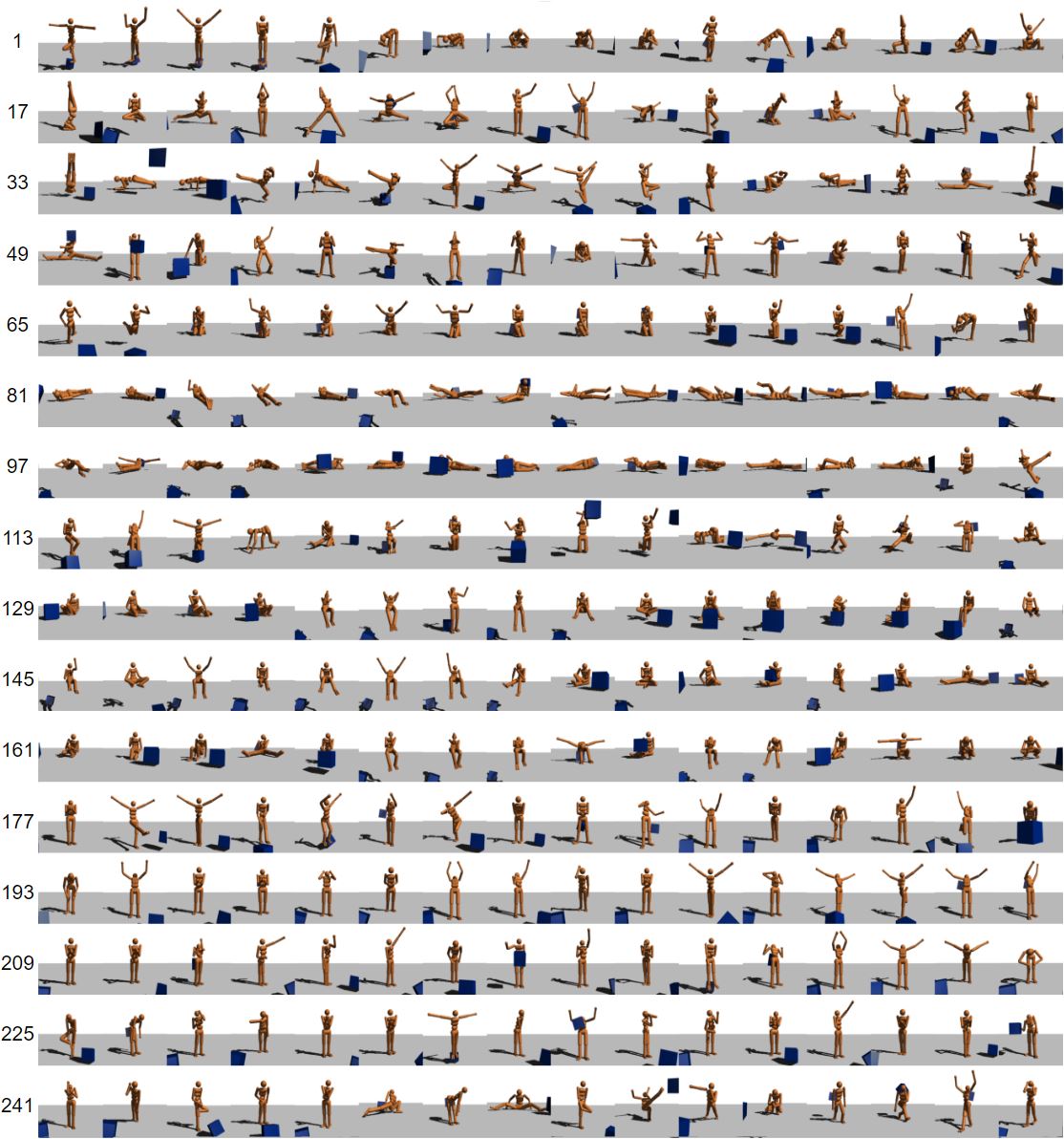}}
\caption{Front views of configurations retrieved from the single-view embedding-diversity dataset, finetuned using a single-view distilled model, then selected using EVA02-E-14+ to evaluate single-view scores.}
\label{fig:selected_cov_1v}
\end{center}
\end{figure}

\begin{figure}
\begin{center}
\centerline{\includegraphics[width=.9999\linewidth]{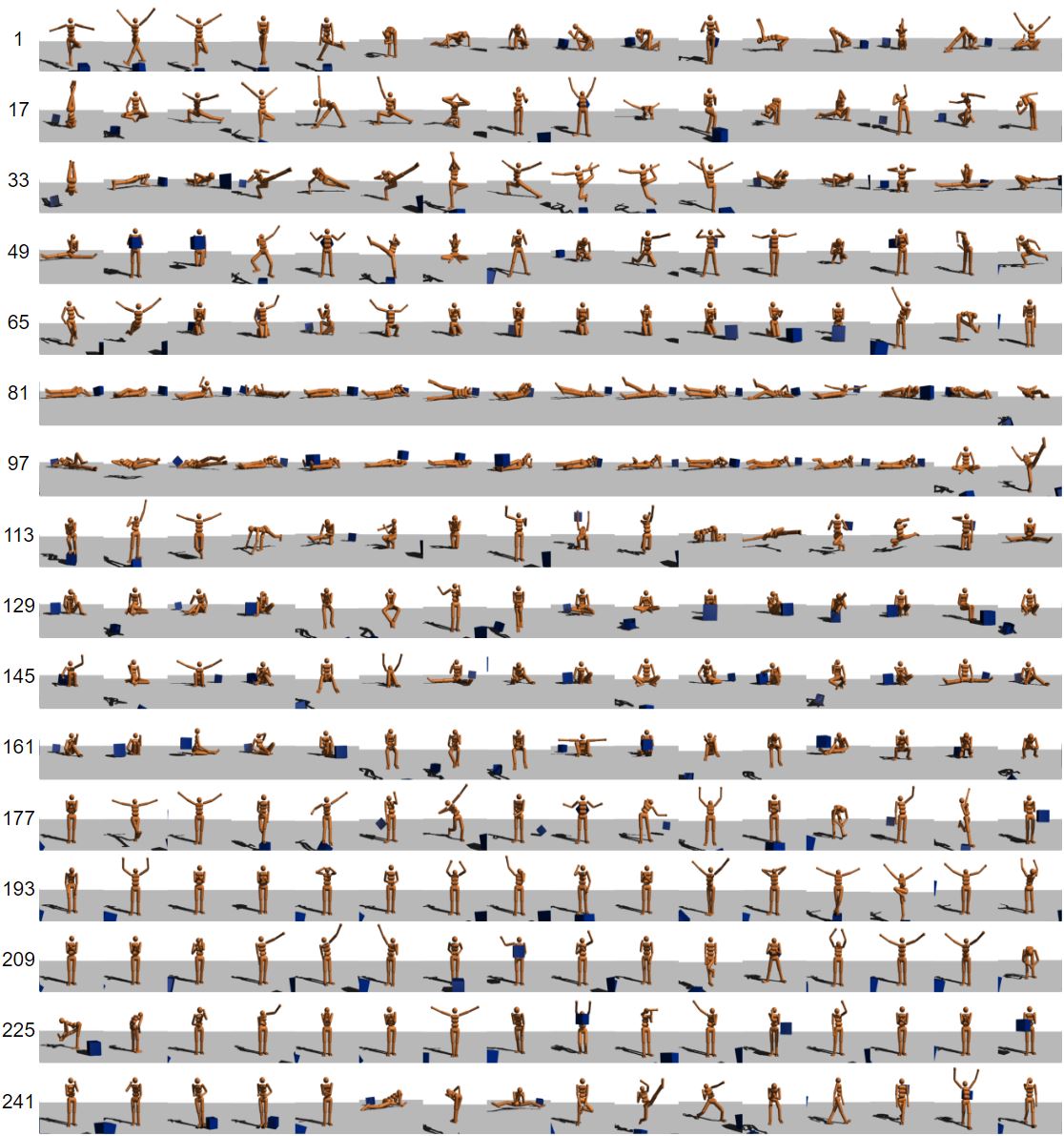}}
\caption{Front views of configurations retrieved from the multiview embedding-diversity dataset, finetuned using a multiview distilled model, then selected using  EVA02-E-14+ to evaluate multiview scores.}
\label{fig:selected_cov_3v}
\end{center}
\end{figure}

\begin{itemize}[itemsep=0.5mm, parsep=0mm]
\item Figures~\ref{fig:retrieved_rand_1v} and~\ref{fig:retrieved_rand_3v} show configurations retrieved from the random-policy dataset, for single-view and multiview scores respectively. 
\item Figures~\ref{fig:retrieved_cov_1v} and~\ref{fig:retrieved_cov_3v} show configurations retrieved from the embedding-diversity dataset, for single-view and multiview scores respectively.  
\item Figures~\ref{fig:finetuned_cov_1v} and~\ref{fig:finetuned_cov_3v} show configurations retrieved from the embedding-diversity dataset and finetuned with the distilled model, for single-view and multiview scores respectively.  
\item Figures~\ref{fig:selected_cov_1v} and~\ref{fig:selected_cov_3v} show configurations retrieved from the embedding-diversity dataset, finetuned with the distilled model, then selected with the VLM, for single-view and multiview scores respectively.  
\end{itemize}


\subsection{Reached Configurations}
\label{sec:reached-configs}
The following figures show frontal views of the best-in-trajectory configurations reached by our LCAs and by the MTRL baselines during a single rollout, based on the multiview text-configuration score.

\begin{figure}
\begin{center}
\centerline{\includegraphics[width=.9999\linewidth]{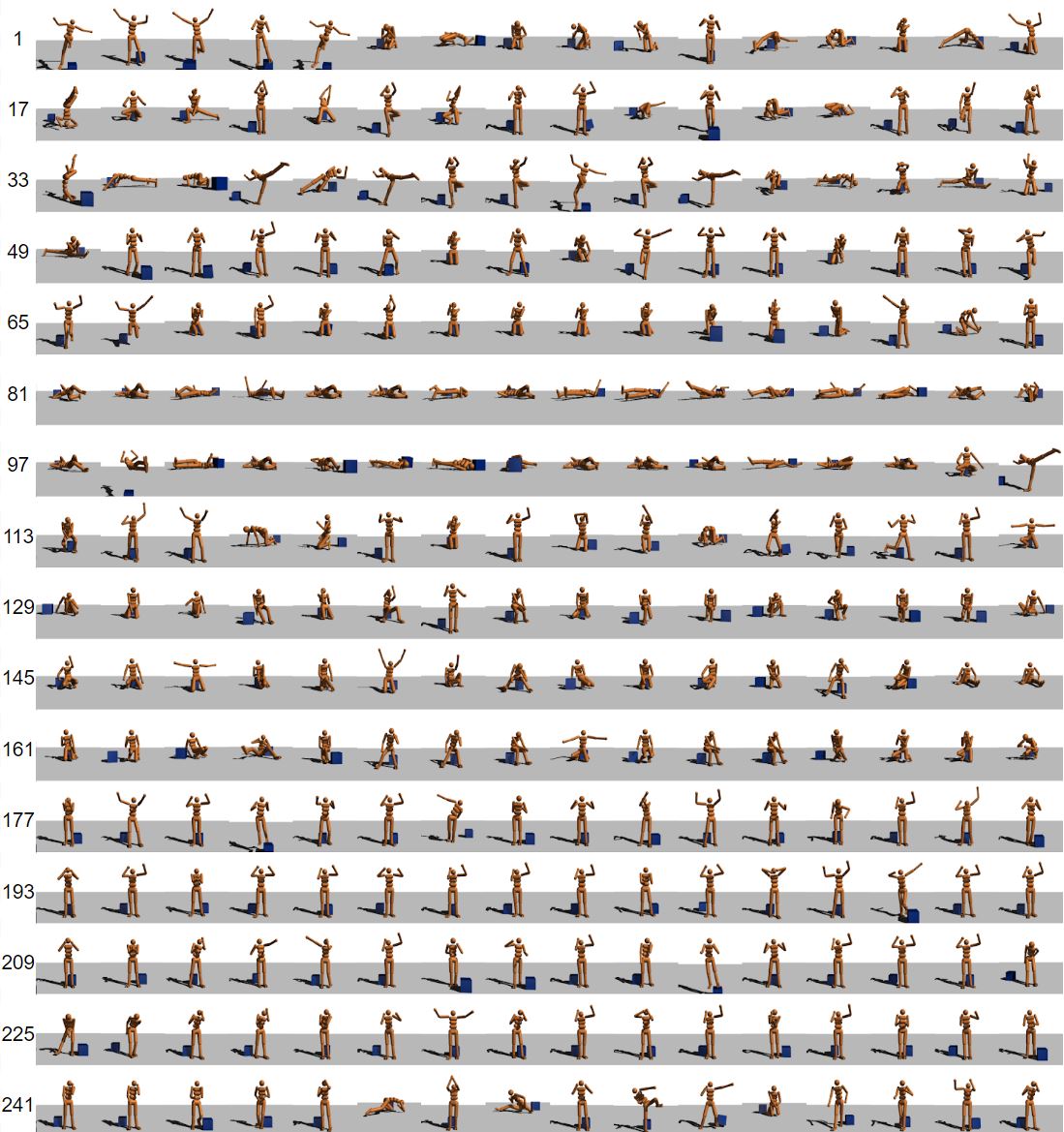}}
\caption{Front views of the best-in-trajectory configurations reached by GCRL-R based on multiview text-configuration score evaluated by EVA02-E-14+.}
\label{fig:reached_rand_3v}
\end{center}
\end{figure}

\begin{figure}
\begin{center}
\centerline{\includegraphics[width=.9999\linewidth]{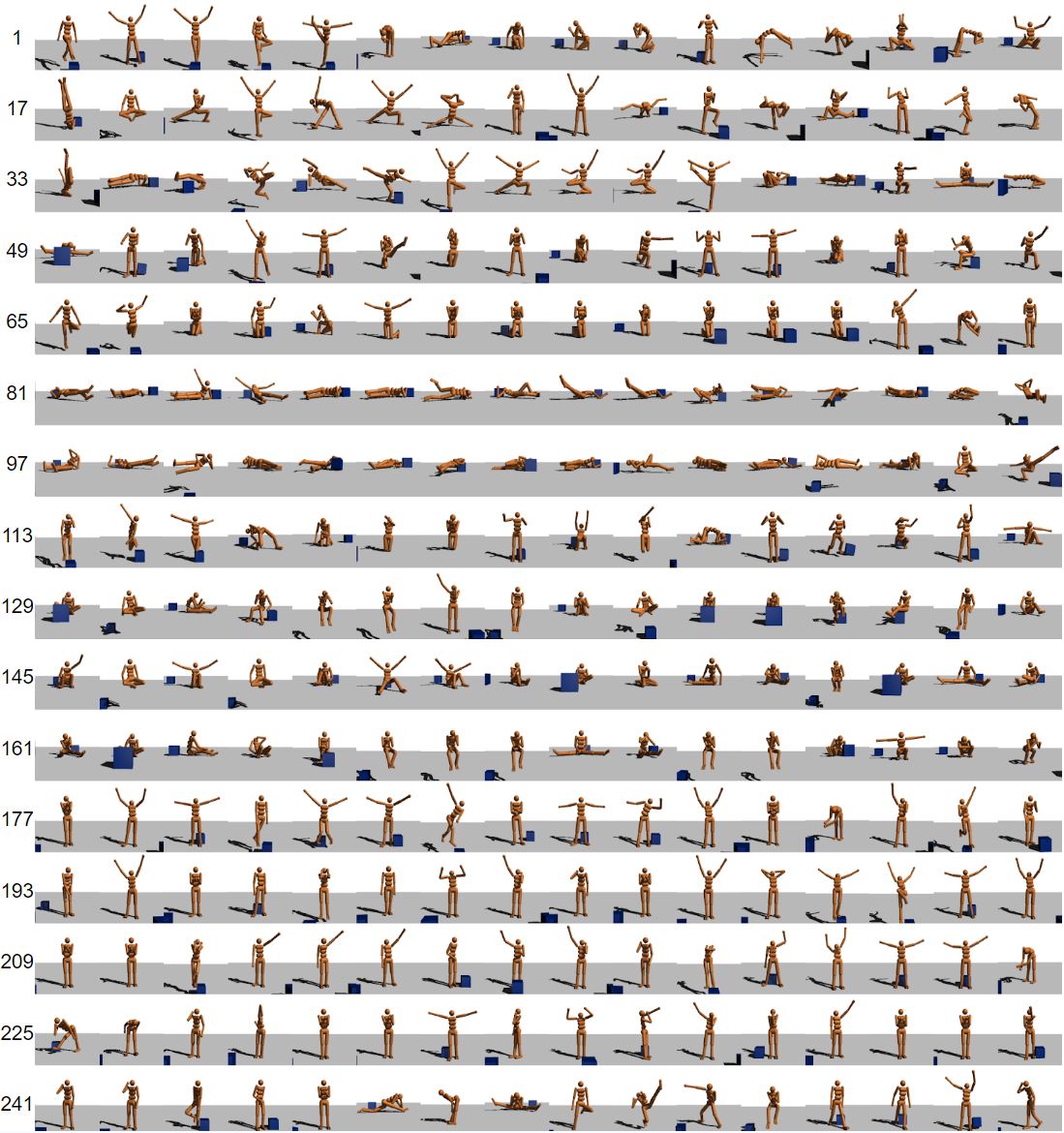}}
\caption{Front views of the best-in-trajectory configurations reached by GCRL-D based on multiview text-configuration score evaluated by EVA02-E-14+.}
\label{fig:reached_cov_3v}
\end{center}
\end{figure}

\begin{figure}
\begin{center}
\centerline{\includegraphics[width=.9999\linewidth]{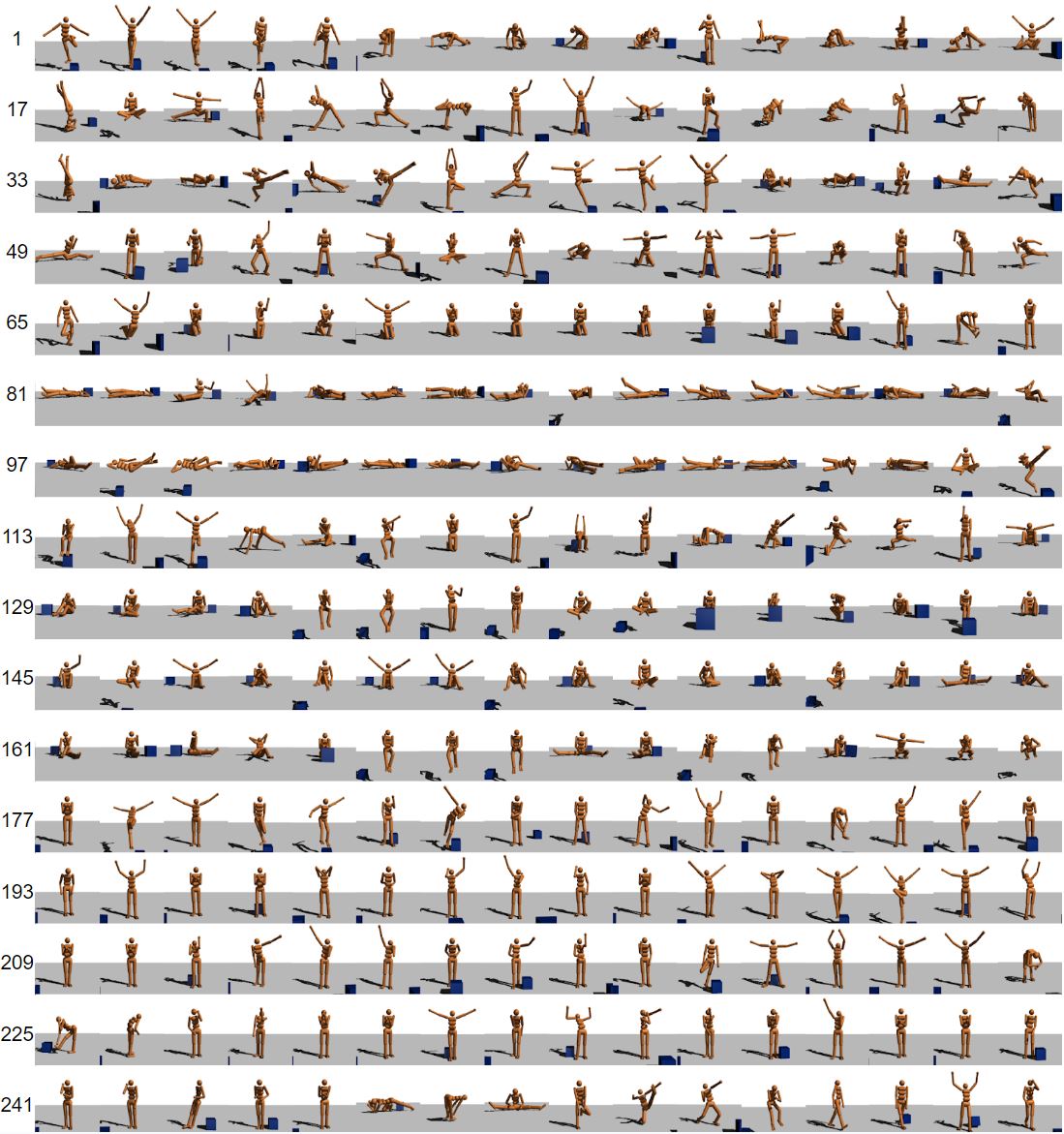}}
\caption{Front views of the best-in-trajectory configurations reached by GCRL-F based on multiview text-configuration score evaluated by EVA02-E-14+.}
\label{fig:reached_fine_3v}
\end{center}
\end{figure}

\begin{figure}
\begin{center}
\centerline{\includegraphics[width=.9999\linewidth]{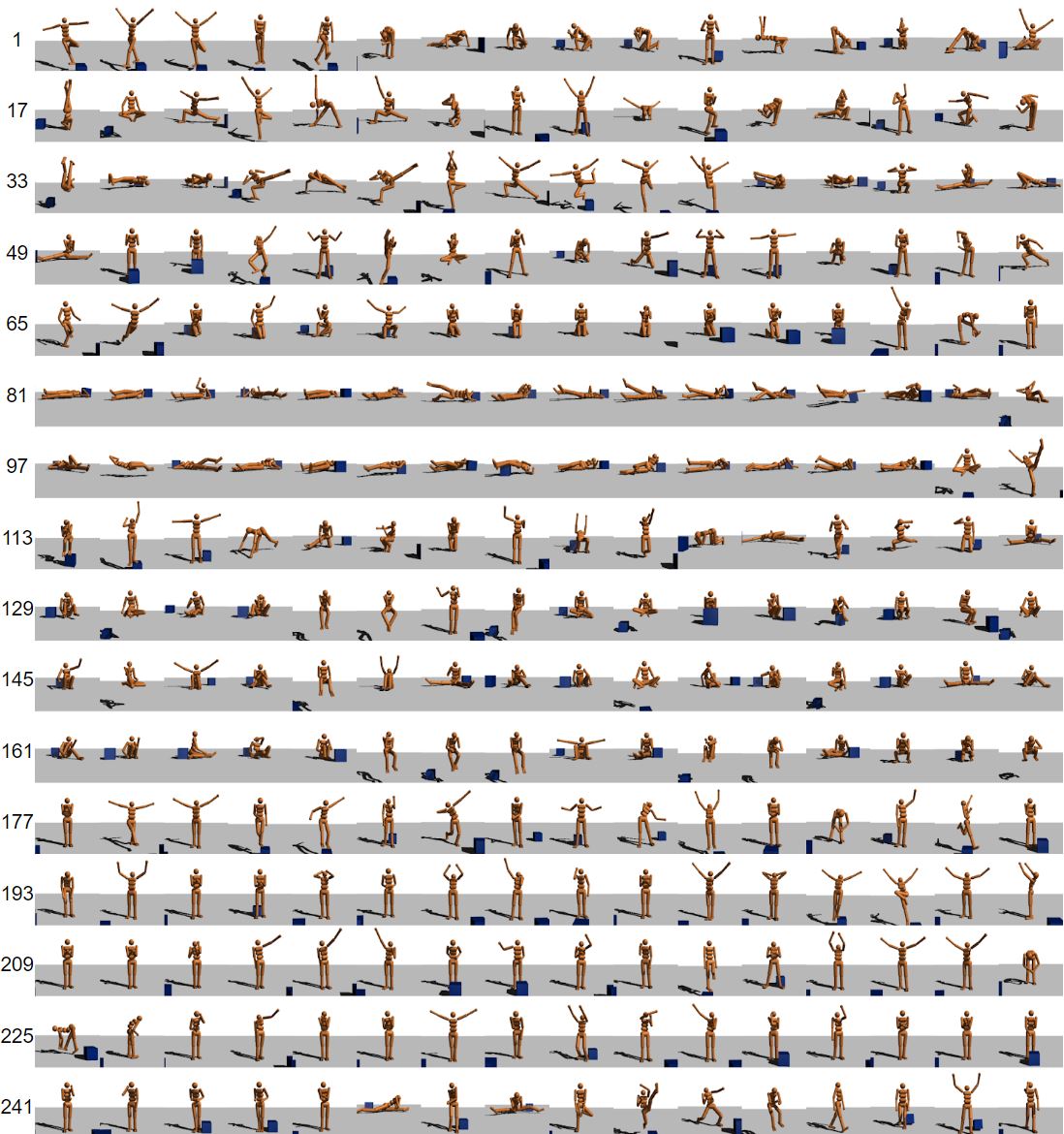}}
\caption{Front views of the best-in-trajectory configurations reached by GCRL-S based on multiview text-configuration score evaluated by EVA02-E-14+.}
\label{fig:reached_sel_3v}
\end{center}
\end{figure}

\begin{figure}
\begin{center}
\centerline{\includegraphics[width=.9999\linewidth]{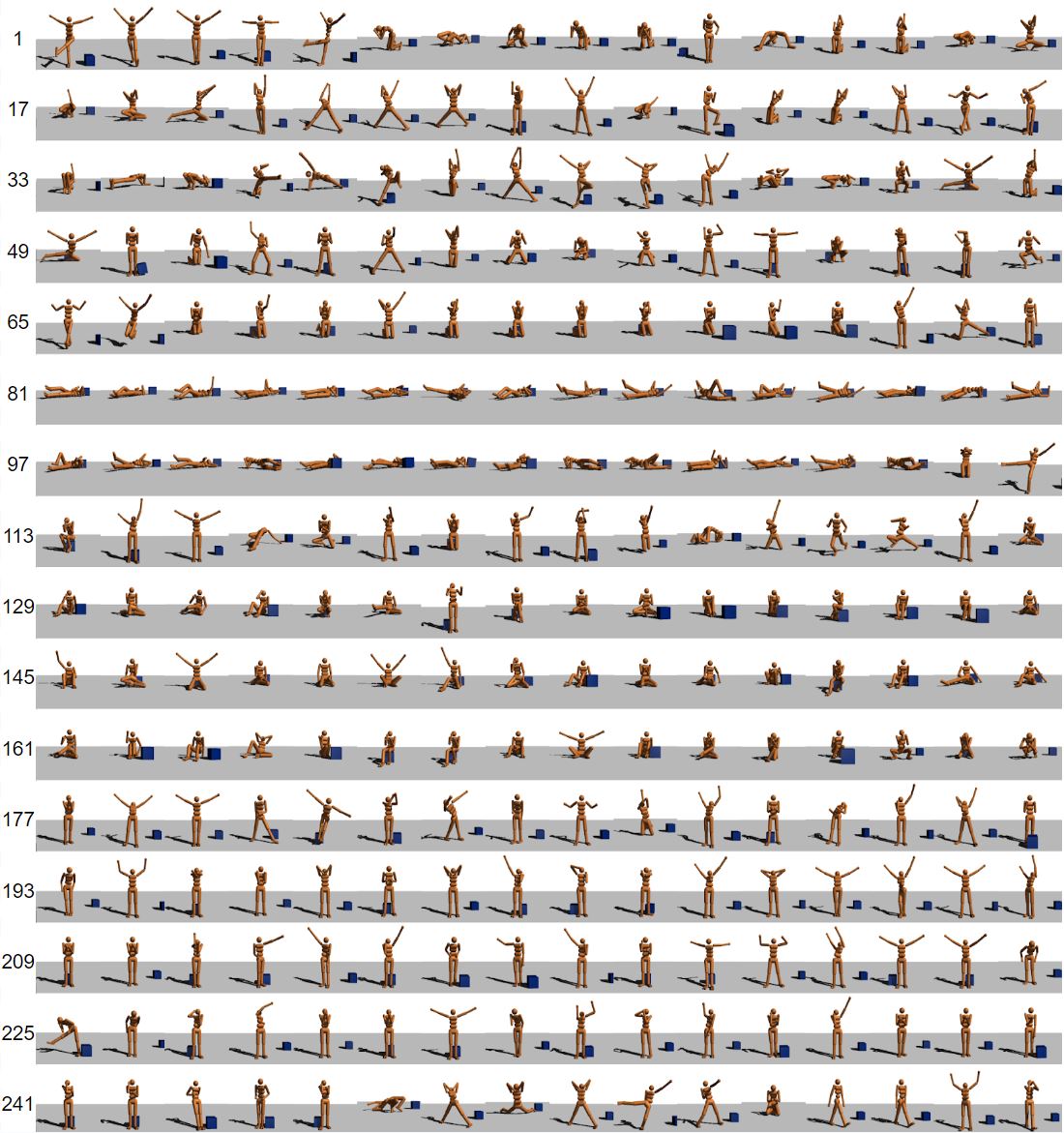}}
\caption{Front views of the best-in-trajectory configurations reached by the MTRL baseline on its training tasks, based on multiview text-configuration score evaluated by EVA02-E-14+.}
\label{fig:reached_mtrltrain_3v}
\end{center}
\end{figure}

\begin{figure}
\begin{center}
\centerline{\includegraphics[width=.9999\linewidth]{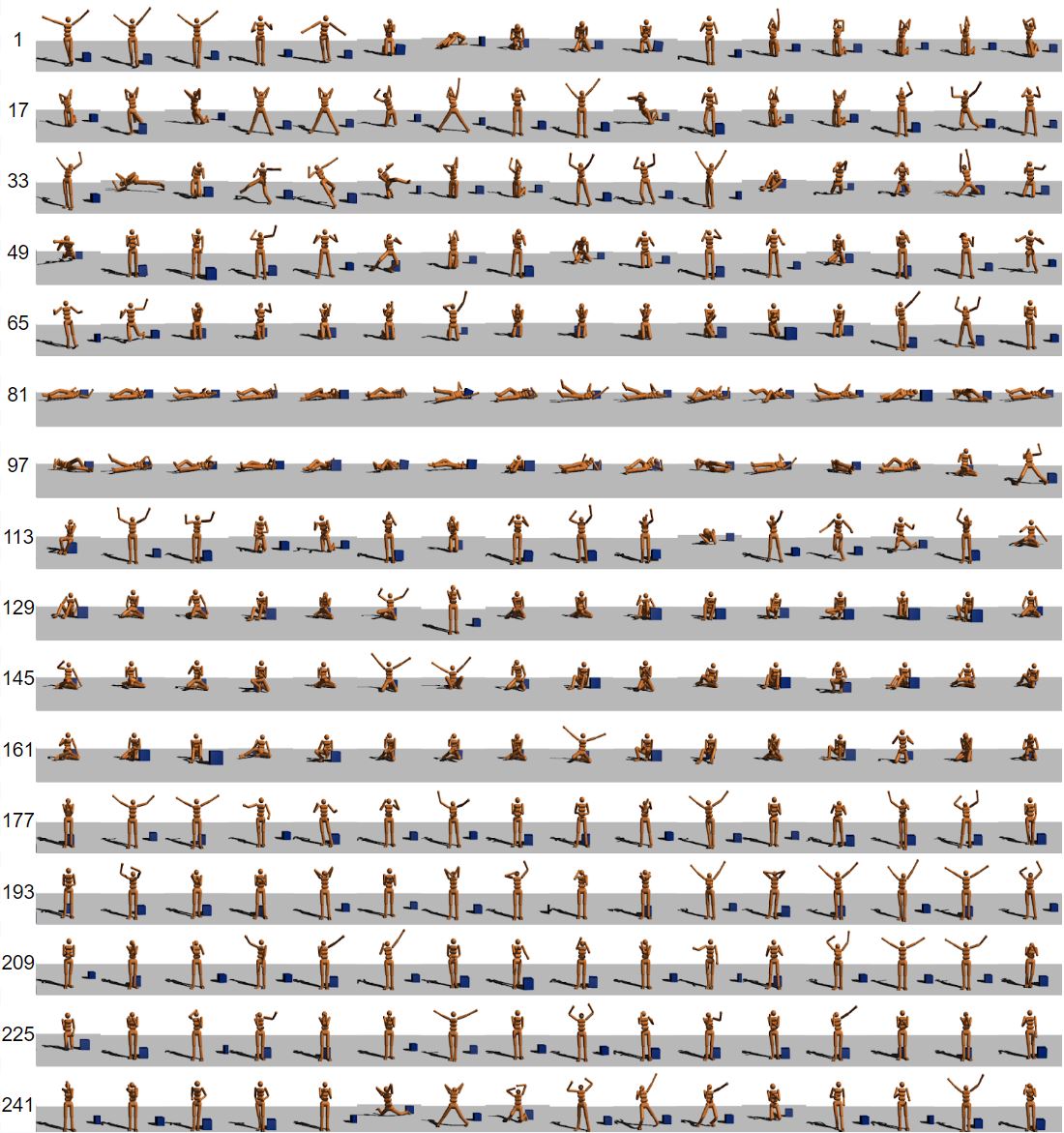}}
\caption{Front views of the best-in-trajectory configurations reached by the MTRL baseline on its test tasks, based on multiview text-configuration score evaluated by EVA02-E-14+.}
\label{fig:reached_mtrltest_3v}
\end{center}
\end{figure}

\begin{itemize}[itemsep=0.5mm, parsep=0mm]
\item Figures~\ref{fig:reached_rand_3v} show the best-in-trajectory configurations reached by GCRL-R.
\item Figures~\ref{fig:reached_cov_3v} show the best-in-trajectory configurations reached by GCRL-D.
\item Figures~\ref{fig:reached_fine_3v} show the best-in-trajectory configurations reached by GCRL-F.
\item Figures~\ref{fig:reached_sel_3v} show the best-in-trajectory configurations reached by GCRL-S.
\item Figures~\ref{fig:reached_mtrltrain_3v} show the best-in-trajectory configurations reached by MTRL evaluated on its training tasks.
\item Figures~\ref{fig:reached_mtrltest_3v} show the best-in-trajectory configurations reached by MTRL evaluated on its test tasks.
\end{itemize}

\end{document}